\documentclass[preprint,12pt,3p]{elsarticle}

\usepackage[utf8]{inputenc}
\usepackage{algpseudocode}
\usepackage[T1]{fontenc}
\usepackage{subcaption}
\usepackage{pdflscape}
\usepackage{adjustbox}
\usepackage{algorithm}
\usepackage{amsfonts}
\usepackage{booktabs}
\usepackage{dashrule}
\usepackage{hyperref}
\usepackage{graphicx}
\usepackage{multicol}
\usepackage{multirow}
\usepackage{tabularx}
\usepackage{amsmath}
\usepackage{xcolor}
\usepackage{lineno}
\usepackage{tikz}
\usepackage{array}
\usepackage[
    justification=justified
]{caption}

\journal{Elsevier}
\biboptions{sort&compress}
\hypersetup{
  colorlinks=true, 
  urlcolor=blue,   
  linkcolor=red,   
  citecolor=green  
}

\begin{document}
\begin{frontmatter}

\title{Multi-Path Long-Term Vessel Trajectories Forecasting\\with Probabilistic Feature Fusion for Problem Shifting}

\author[1,2,3]{Gabriel Spadon\corref{equal}}
\ead{spadon@dal.ca}
\ead[url]{www.spadon.com.br}
\author[1,2]{Jay Kumar\corref{equal}}
\author[4]{Derek Eden}
\author[4]{Josh van Berkel}
\author[4]{Tom\\Foster}
\author[5]{Amilcar Soares}
\author[3]{Ronan Fablet}
\author[1,6]{Stan Matwin}
\author[2]{Ronald Pelot}

\cortext[equal]{These authors contributed equally to this work.}

\affiliation[1]{organization={Institute for Big Data Analytics, Dalhousie University}, addressline={Halifax -- NS, Canada}}
\affiliation[2]{organization={Industrial Engineering Department, Dalhousie University, Halifax -- NS, Canada}}
\affiliation[3]{organization={Departement de Genie Mathematique et Electrique, IMT Atlantique}, addressline={Brest, France}}
\affiliation[5]{organization={Department of Comp. Sci. and Media Tech., Linnaeus University}, addressline={Vaxjo, Sweden}}
\affiliation[6]{organization={Institute of Computer Science, Polish Academy of Sciences}, addressline={Warsaw, Poland}}
\affiliation[4]{organization={DHI Water \& Environment Inc.}, addressline={Ottawa, Canada}}

\begin{abstract}
This paper addresses the challenge of boosting the precision of multi-path long-term vessel trajectory forecasting on engineered sequences of Automatic Identification System (AIS) data using feature fusion for problem shifting. We have developed a deep auto-encoder model and a phased framework approach to predict the next 12 hours of vessel trajectories using 1 to 3 hours of AIS data as input. To this end, we fuse the spatiotemporal features from the AIS messages with probabilistic features engineered from historical AIS data referring to potential routes and destinations. As a result, we reduce the forecasting uncertainty by shifting the problem into a trajectory reconstruction problem. The probabilistic features have an F1-Score of approximately 85\% and 75\% for the vessel route and destination prediction, respectively. Under such circumstances, we achieved an R2 Score of over 98\% with different layer structures and varying feature combinations; the high R2 Score is a natural outcome of the well-defined shipping lanes in the study region. However, our proposal stands out among competing approaches as it demonstrates the capability of complex decision-making during turnings and route selection. Furthermore, we have shown that our model achieves more accurate forecasting with average and median errors of 11km and 6km, respectively, a 25\% improvement from the current state-of-the-art approaches. The resulting model from this proposal is deployed as part of a broader Decision Support System to safeguard whales by preventing the risk of vessel-whale collisions under the \textit{smartWhales} initiative and acting on the Gulf of St. Lawrence in Atlantic Canada.
\end{abstract}



\end{frontmatter}



\section{Introduction \& Background}

\begin{figure}[!h]
    \centering
    \includegraphics[width=.9\linewidth]{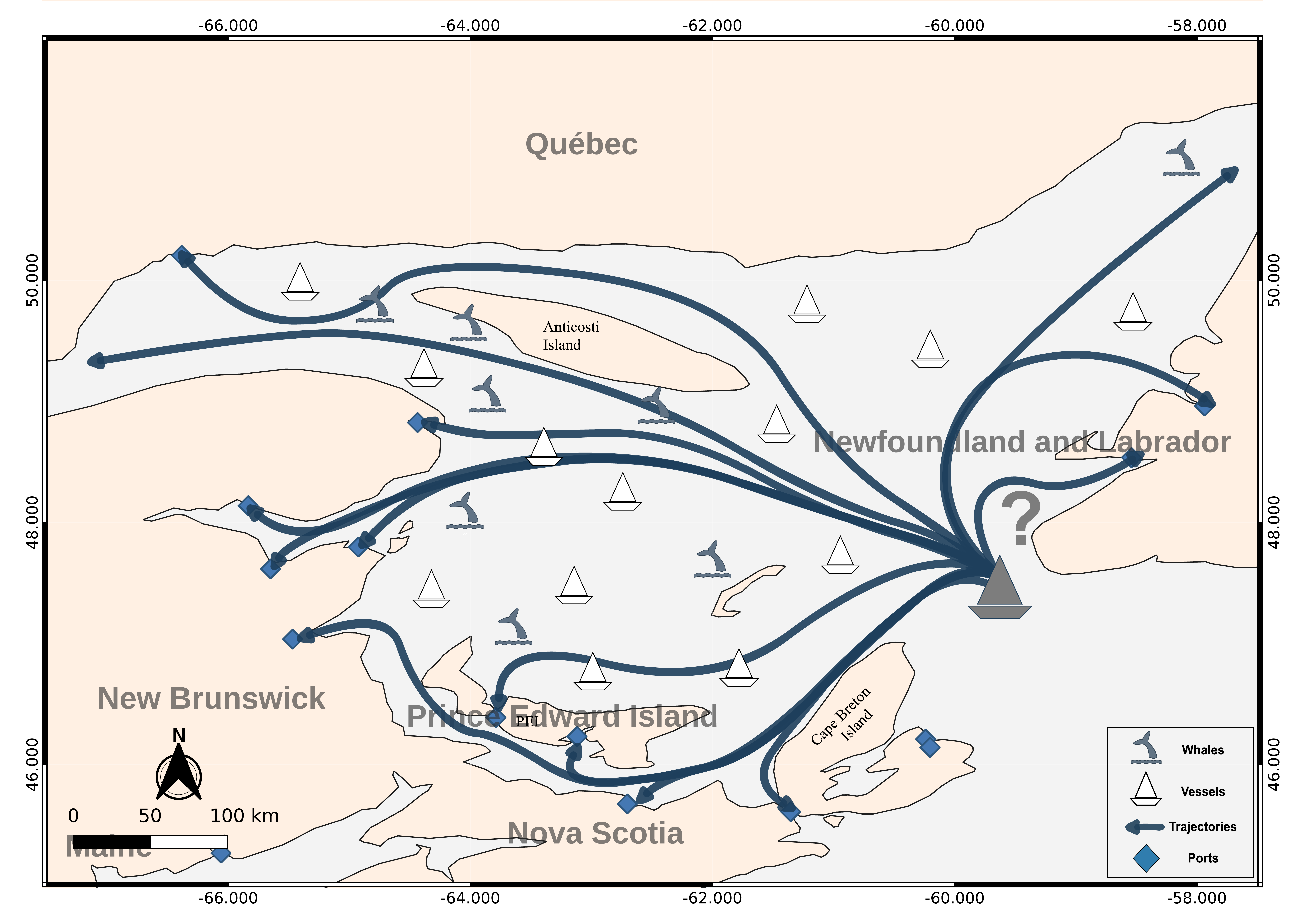}
    \caption{Potential routes a vessel may take to enter the Gulf of St. Lawrence through the Cabot Strait. Whale and vessel symbols represent high vessel traffic in the region and the possibility of marine life encounters along the possible routes. Possible destinations include Quebec, New Brunswick, Nova Scotia, Newfoundland and Labrador ports. A vessel can anchor in one of these ports, continue into the St. Lawrence River, or exit the gulf through the Strait of Belle Isle. The figure illustrates that vessels can have a common start point but may take different routes, resulting in different trajectories.}
    \label{fig:vessel-decision}
\end{figure}

The International Maritime Organization (IMO) is essential in enhancing maritime safety, efficient navigation, and the prevention of marine pollution by ships~\cite{IMO}.
One of the key technologies used to achieve these goals is the Automatic Identification System (AIS), which tracks marine vessels' locations worldwide.
Canadian regulations require every vessel planning to enter Canada to equip themselves with an AIS transceiver for sharing and receiving navigational data (Section 65 of the Navigation Safety Regulations\footnote{~https://laws-lois.justice.gc.ca/eng/regulations/SOR-2020-216}).
The AIS system uses a network of satellites and ground-based stations to track vessels' location, speed, and other relevant data in real-time.
AIS allows for the real-time tracking of vessels, enhancing safety measures by improving the visibility of vessel movements for individual ships and shore-based facilities like the Coast Guard.
By actively using AIS data in maritime applications, authorities can track vessels, monitor their movements, and ensure compliance with regulations.
These organizations can act in distress situations to prevent incidents, such as collisions~\cite{fournier2018past, Goerlandt2011:ShipCollision, Chen2019:ShipCollision}, enhancing safety regardless of their current route (see Figure~\ref{fig:vessel-decision}).

Research on AIS data has opened the doors to enhance maritime safety and security through solving various issues such as the detection of suspicious events~\cite{campbell2022detection}, monitoring traffic to avoid collisions~\cite{rong2020data, spadon2022unfolding}, and evaluating noise or environmental pollution levels~\cite{peng2022establishment, pichegru2022maritime}.
Recently, the problem of forecasting long-term trajectories~\cite{nguyen2018multi, patmanidis2016maritime, zhang2018ais, uney2019data} of vessels gained significant attention to reduce vessel accidents, detect anomalous trajectories, and improve path planning --- especially for autonomous shipping~\cite{faghih2014accident, le2013unsupervised, d2018maritime, forti2019unsupervised, forti2019anomaly, ferreira2022semi, newaliya2021review}.
The study of trajectory forecasting has a diverse background, where different techniques have been employed to create models that predict vessel movement, either by observing a single vessel's data or different vehicles.
One of the most basic models is the constant velocity model \cite{DBLP:journals/tits/XiaoFZG20}, derived from the AIS messages' geographical coordinates and the vessel's Speed over Ground (SOG) and Course over Ground (COG) at the last recorded timestep, assuming no change in speed and course for the forecasting duration.
Other models surpass these assumptions and use the trajectory's geometry to produce predictions, such as the Ornstein--Uhlenbeck process is adapted for vessel trajectory forecasting~\cite{pallotta2014context}.
Deep Learning has advanced significantly in recent years, allowing for the creation of models that combine different data types.
However, the real-world use of these models is often limited due to the difficulty of synchronizing and preparing diverse data in streaming conditions.

In a recent study by Nguyen et al. (2024)~\cite{nguyen2024}, authors proposed an approach to represent four AIS-related features into embeddings, {\it i.e.}, a learned numerical representation of data where similar data points have a similar representation based on their meaning.
Their study adopted the transformer encoder in a generative-driven approach to forecasting long-term trajectories, as proposed by Nguyen et al. (2021)~\cite{DBLP:journals/corr/abs-2109-03958}.
However, their technique confines the ship's trajectory to the centroid of the grid cells, requiring longer training sessions on a large model so it can learn the many patterns that distinguish each grid cell.
This approach may be effective when ships adhere to nearly constant velocity models, but its potential to learn more individual moving patterns is overlooked.
Besides this, the authors implemented their solution using teacher forcing~\cite{NIPS2016_16026d60}.
Instead of inputting the previous step predictions as input to the next step, the technique trains the network with ground truth data, as that information is known during the training phase.
During testing, the network is rewired to revert the need for ground truth data, expecting that the learned weights during forced training will perform better in the testing phase.
Specifically, the authors forecast information about the vessel's speed and course, which can be mathematically calculated when having information about consecutive coordinates.
Therefore, unless the model provides a perfect estimation of the vessel's speed and course at every time step, the information used for predicting the next step of the sequence tends to be uncertain.

In different studies, Capobianco et al. (2021)~\cite{DBLP:journals/taes/CapobiancoMFBW21} and Capobianco et al. (2022)~\cite{TAES-823} have reviewed the effectiveness of recurrent encoder-decoder neural networks in trajectory forecasting.
In both studies, the authors utilized comparable network architectures that employ an encoder to produce an embedding; this embedding is then utilized in an attention mechanism to allow for multi-step forecasting and take advantage of both short and long-term dependencies learning, and the network decoder decodes the embedding and yields the forecasted trajectory.
However, the issue in their modeling approach is not related to the auto-encoder model but rather to how the attention mechanism is employed.
When attention is used over a sentence as seen in Large Language Models (LLMs) such as ChatGPT\footnote{~https://openai.com/chatgpt} or Bert\footnote{~https://github.com/google-research/bert}, the mechanism learns the significance (or importance) of words in the sentence, helping the model to focus on the key points of the sentence and respond accordingly to the LLM operator.
However, sequences of AIS messages and sentences ({\it i.e.}, sequences of words) have intrinsically different abstractions.
In an AIS forecasting model, the knowledge that can be leveraged from the beginning of the sequence is less important than the one that can be extracted from later points in the same sequence because of the spatiotemporal nature of the data.
While the intermediate points of the sequence can aid in more accurate long-term trajectory forecasting, they also increase the uncertainty of the modeling problem in the short term because the model does not follow the correct order of events in the sequence.

Using a different approach, this paper advances previous works by using an equal-size hexagon grid for extracting probabilistic features that aid in identifying the possible route and destination of different vessels --- these features allow for a finer approximation of the vessel path without being constrained by the centroids of the cells.
This means that, differently from Nguyen et al. (2024)~\cite{nguyen2024}, we tackle this problem in a regression rather than a classification fashion --- where the ocean grid is used to distill spatial knowledge that will be fused with the existing trajectory data instead of working as an anchor for the trajectory forecasting task.
Accordingly, our approach involves shifting the forecasting problem into a reconstruction by leveraging partial information about the future trajectory distilled from the grid system we previously created.
Then, by fusing engineered data about the route and destination obtained from our spatial grid system with the AIS messages, we gain a deeper understanding of vessel behavior and can improve the precision of the forecasted ({\it i.e.}, reconstructed) trajectory.
As a key component of our approach, we developed a deep neural network model.
This model has the capability to predict 12 hours of AIS data based on 3 hours of input data, with one AIS message being sent every 10 minutes.
The model was designed to operate over the Gulf of St. Lawrence in Canada, where vessels can take different routes to reach various regional ports, with the purpose of identifying the paths that cargo and tanker vessels will take in an area with many possible destinations (see Figure~\ref{fig:vessel-decision}).

The proposed model utilizes a spatio-temporal encoder-decoder architecture designed to enhance the trajectory reconstruction precision through phased feature learning and targeted attention mechanisms.
The encoder comprises stacked CNN, each equipped with dual $1\mathbf{D}$ convolutions.
These layers are specifically engineered to extract both short- and long-term dependencies from the input features using different kernel sizes, preparing a comprehensive spatial context for subsequent temporal processing.
Following the CNN, the model employs a recurrent neural network (RNN) to handle the sequential data, which captures the temporal dynamics essential for understanding how the features evolve.
This process is augmented with a positional-aware attention mechanism, pivotal in prioritizing more recent timesteps of the input sequence.
Unlike the architectures described by Capobianco et al. (2021)~\cite{DBLP:journals/taes/CapobiancoMFBW21} and Capobianco et al. (2022)~\cite{TAES-823}, our model's attention mechanism strategically emphasizes the latest data points, thereby ensuring they have a greater impact on the learning process.
This methodological focus on the most current and relevant data points allows for a seamless and dynamic transition from input to output, enhancing the precision of the trajectory reconstruction task.
By integrating these refined spatial and temporal features, our model produces accurate, timely, and contextually relevant predictions, reducing the uncertainty of long-term multi-path trajectory forecasting.

We conducted a comprehensive series of experiments to assess the performance of our model under various conditions.
When assessing the capability of the spatial grid system to distill trustworthy information about the future location of a given trajectory, we achieved an F1-Score of approximately 85\% and 75\% for predicting vessel routes and destinations.
At the same time, by excluding the engineered data from the spatial grid with the trajectory data and using basic features (such as coordinates, speed, and course information), the model's decision-making capacity was limited.
This was evident as the model excelled in predicting straightforward trajectories but struggled with complex or curved paths, initially showing mean and median errors of 13 km and 8 km, respectively.
However, we significantly improved decision-making and reduced these errors by integrating probabilistic features showing mean and median errors of 11 km and 6 km, respectively.
Remarkably, our solution achieved an $R^2$ score above 98\% for cargo and tanker vessels.
This high score primarily resulted from the frequent occurrence of nearly straight paths in the dataset.
Nonetheless, our model demonstrated superior performance in navigating complex routes and responding to sudden changes in trajectory.
Therefore, this research sets a new benchmark in long-term multi-path vessel trajectory forecasting and holds significant implications for marine safety and conservation efforts.
By employing spatio-temporal data, probabilistic feature modeling, and sophisticated neural network techniques, we have paved the way for policies to increase marine safety and protect endangered species like the North Atlantic Right Whale (NARW) as a use case.
The resulting model from this study is currently deployed as part of a broader Decision Support System to safeguard whales by preventing the risk of vessel-whale collisions under the \textit{smartWhales} initiative and acting on the Gulf of St. Lawrence in Atlantic Canada.

This paper is structured into five sections besides the introduction, which offers an overview of the problem, existing literature, and the solutions we propose.
Specifically, Section~\ref{sec:problem} covers the core problem of this research and outlines our solution that leverages a spatial grid system for knowledge extraction.
Section~\ref{sec:proba-model} details the probabilistic model we developed to encapsulate spatial information pertinent to the routes and destinations of vessel trajectories.
Section~\ref{sec:deep-model} discusses the deep neural network architecture tailored for trajectory reconstruction tasks.
Section~\ref{sec:results} presents our findings, providing a comprehensive discussion on the capacity of our proposed model.
Finally, Section~\ref{sec:conclusions} summarizes our conclusions and final remarks.
Supplementary concepts and definitions are provided in the Annex.

\section{Problem Statement \& Methodology}
\label{sec:problem}

The problem addressed in this paper and the achievements of the contributions were rooted in the Canadian Space Agency's \textit{smartWhales} initiative, which is a strategic response to the escalating issue of vessel-whale collisions in the Gulf of St. Lawrence.
The related project (1 of 5 executed through the program), executed by WSP\footnote{~{https://www.wsp.com/en-gl}} in collaboration with DHI Water \& Environment\footnote{~{https://www.dhigroup.com}}, Dalhousie University, the Institut des Sciences de la mer de Rimouski (ISMER)\footnote{~{https://www.ismer.ca}}, and the Canadian Whale Institute\footnote{~{https://www.canadianwhaleinstitute.ca}}, applies novel ecological modeling techniques and, among other data sources, Earth Observation Data to enable predictive hindcast and forecast modeling of NARW movements.
This and the vessel trajectory forecast model technique proposed in this paper formed the foundation of a Decision Support System capable of forecasting potential whale-life-threatening interactions between vessels and NARWs in the Gulf of St. Lawrence.
The vessel-whale collisions issue, our underlying motivation, gained particular spotlight in 2017, a year marked by the reported fatalities of at least 12 NARW, highlighting the considerable risks posed by maritime activities, including shipping and fishing, to the survival of this species~\cite{davies2019mass}.
With the constant threat of collision and the persisting NARW endangered status, creating a model for forecasting multi-path long-term ship trajectories came as one of the ideas for aiding in solving this issue.

\begin{figure}[!h]
    \centering
    \includegraphics[width=0.495\textwidth]{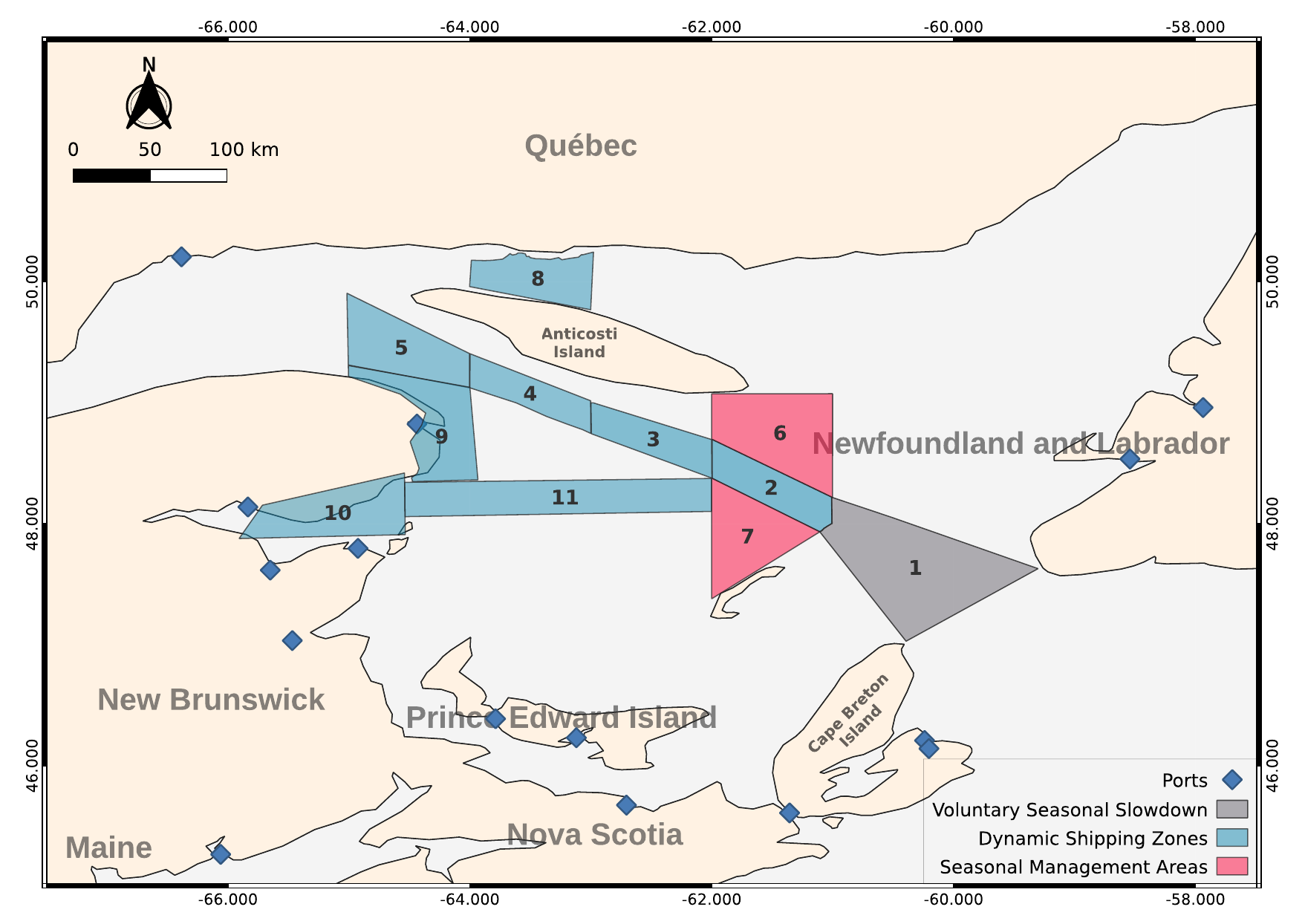} 
    \includegraphics[width=0.495\textwidth]{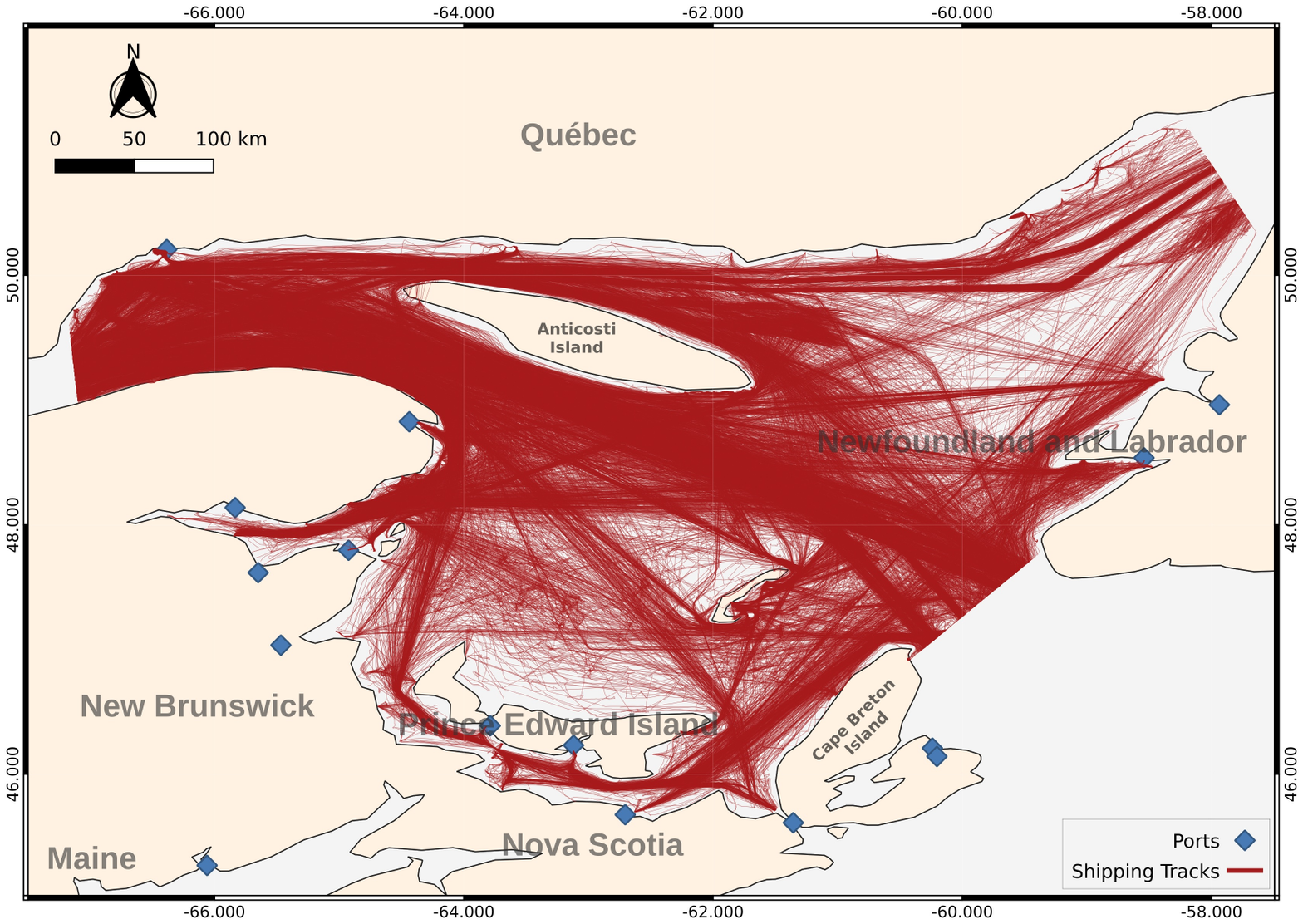}
    \caption{This study centers on the Gulf of St. Lawrence, with traffic policy zones defined by Transport Canada\protect\footnotemark. These zones include voluntary seasonal slowdowns, dynamic shipping zones, and seasonal management areas, which are illustrated on the left-hand map. The regulated areas overlay main vessel traffic routes, such as the main shipping corridor between the Cabot Strait (between Cape Breton Island and Newfoundland) and the estuary of the St. Lawrence River routes (polygons 1--5). These areas cover the ports in the zone between New Brunswick and Québec (polygons 6--12) and the Jacques Cartier Strait between Anticosti Island and Northern Quebec (polygon 8). The visualization on the right-hand side displays vessel tracks (2015--2020) in the area, revealing inconsistent paths within shipping lanes.}
    \label{fig:AIO-Traffic}
\end{figure}
\footnotetext{~https://bit.ly/45DilrO}

\begin{figure}[!h]
    \centering
    \includegraphics[width=.9\linewidth]{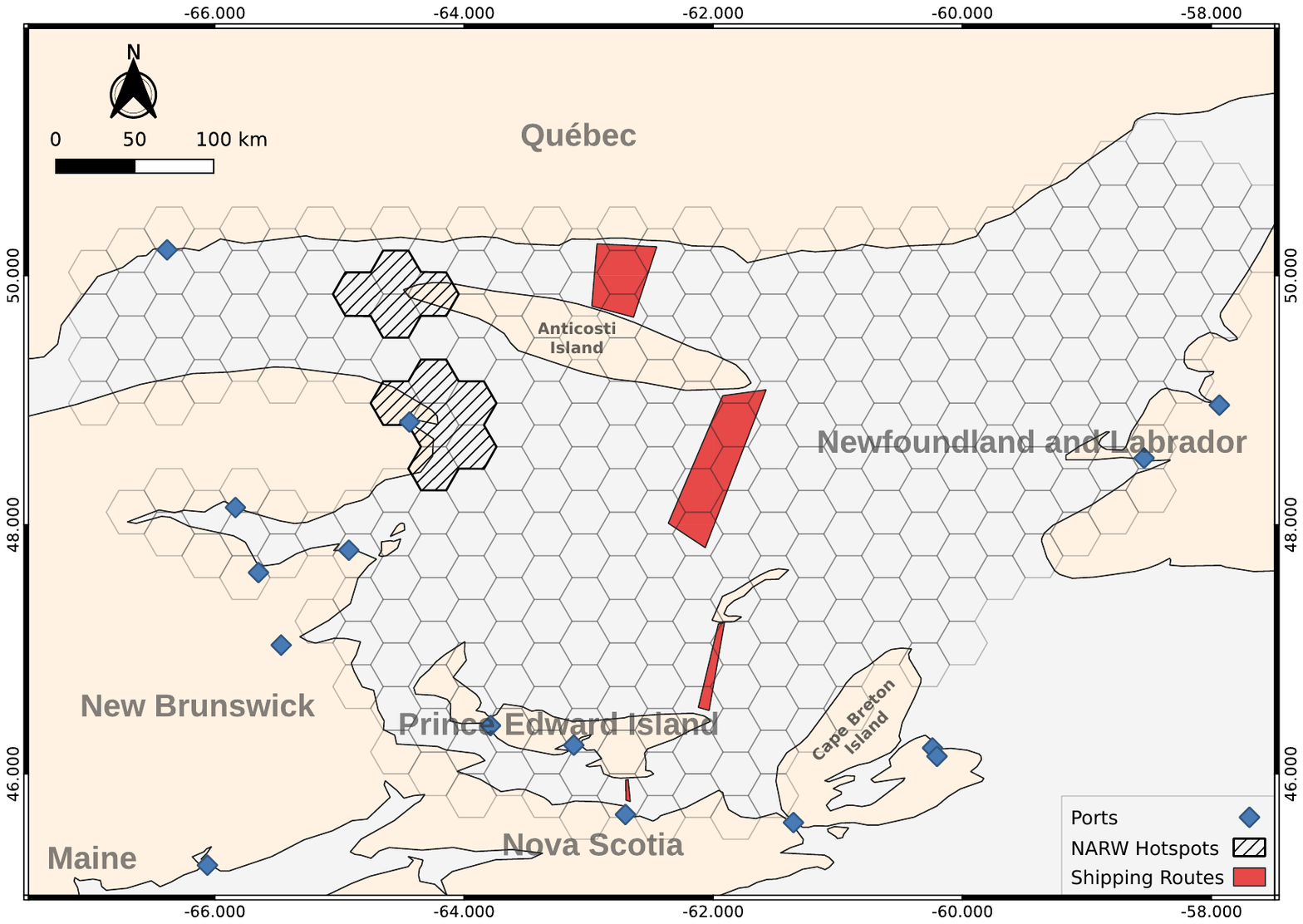}
    \caption{Gulf of St. Lawrence overlain with our hexagonal study grid. Hatched grid cells indicate hotspots for North Atlantic Right Whales (NARWs)~\cite{o2022repatriation}. Most vessels' trajectories intersect with the red route polygons at some point during their journey through the gulf. These polygons segment the gulf from north to south, covering major inbound and outbound movements inside the gulf. However, it is important to note that not all paths are required to cross a route polygon. Vessels whose paths cross the route polygons are generally easier to forecast, while those that do not pose intricate forecasting scenarios.}
    \label{fig:NARW-Paths}
\end{figure}

With such a model in hand and real-time data acquisition, the \textit{smartWhales} initiative can facilitate proactive mitigation steps that aim to minimize collision risks by providing advanced predictive analytics that anticipate the trajectories of vessels concerning the known habitats and movements of NARWs.
In this sense, by adjusting vessel routes (see Figure~\ref{fig:AIO-Traffic}{\bf B}) to avoid these critical areas (see Figure~\ref{fig:AIO-Traffic}{\bf A}), the \textit{smartWhales} initiative aims to contribute to a significant reduction in the likelihood of harmful vessel-whale encounters mainly when vessel paths navigate across collision hotspot areas (see Figure~\ref{fig:NARW-Paths}).
As vessels navigate, their historical data provide us with insights into their likely trajectory, along with observations on collective vessel movement patterns~\cite{spadon2022unfolding}; the same can be observed among people commuting~\cite{spadon2019reconstructing, lei2022forecasting, alves2021commuting}, patient trajectories in hospitals~\cite{rodrigues2021ligdoctor}, and disease-spreading phenomena~\cite{spadon2022evolution}.

\begin{figure}[!t]
    \centering
    \begin{subfigure}[b]{.475\textwidth}\centering
        \includegraphics[width=\textwidth]{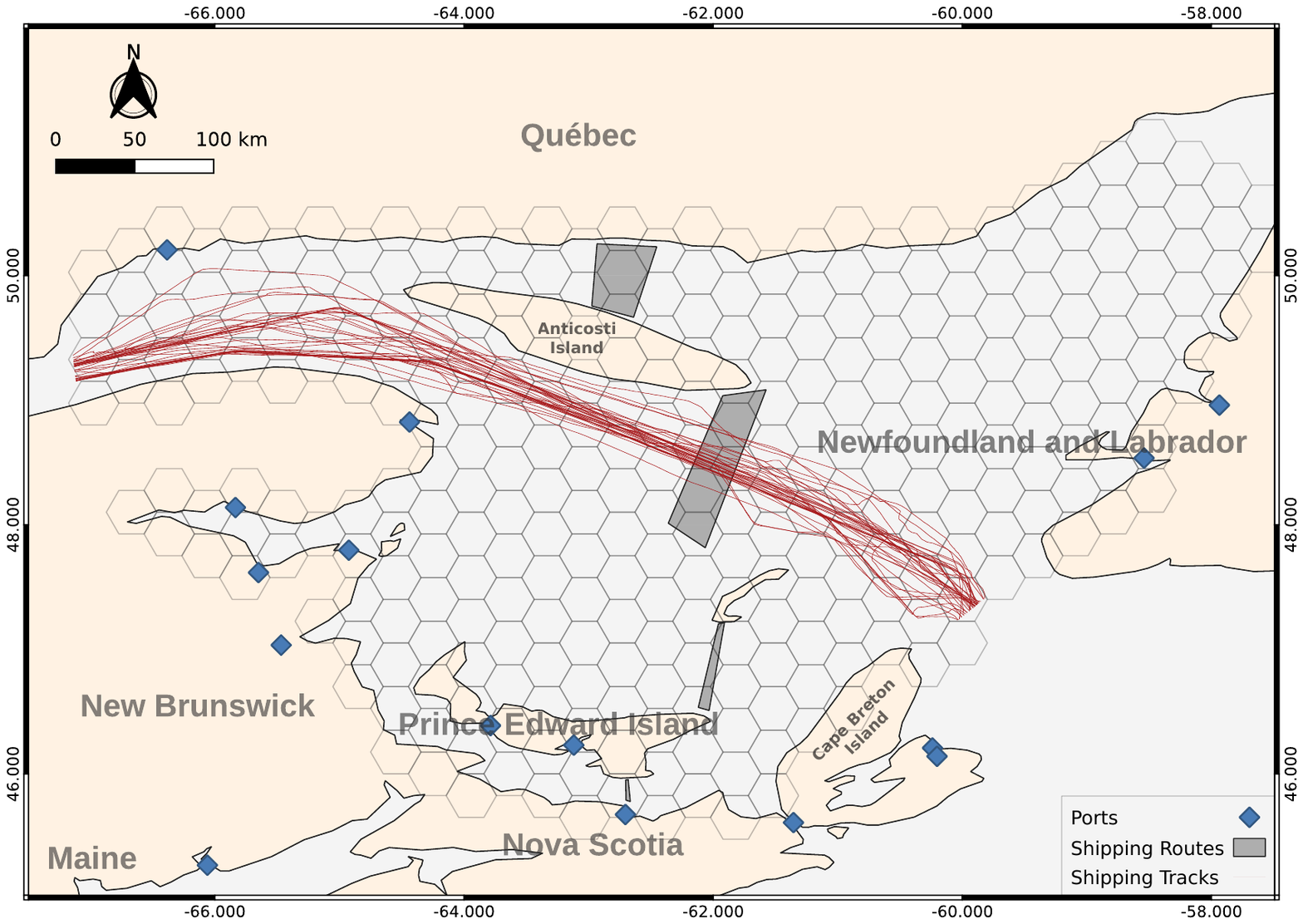}
        \caption{Path 1}
        \label{fig:1}
    \end{subfigure}
    \begin{subfigure}[b]{.475\textwidth}\centering
        \includegraphics[width=\textwidth]{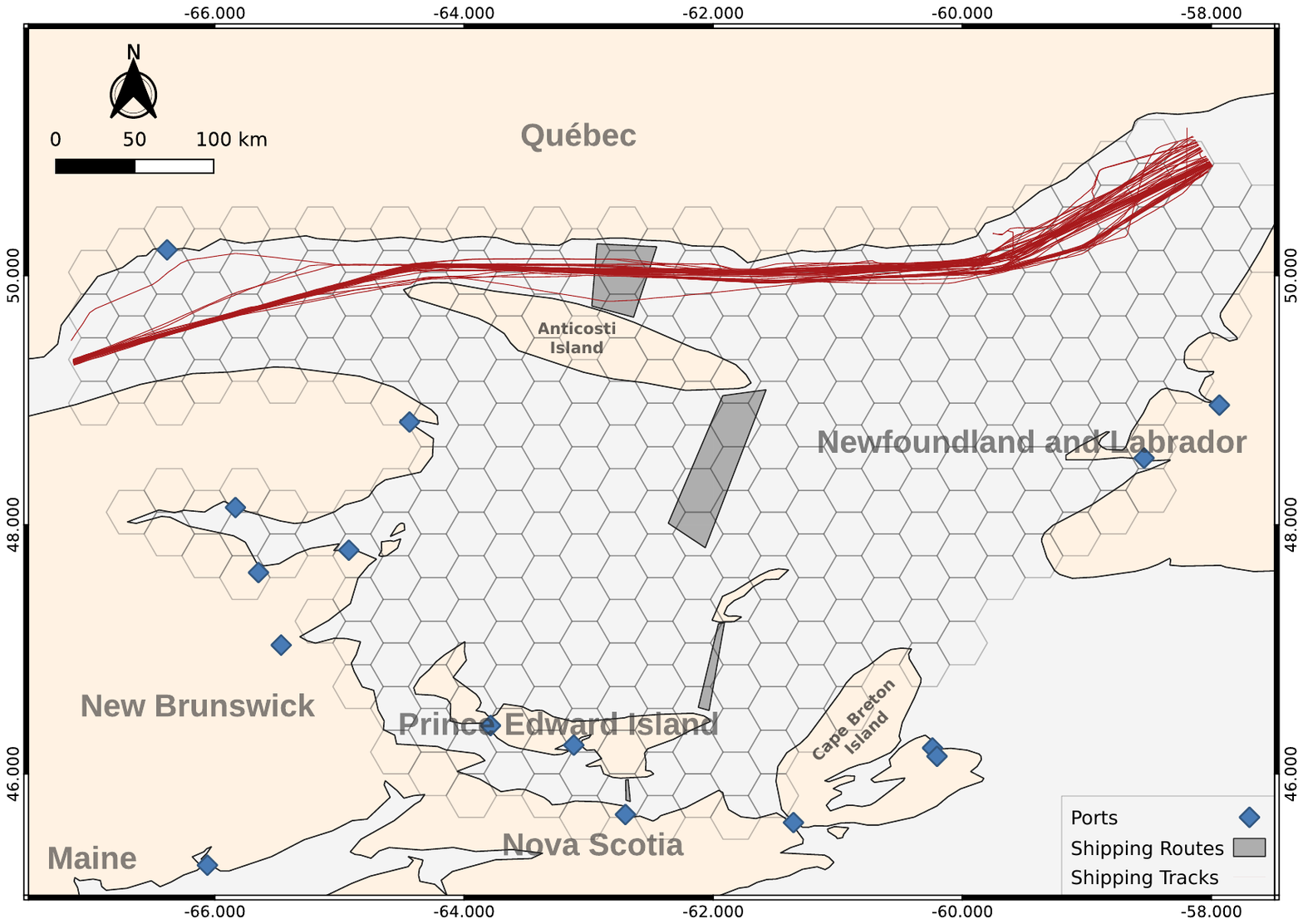}
        \caption{Path 2}
        \label{fig:2}
    \end{subfigure}
    \vspace*{\floatsep}
    \begin{subfigure}[b]{.475\textwidth}\centering
        \includegraphics[width=\textwidth]{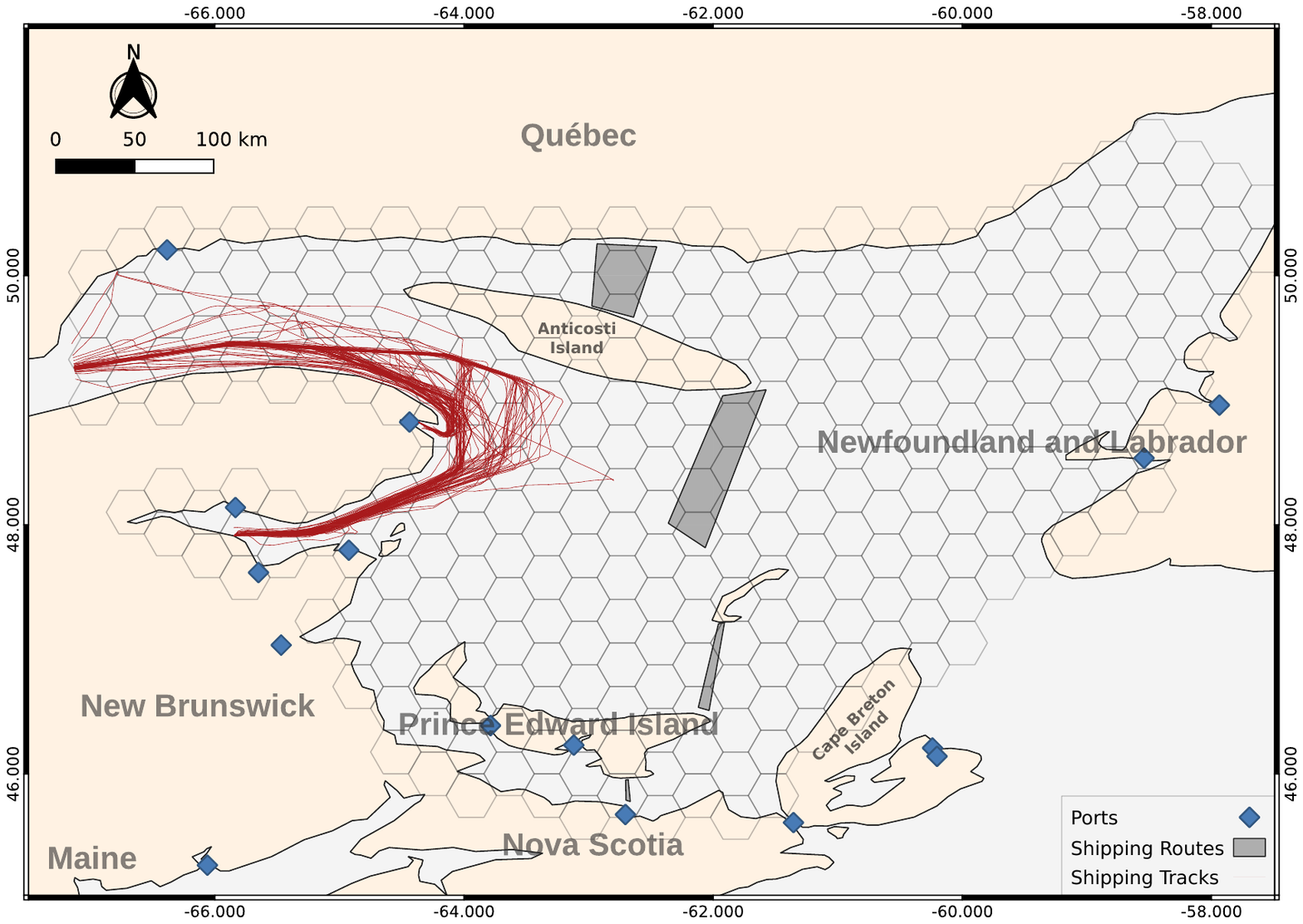}
        \caption{Path 3}
        \label{fig:3}
    \end{subfigure}
    \begin{subfigure}[b]{.475\textwidth}\centering
        \includegraphics[width=\textwidth]{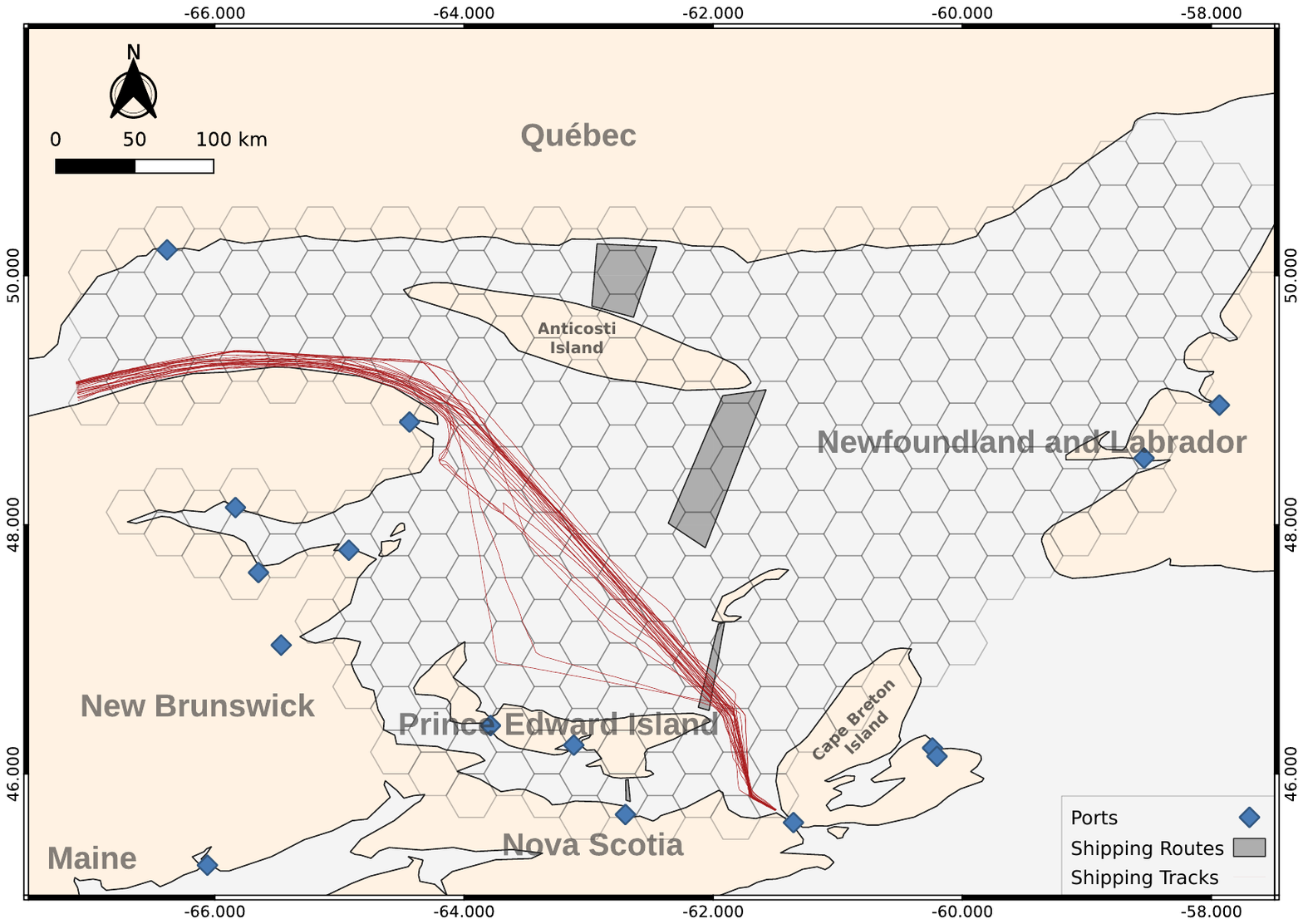}
        \caption{Path 4}
        \label{fig:4}
    \end{subfigure}
    \caption{The Gulf of St. Lawrence region has multiple paths with high vessel traffic. This figure shows four of them, which do not represent our dataset's complete set of trajectories but cover three major ones (1, 2, and 4) that cross a route polygon. Path 3, on the other hand, does not intersect with any route polygon. Instead, it represents tracks that interact with all major hotspots of NARW in the gulf.}
    \label{fig:all-paths}
\end{figure}

Accordingly, in the Gulf of St. Lawrence, ships can take various paths even if their start and end are known beforehand (see Figure~\ref{fig:vessel-decision}).
This is because no physical grid directs ships into specific lanes, unlike city streets~\cite{SPADON201718, 10.1007/978-3-319-93698-7_21, SPADON2019209}.
Although traffic management policies designate shipping lanes, the lanes do not restrict the movement of vessels as they may enter or exit lanes at different points throughout the gulf.
Predicting ship trajectories becomes increasingly challenging for longer distances as the predicted paths deviate significantly from actual paths with increased elapsed time and travel distance.
In simpler terms, the further a ship travels, the larger the uncertainty of its future position, making it difficult to predict accurately.
To overcome this challenge, we need a strategy that provides increased information for the forecasting task.
This can be achieved by fusing information about a given vessel's possible route and destination, which changes the forecasting paradigm into reconstruction while reducing the uncertainty of the vessel's future positioning.
However, such information is unavailable within the AIS data and would require a modeling strategy to provide trustworthy information about unseen and unknown trajectories in the ocean.

To solve this problem, we first split the gulf into hexagonal grid cells (see Figure~\ref{fig:NARW-Paths}) of $0.3^\mathrm{o}$ on EPSG:4269\footnote{~https://epsg.io/4269.} projection.
The shape of the cells is due to the Earth's curvature, as hexagonal cells can divide the Earth's surface uniformly.
On top of such grid, Figure~\ref{fig:all-paths} shows some paths discovered from historical data, highlighted by polygons placed in high-traffic areas.
Vessels crossing these polygons are more likely to interact with a whale hotspot area based on the historical movement behaviors in the same region.
In this sense, comparing trajectories in high-traffic areas, we can observe that most tracks intersect with route polygons at some point during their course.
Besides, paths like the one in Figure~\ref{fig:all-paths}{\bf C} are the intricate patterns of the dataset as they do not intersect with any route polygons and represent curvy-shaped moving patterns.
Based on all these different moving behaviors, we developed a probabilistic model (see Algorithm~\ref{algo:pdestination}) that leverages the relationship between grid cells, route polygons, and historical vessel tracks to predict the route taken by a vessel and the destination ({\it i.e.}, endpoint ) of that same voyage.

\begin{figure}[!b]
    \centering
    \hfill
    \begin{subfigure}[b]{.49\textwidth}
        \centering
        \includegraphics[width=\textwidth]{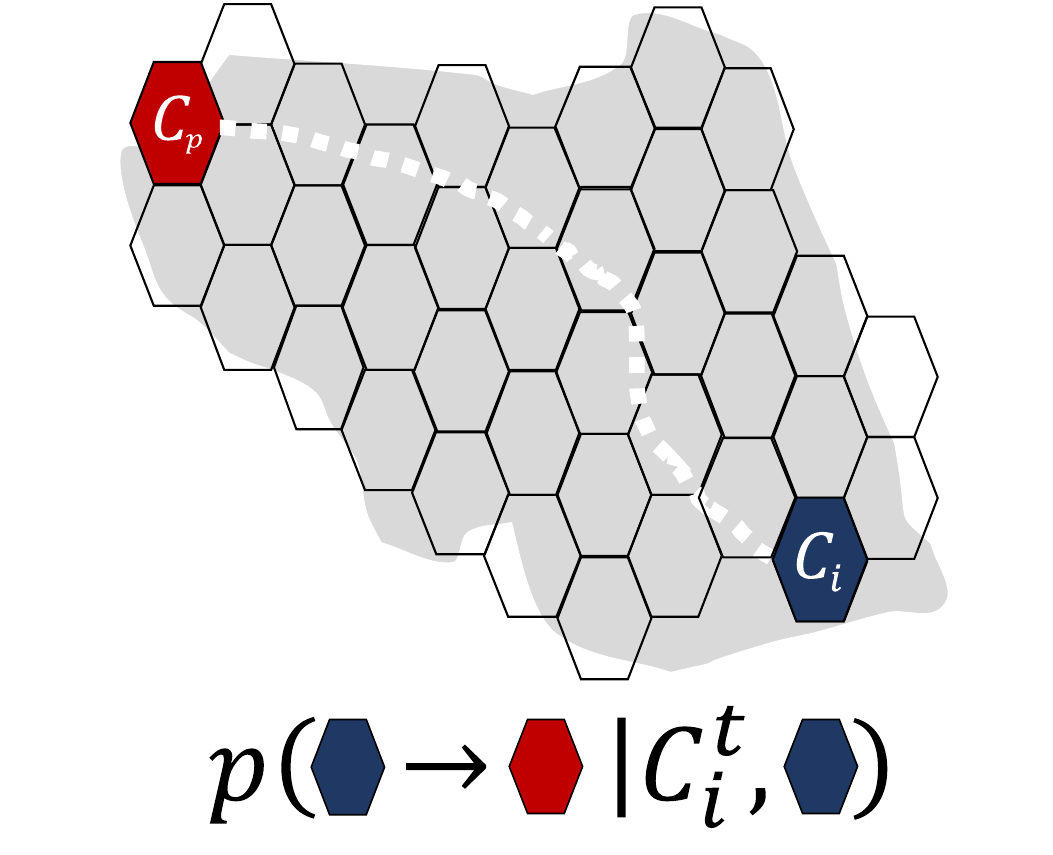}
        \caption{Traditional Formulation}
        \label{fig:trad}
    \end{subfigure}
    \begin{subfigure}[b]{.49\textwidth}
        \centering
        \includegraphics[width=\textwidth]{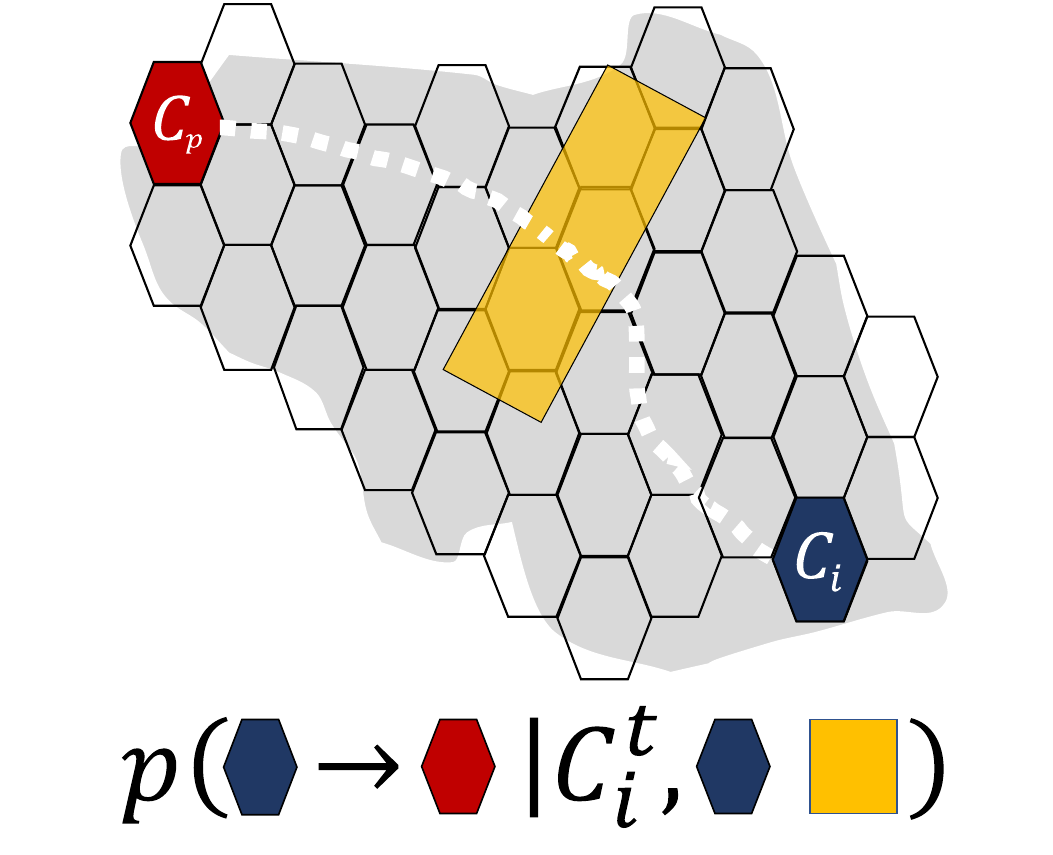}
        \caption{Proposed Formulation}
        \label{fig:ours}
    \end{subfigure}
    \hfill
    \caption{Probabilistic formulations of long-term trajectory forecasting. In the traditional method (a), the probabilistic matrix is utilized to determine the most likely whereabouts of a vessel by analyzing its movements concerning its starting point ($C_{i}$) on a grid using historical data. Our proposed approach (b) follows a similar method but with new features and a revisited understanding. In addition to historical data, the potential route depends on traversing the polygons. Essentially, our model employs route polygons as a type of memory to provide more precise guidance on the vessel's decisions at different moments in the past, aiming for a more reliable and effective probabilistic-based inference system.}
    \label{fig:formulations}
\end{figure}

Understanding the historical movements and patterns of vessels is essential in this approach.
The proposed probabilistic model, as shown in Figure~\ref{fig:formulations}{\bf B}, considers the relationship of tracks with route polygons and grid cells that highlight major paths in the gulf.
By calculating the (1) number of tracks going to a particular destination (a grid cell), (2) the direction of each track at the point where it intersects with a route polygon, and (3) the probability score of each surrounding cell as a potential destination for each track, we can gain valuable insights into ship behavior and the routes they take.
The initial step of this approach ({\it i.e.}, probabilistic model) is to predict the probable destination (grid cell as endpoint) while considering the vessel's current position.
This is achieved by computing the difference, through the Euclidean distance (see Equation \ref{eq:Euclidean}), between the motion behavior of the vessel in the grid cell where it currently lies and the historical motion behavior of vessels for each possible destination.
We consider grid information and AIS messages to determine probability scores and potential destinations.

In the following, we provide a detailed description of the methodology we use to achieve the proposed solution, and for simplicity's sake, we refer to forecasting and reconstruction indistinctly.
Furthermore, for the sake of disambiguation, this research proposes two models: (1) a \textit{probabilistic model} for predicting end-point of a track, and (2) \textit{deep-learning model} to reconstruct the trajectories while using the output of probabilistic model as embedded features.
Such models worked on a phased framework, contributing to the building of a feasible solution for the same problem.

\subsection*{Methodological Framework}

\begin{figure}[!b]
    \centering
    \includegraphics[width=\linewidth]{"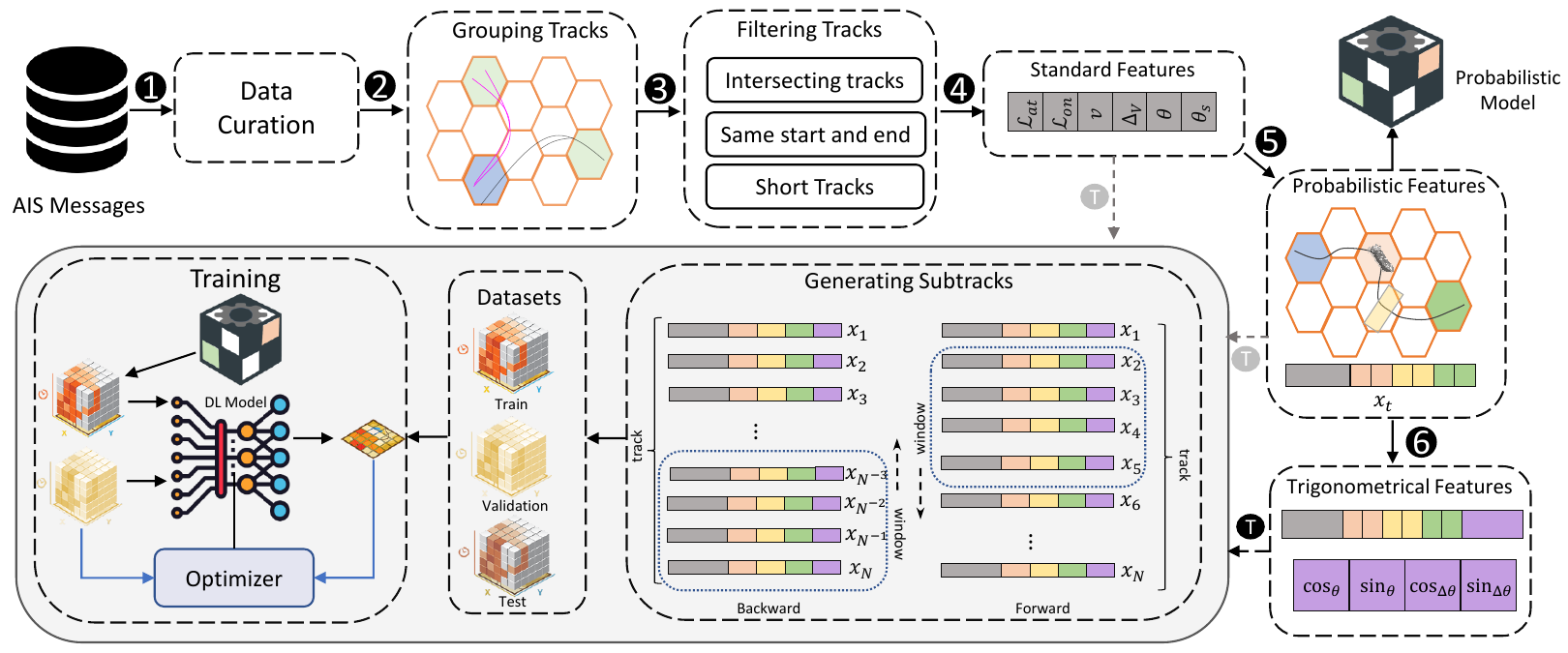"}
    \caption{Visual representation of our pipeline. We start with a subset of AIS messages that cover the Gulf of St. Lawrence. After curating the data and separating messages by MMSI into tracks {\bf (Step~1)}, we then create shorter segments (tracks) with end-to-end trip behavior and group these tracks based on common start and end points {\bf (Step~2)}. This means the dataset is divided into subsets with a balanced number of trajectories based on their unique patterns, and then the samples are weighted according to their population. All short tracks, self-intersecting tracks, and tracks having the same start and end port are removed {\bf (Step~3)}. We reverse the tracks and add them to the dataset as new tracks. Using positional attributes, bearing, speed, and their rate of change is calculated to form the {\it Standard Features} set {\bf (Step~4)}. This features set is further extended by {\it Probabilistic Features}, which are current grid cell (orange color), route polygon (yellow color), and last grid cell (green color) centroids, and for that, we consider the start and end of each track and the closest route polygon placed in the gulf {\bf (Step~5)}. These new features are used to compute the probabilistic model. We then add {\it Trigonometric Features} for each message of tracks to extend the existing feature set {\bf (Step~6)}. In {\bf Step~T}, for each track, using the sliding window technique, we generate sub-sequences (trajectories) in contiguous temporal slices of up to 15 hours to construct training, validation, and testing datasets. The first three hours serve as the input to the deep learning model, and the following 12 hours are the observed data to compare to the model prediction. After applying a temporal mini-batch training method, we optimize the neural network to identify the best subset of parameters that can closely forecast the samples in the training set. When training with probabilistic features, each sample is updated to have the additional engineered features from the probabilistic model before feeding to the neural network. This training process is repeated several times until there is no further improvement when testing the model on the validation-reserved data, and after training, we evaluate it on the test data.}
    \label{fig:pipeline}
\end{figure}

In our trajectory forecasting pipeline, as depicted in Figure~\ref{fig:pipeline}, we extract MMSI, timestamps, vessel type, and coordinates from AIS messages recorded between 2015 and 2020.
Although over ten types of vessels are included in the data, however, our study only focuses on cargo and tanker vessels due to their distinguishable quasi-linear movement behavior {\bf (Step~1)}.
Our pipeline uses tracks, which are ordered sequences of a vessel's positions from a source port to a destination, instead of complete voyages.
Creating tracks from AIS messages involves segmenting voyages based on the time and distance between consecutive messages to identify sub-sequences with end-to-end trip behavior.
In our study, a new track starts if the interval between two consecutive AIS messages exceeds 8 hours or the distance of 50 Km or change in the bearing between two consecutive points varies by more than $\pm 45^{g}$. It is obvious in the ocean that some routes are more busy than others thus each route has its own density of vessels. 
We weigh the tracks based on their starting and ending grid, assigning a weight value corresponding to the vessel density of that moving pattern with respect to the entire dataset.
Such weight is used for stratified dataset sampling, assuring that distinct mobility patterns will be equally used for model training and validation {\bf (Step~2)}.

In {\bf Step~3} of our data processing pipeline, we discard messages within one kilometer of the nearest port as vessels are likely anchored in these areas.
Additionally, we discard self-intersecting tracks as well as tracks that start and end at the same port, often observed in fishing vessels.
After this filtering, we interpolate points of each track with uniform interval of 10 minutes between two consecutive points.
Next, we remove tracks that are outliers from the dataset.
Specifically, we identify outlier tracks within a grid cell where the number of tracks of a particular stratification pattern is less than or equal to five.
That is the case of tracks of vessels patrolling around marine protected areas that may have been mislabeled during data collection and thus need to be removed.
The resulting dataset consists of vessels grouped by their MMSI and divided by their type, with sets of tracks containing only coordinates.
Finally, in the preprocessing stage, we reverse the tracks and add them to the dataset as new tracks to increase the sample size for the experiment.
In step {\bf Step~4}, we calculate the speed displacement of the vessel in knots and its acceleration in knots per hour based on the distance between the consecutive AIS messages.
We do the same to compute the bearing of the vessel and the rate of the change of the bearing in gradian (see Appendix).
The resulting information is called the {\it Standard Features} (i.e., longitude, latitude, speed, bearing, change in speed, change in bearing) of each track.

Afterward, we split the dataset into three subsets: training, validation, and testing.
The division is based on vessel type and MMSI, following the stratified sampling weights defined earlier.
We use the training set to construct the probabilistic model.
This model is capable of identifying common routes and predicting their probable destinations.
The presence of multiple routes between frequently traveled source and destination ports highlights the need to create route polygons to direct the forecasting models.
In this sense, the set of {\it Standard Features} of each track is extended with the current grid cell, route polygon, and last grid cell centroids {\bf (Step~5)}.
We refer to this extended feature set as {\it Probabilistic Features}.
Subsequently to this step, in {\bf Step 6}, we further extend the {\it Probabilistic Features} by calculating trigonometrical transformation on top of the vessels' location and bearings resulting in the {\it Trigonometrical Features}.
Either of the three feature sets can be used for training the deep learning model in \textbf{Step T}, but not multiple simultaneously.
For example, in Figure \ref{fig:pipeline}, the {\it Trigonometrical Features} is the input of the deep learning model, whereas arrows from the other feature sets are greyed out.

The target of our approach is to forecast long term trajectory up to 12 hours by feeding 3 hours trajectory to the model. The tracks of the vessels are usually in days so smaller sequences (sub-tracks) are needed. To generate smaller sequences of tracks for the deep learning model, we generate {\it trajectories} that are up to 15 hours long using a sliding window algorithm.
This algorithm processes 3 new messages simultaneously, iterating over a track every 30 minutes.
This aims to ensure that the input data is shaped into vectors of equal size, ranging from 1 to 3 hours, by padding any missing values.
The output vectors always have a size of 12 hours, and no padding is applied to them.
The deep learning model uses the first 3 hours of the output vector as input, while the remaining 12 hours are observed data used to evaluate the model's prediction.
We proceed to neural network engineering and training after generating the input features normalized between $[0,1]$.
Regardless of the feature set we use, we expect the model to generate a pair of normalized coordinates in the range of $[0,1]$ as output, which, when denormalized, should resemble the anticipated output of the vessel trajectory.
It is important to mention that probabilistic model is learned with tracks in training dataset whereas deep learning model is trainined over the trajectories (sub-tracks or subsequences). 
The training of deep learning model continues until no improvement is seen on the validation-reserved dataset slice, after which the models are evaluated on the test dataset.
The training and validation are conducted simultaneously on tankers and cargo vessels.
However, the final testing is performed separately so that we can account for the individual performance by vessel type.
We also allow the last element in the training input vector to be present as the first element of the output vector (overlap of 10 minutes).

Our experiments involve testing different sets of features and network architectures.
Our contributions are on feature engineering and neural network modeling, which together can make complex long-term path decisions.
We conducted experiments with three distinct sets of features and displayed the results of other neural network models in an ablation fashion.
We covered comparisons with the literature and state-of-art, including simple RNNs~\cite{doi:10.1080/20464177.2019.1665258, s20185133, s21217169}, Bidirectional RNNs~\cite{s18124211, jmse10060804, 9721877}, CNNs/RNNs AutoEncoders~\cite{MURRAY2021107819, 9768818, 9626839}, and combinations of those with and without the transformer-based attention mechanism, all of which have less than 1.5M trainable parameters.
We chose not to test models that exceeded the threshold due to their large size and long training time requirements.
As AIS is a type of streaming data, it requires models that can train and produce fast outputs.
Our proposal suggests using smarter features in combination with simpler, well-tailored models.
Additionally, in the Gulf of St. Lawrence context, there is not enough data and information available to justify using larger and deeper models.
These models may not necessarily provide better results but will require more resources and take longer training sessions.

Differently from the literature, our model uses parallel Convolutional Neural Networks (CNNs) to extract local and global features from input data through dilated convolutions.
The CNNs' outputs are combined and processed by a Bidirectional Long Short-Term Memory (Bi-LSTM) with a Position-Aware Attention mechanism, efficiently encoding the temporal sequence, giving higher importance to later timesteps.
The resulting latent space is then decoded through a second Bi-LSTM and multiple Dense layers, giving the final output as normalized latitude and longitude values.
This end-to-end, customizable architecture achieves superior performance in predictive geospatial tasks, mainly when trained with probabilistic and trigonometrical features.
To start this discussion, subsequently, we formalize our probabilistic model while using Table~\ref{tab:notation-1} to describe the notations used from hereinafter.

\begin{table}[!h]
    \centering
    \caption{Symbols and notations relevant for the probabilistic model formulation understanding.}
        \begin{tabularx}{\textwidth}{c X}
            \toprule
            Notation & Definition \\
            \midrule
            $\tau$ &  An ordered sequence of $t$ points \\
            $x_{t}$ & A point in a $\tau$ at instant $t$, in format $\left<\mathcal{L}_{on}, \mathcal{L}_{at}, S, \Delta_{v}, \theta, \Delta_{\theta} \right>$ \\
            $\mathcal{L}_{on}$ & Longitude in degrees \\
            $\mathcal{L}_{at}$ & Latitude in degrees \\
            $v$ & Speed in knots \\
            $\Delta_{v}$ & The acceleration in knots per unit time of the vessel \\
            $\theta$ & Bearing in gradian \\
            $\Delta_{\theta}$ & The rate of change of the bearing in gradian per unit time \\
            $\mathbf{M}$ & Route probability matrix $\mathbf{M} \in \mathbb{R}^{c \times r \times z}$, where $c$ is the number of grid cells, $r$ is the number of route polygons and $z$ is number of tracks from $c$ passing through $r$ \\
            $\widetilde{\mathbf{M}}$ & Destination probability matrix $\widetilde{\mathbf{M}}  \in \mathbb{R}^{c \times c \times z}$, where $c$ is the number of grid cells and $z$ is number of tracks passing through a given grid cell \\
            $C$ & Set of all grid cells $C=\left\{ C_{i} \right\}^{c}_{i=1}$\\
            $D$ & Subset of destination grid cells $D=\left\{ D_{j} \right\}^{d}_{j=1}$ where $D \subseteq C$ \\
            $T$ & Collection of tracks $T=\left\{ \tau_{i} \right\}^{n}_{i=1}$, where $n$ is the number of tracks \\
            $T_{C_{i}}$ & Subset of tracks that have passed through a specific grid cell $C_{i}$ \\
            $T_{D_{j}}$ & Set of tracks that ended up in $D_{j}$ \\
            $S_{\tau}$ & A tuple of motion statistics of track $\tau$ in format $\{\mathcal{L}_f, \mathcal{L}_e, \widetilde{\theta}_g, \Psi_{\theta_g}\}$ \\
            $\mathcal{L}_f$ & First AIS coordinates of track $\tau_i$ within a grid cell $C_i$ \\
            $\mathcal{L}_e$ & Last AIS coordinates of track $\tau_i$ within a grid cell $C_i$ \\
            $\widetilde{\theta}_g$ & The median of the course values of vessel movements in a grid cell \\
            $\Psi_{\theta_g}$ & The Gaussian Kernel Density Estimation-based entropy of the bearing values \\
            $\delta$ & Euclidean distance threshold for motion statistics, first quantile of all pre-computed distances \\
            $\mathbf{I}(\cdot)$ & Indicator function, outputs 1 if a $\tau$ satisfies the input conditions, else 0 \\
            $E$ & Euclidean distance function, as defined in Equation~\ref{eq:Euclidean} \\
            $\mathcal{L}^{R}$ & Predicted route centroid coordinates, where $R \subseteq C$ \\
            $\mathcal{L}^{N}$ & Current grid cell's centroid coordinates, where $N \subseteq C$ \\
            $\mathcal{L}^{D}$ & Predicted destination grid cell's centroid coordinates, where $D \subseteq C$ \\
            \bottomrule
        \end{tabularx}%
    \label{tab:notation-1}%
\end{table}

\section{Probabilistic Feature Model}
\label{sec:proba-model}

A trajectory is a sequence of points representing the location of an object over time, and, in this paper, a trajectory $\tau$ is represented as an ordered sequence of $t$ points in time, such as $\tau = \left<x_{1}, x_{2}, \ldots, x_{t}\right>$, where a point $x_{t} \in \tau$ at time $t$ is also an ordered sequence of features from the AIS message such as $x_{t} = \left<\mathcal{L}_{on}, \mathcal{L}_{at}, v, \Delta_{v}, \theta, \Delta_{\theta} \right>$; where $\mathcal{L}_{on}$ is the longitude, $\mathcal{L}_{at}$ the latitude, $v$ the speed in knots, $\theta$ the bearing in gradian, $\Delta_{v}$ the acceleration in knots per hour, and $\Delta_{\theta}$ the rate of change of the bearing in gradian.
We refer to the set of values in $x_{t}$ of each AIS message as \textit{Standard Features}.
The speed, bearing, and deltas can be mathematically calculated using sequences of coordinates (see Annex for details).

Our approach involves probabilistic modeling where we consider a 310-cell grid that overlays the Gulf of St. Lawrence and six-year historical AIS data to create a Route Probability Matrix $\mathbf{M} \in \mathbb{R}^{c \times r \times z}$ and a Destination Probability Matrix $\widetilde{\mathbf{M}} \in \mathbb{R}^{c \times c \times z}$.
Here, $c$ represents the number of cells, $r$ is the number of possible routes (route polygons), and $z$ is the number of tracks passing through a grid cell or route polygon.
Through those, we aim to generate a probable route and destination that can be used as features for deep-learning models to forecast vessels' trajectories.
It is important to note that our probabilistic model does not predict the next grid cell of the vessel during the vessel displacement in the trajectory.
Instead, it predicts the probable long-term decisions that vessels will make.
In the end, we will have a collection of conditional probabilities related to trajectories $T$, grid cells $C$, and destination grid cells $D$.
We use $T_{C_{i}}$ to refer to a subset of tracks that have passed through a particular grid cell $C_{i}$ and $T_{D_{j}}$ to represent the set of tracks that ended up in $D_{j}$, $i \neq j$.
For a simple conditional probability formulation where $z$, the individual tracks interacting with the grid cells, is not taken into account, one could populate $\mathbf{M}_{ij}$ with the conditional probability of tracks crossing a route polygon $R_{k}$ given it crossed the grid cell $C_{i}$ as $\mathbb{P}(R_k | C_{i}) = |T_{R_k} \cap T_{C_{i}}| \div |T_{C_{i}}|$ and populate $\widetilde{\mathbf{M}}_{ij}$ with the conditional probability of a trajectory crossing a grid cell $D_{j}$ given it crossed the grid cell $C_{i}$ as $\mathbb{P}(D_{j} | C_{i}) = |T_{D_{j}} \cap T_{C_{i}}| \div |T_{C_{i}}|$.

\begin{figure}[!ht]
    \centering
    \begin{adjustbox}{width=11cm, trim=1cm 1cm 3cm 1.5cm, clip, frame}
        \includegraphics{"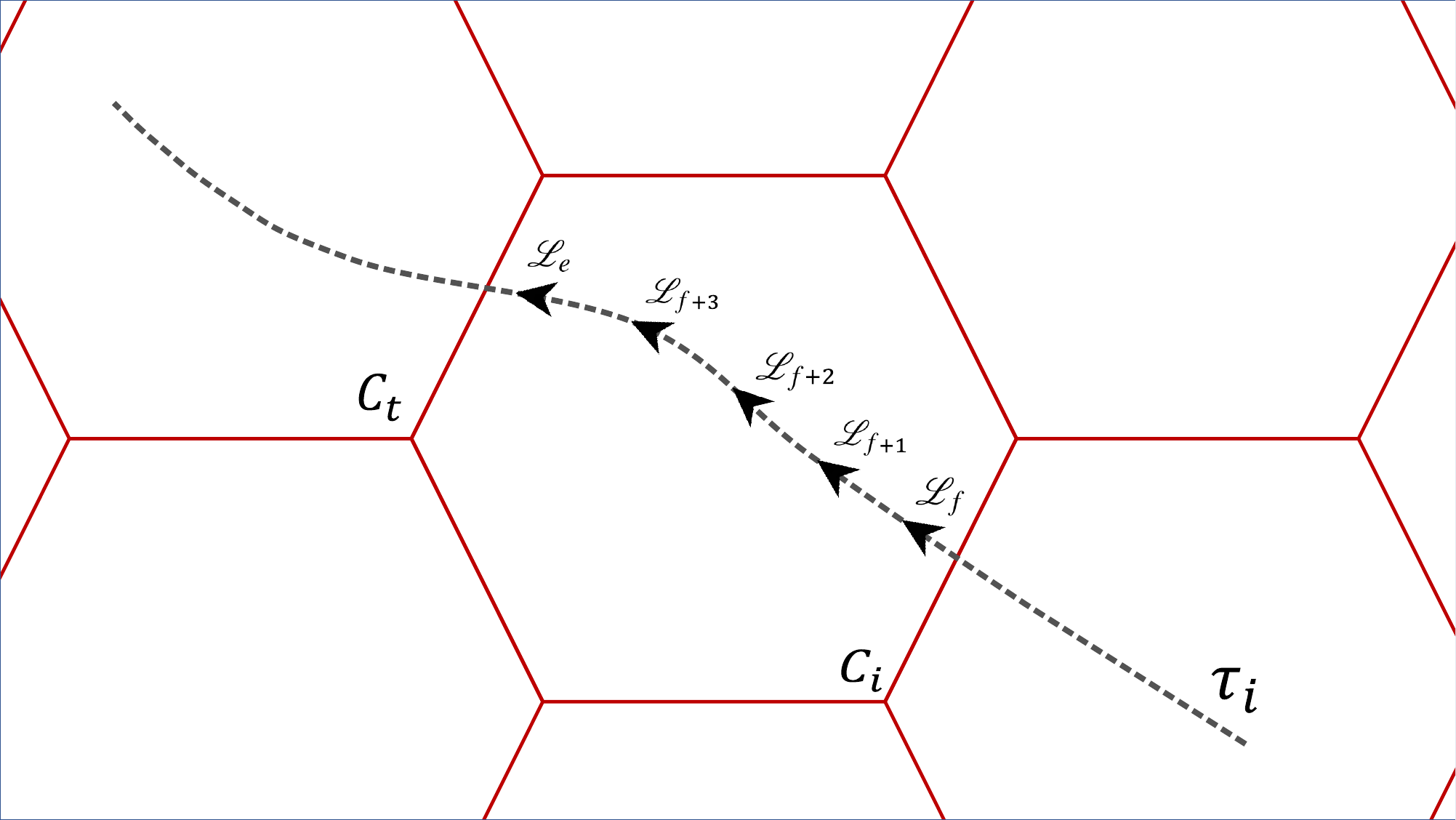"}
    \end{adjustbox}
    \caption{An example of AIS coordinates $\mathcal{L}_{f}$ first and $\mathcal{L}_{e}$ last of a vessel inside a grid cell $C_{i}$ (current) and $C_{t}$ (subsequent) on the vessel trajectory $\tau_{i}$.}
    \label{fig:pingincell}
\end{figure}

Unfortunately, conditional probabilities alone cannot accurately describe the transition between grid cells.
This would lead to a bias towards shipping routes with higher vessel traffic over time.
Moreover, assuming an equal distance between all grid cells neglects the importance of the vessel course and its distance (from the last AIS message location) to the nearest route polygon.
Because of that, we consider the relationship of the vessel location with the route polygon.
We do so by accounting for the trajectory $\tau_{i}$ movement direction while passing through $C_i$, as depicted in Figure~\ref{fig:pingincell}.
Such directional information ($S_{\tau_{i}}$) is stored on the last axis of our probability matrices $\mathbf{M}~/~\widetilde{\mathbf{M}}$, for each destination grid cell and route polygon, respectively.
We refer to the motion direction ($S_{\tau_{i}}$)  as \textit{motion statistics}.
The motion statistics for each track $\tau_i \in T_{C_i}$ within a particular grid cell consist of a four-value tuple, which is represented by $S_{\tau_i} = \{\mathcal{L}_f, \mathcal{L}_e, \widetilde{\theta}_g, \Psi_{\theta_g}\}$.
Within a grid cell $C_i$, $\mathcal{L}_f$ and $\mathcal{L}_e$ represent the first and last pair of coordinates, respectively, of $\tau_i$ --- {\it e.g.}, $\mathcal{L}_f = \Big<\mathcal{L}_{lon}^{f}, \mathcal{L}_{lat}^{f}\Big>$; $\widetilde{\theta}_g$ is the median of the course values of messages inside $C_i$; and, $\Psi_{\theta_g}$ is the Gaussian Kernel Density Estimation-based entropy of the course values of messages inside $C_{i}$. A hexagonal grid cell has geometrically six neighbors, and $\Psi_{\theta_g}$ is responsible for capturing the movement and change in the vessel's direction inside a particular grid cell.  Each $S_{\tau_i}$ of historical track is associated with a route polygon $S_{\tau_i} \rightarrow R_{k}$ and a destination $S_{\tau_i} \rightarrow D_{\tau_i}$, in which, motion statistic for each track $\tau \in T$ is computed.
Our aim here is to define feasible transitions between routes' and destinations' grid cells.

The similarity of a new trajectory $\tau_{n}$ and a historical track $\tau_{i}$ in a given grid cell $C_{i}$ is computed using Euclidean distance of their motion statistics $S^{C_{i}}_{\tau_{n}}$ and $S^{C_{i}}_{\tau_{i}}$ in $C_{i}$ as follows:
\begin{equation}
    \hfill
    E\left(S^{C_{i}}_{\tau_{n}}, S^{C_{i}}_{\tau_{i}}\right) = \sqrt{\left(\mathcal{L}_f^{\tau_{n}} - \mathcal{L}_f^{\tau_{i}}\right)^2 +  \left(\mathcal{L}_e^{\tau_{n}} - \mathcal{L}_e^{\tau_{i}}\right)^2 + \left(\widetilde{\theta}_g^{\tau_{n}} - \widetilde{\theta}_g^{\tau_{i}}\right)^2 + \left(\Psi_{\theta_g}^{\tau_{n}} - \Psi_{\theta_g}^{\tau_{i}}\right)^2}
    \hfill
    \label{eq:Euclidean}
\end{equation}

\noindent
where $\mathcal{L}_f$ and $\mathcal{L}_e$ represent starting and ending coordinates; and, $\widetilde{\theta}_g$ and $\Psi_{\theta_g}$ refer to the median and entropy of bearing values, respectively.
The distance is calculated between the current track $\tau_{n}$ and each historical track $\tau_{i}$ inside grid cell $C_i$.
In this calculation, every value in $S_{\tau}$ is normalized between the range $[0,1[$, which ensures that all features have equal influence on the distance measure.
The lower the Euclidean distance, the more likely a vessel will take that route or reach its more probable destination.

\begin{figure}[!t]
    \centering
    \begin{subfigure}[b]{0.49\textwidth}
        \includegraphics[width=\textwidth]{"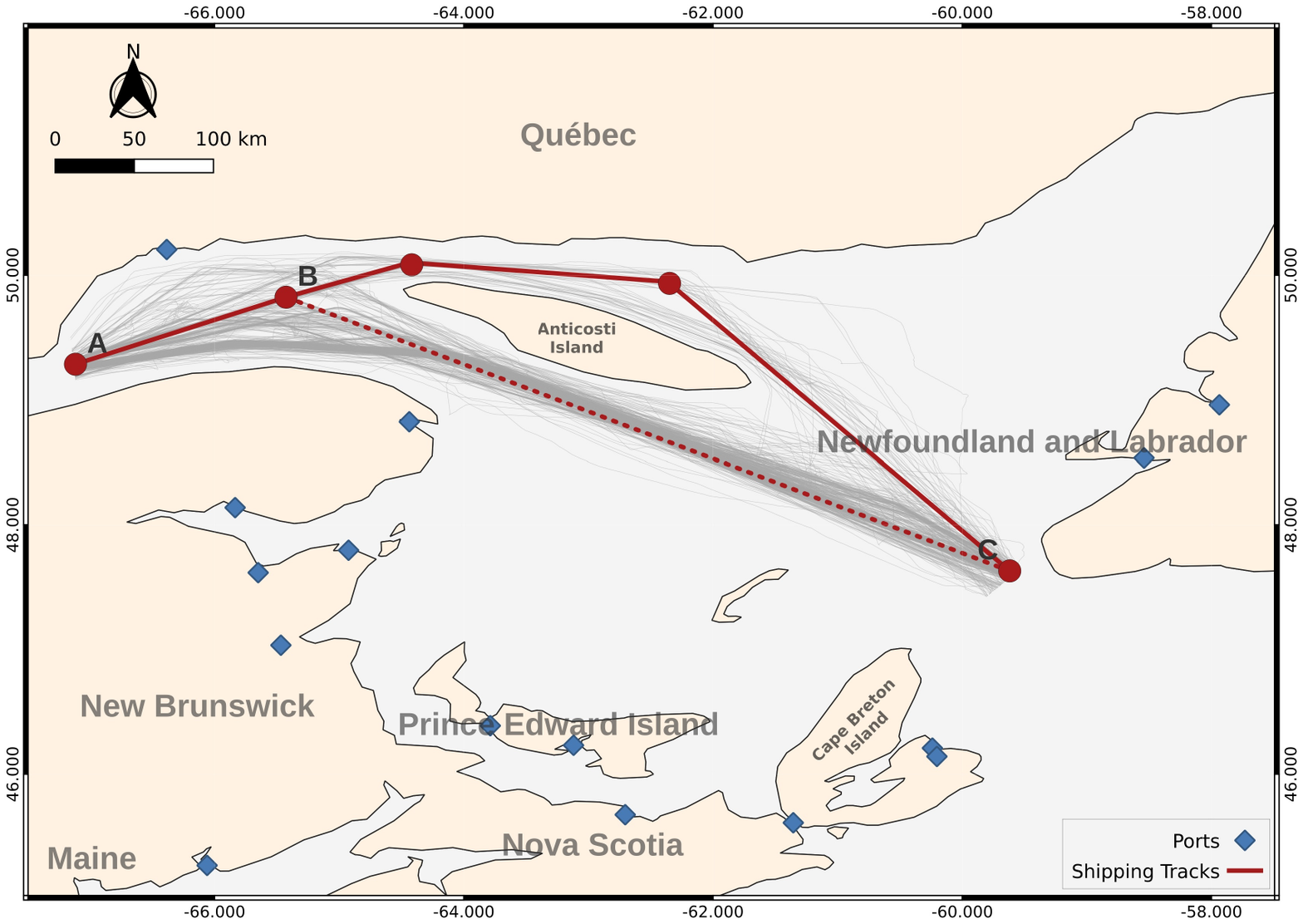"}
        \caption{}\label{fig:twoways}
        \vspace{.5mm}
    \end{subfigure}
    \begin{subfigure}[b]{0.495\textwidth}
        \includegraphics[width=\textwidth]{"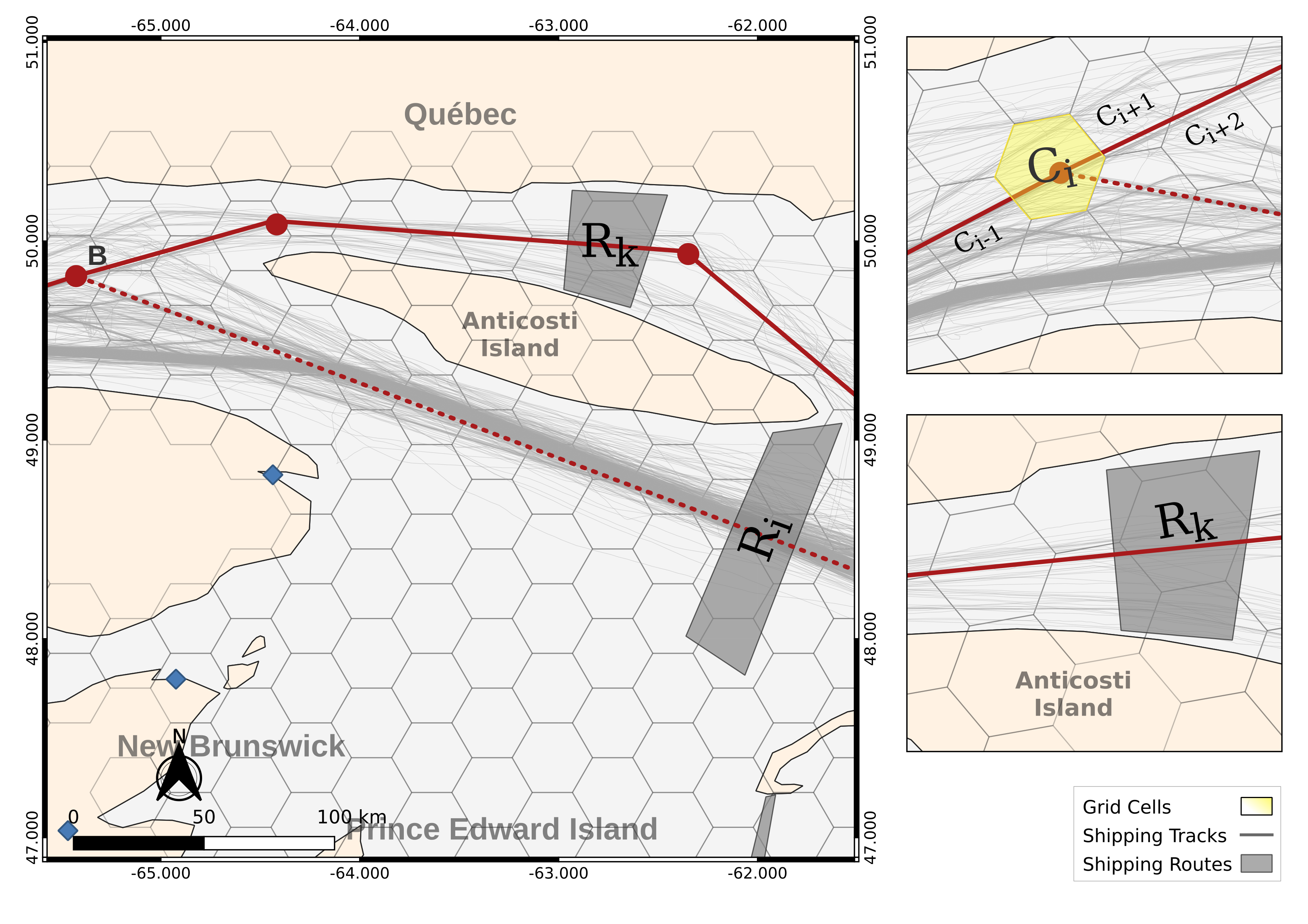"}
        \caption{}\label{fig:gridpolygon}
        \vfill
    \end{subfigure}
    \caption{The image {\bf (a)} displays a density map depicting two routes from a starting point A towards the endpoint C. In image {\bf (b)}, the region has been divided into hexagonal grid cells of 0.3$^\mathrm{o}$ using {\it ESPG:4269} projection, where a grid cells are referred to as $C_{i}$ and possible routes as $R_{i}$ and $R_{k}$. The solid red line shows the probable movement of a vessel between B and C, and the dashed line shows a less probable one.}
    \label{fig:path_grid_prob}
\end{figure}

As an example of the probabilistic process, Figure~\ref{fig:path_grid_prob} depicts a vessel moving from point A to C, with a decision point at the segment between point B and C.
At this decision point, there are two possible paths, each of which interacts with a different route polygon ($R_i$ and $R_k$).
The solid red line represents the most probable path, while the dashed line shows the least probable path.
To define a probable route of a vessel, we need to consider the location of the route polygon ($R_k$ or $R_i$), the vessel's location ($C_i$) during track $\tau_{n}$, and the motion statistics.
We define the probability function for choosing a route as follows:
\begin{equation}
    \hfill
    \mathbb{P}(R_{k} | C_{i}, \tau_{n}) = \frac{ \sum_{\tau_{i} \in T} \mathbf{I} \Big( x_{t} \in \tau_{i} \wedge \exists x_{t} \in R_{k} \, \wedge\, E(S_{\tau_n}, S_{\tau_{i}}) \leq \delta \, \wedge\, \exists x_{t} \in C_{i} \Big) }
    { \sum_{\tau_{i} \in T} \mathbf{I}\Big( x_{t} \in \tau_{i} \, \wedge\, E(S_{\tau_n}, S_{\tau_{i}}) \leq \delta \, \wedge\, \exists x_{t} \in C_{i}\Big) }
    \hfill
    \label{eq:route}
\end{equation}

\noindent
where $\delta$ is a threshold representing the first quantile value of all pre-computed distances between historical motion statistics as defined in Algorithm \ref{algo:pdestination}, explained in the Methods section.
Additionally, the probability for the destination $D_{j}$ is formalized with respect to the route polygon $R_k$, using the following equation:
\begin{equation}
    \hspace{-1.5cm}
    \mathbb{P}(D_{j} | C_{i}, R_{k}, \tau_{n}) = \frac{ \sum_{\tau_{i} \in T} \mathbf{I}\Big(x_{t} \in \tau_{i} \, \wedge\, \exists x_t \in D_{j} \, \wedge\, E(S_{\tau_n}, S_{\tau_{i}}) \leq \delta \, \wedge\, \exists x_{t} \in C_{i} \, \wedge\, \exists x_{t} \in R_{k}\Big)}
    { \sum_{\tau_{i} \in T} \mathbf{I}\Big( x_{t} \in \tau_{i} \, \wedge\, E(S_{\tau_n}, S_{\tau_{i}}) \leq \delta \, \wedge\, \exists  x_{t}  \in C_{i} \, \wedge\, \exists x_{t} \in R_{k}\Big)}
    \hfill
    \label{eq:dest-route}
\end{equation}

\noindent
in cases where the route polygons are known not to interact with the historical tracks, such as in the case of Figure~\ref{fig:3}, we can simplify the conditional probabilistic destination model from above as follows:
\begin{equation}
    \hfill
    \mathbb{P}(D_{j} | C_{i}, \tau_{n}) = \frac{ \sum_{\tau_{i} \in T} \mathbf{I}\Big(x_{t} \in \tau_{i} \, \wedge\, \exists x_{t} \in D_{j} \, \wedge\, E(S_{\tau_n}, S_{\tau_{i}}) \leq \delta \, \wedge\, \exists x_{t} \in C_{i} \Big)}
    { \sum_{\tau_{i} \in T} \mathbf{I}\Big(x_{t} \in \tau_{i} \, \wedge\, E(S_{\tau_n}, S_{\tau_{i}}) \leq \delta \, \wedge\, \exists x_{t} \in C_{i}\Big)}
    \hfill
    \label{eq:dest-only}
\end{equation}

\noindent
Equations~\ref{eq:route}, \ref{eq:dest-route}, and \ref{eq:dest-only} take into account all the tracks in the dataset.
The selection of tracks in the numerator and denominator is guided by the indicator function $\mathbf{I}(\cdot)$, which outputs 1 if a $\tau_{i}$ satisfies all conditions.

Equation~\ref{eq:route} is defined in terms of the ratio of total historical tracks that have visited $C_{i}$ out of which tracks have visited $R_{k}$.
In other words, a track for the indicator function is selected if: {\bf (a)} any point $x_{t}$ of track $\tau_{i}$ falls within the route polygon $R_{k}$ and $C_{i}$; and {\bf (b)} the Euclidean distance $E$ of motion statistics within $C_{i}$ is less than a threshold $\delta$.
Equation~\ref{eq:dest-route} considers the destination probability with respect to the intersection with a route polygon $R_k$, which has the same criteria as in Equation~\ref{eq:route} but with a change in {\bf (b)}: a point within the track falls into the destination cell $D_{j}$, instead of the route polygon.
For Equation~\ref{eq:dest-only}, which calculates destination probability without considering an interacting route polygon, the criteria are much the same as \ref{eq:route} and \ref{eq:dest-route}, but it no longer requires a point in the track to fall into the route polygon $R_{k}$.

The proposed models use conditional probability to add new features to vessel trajectory data.
These features, called \textit{Probabilistic Features}, provide additional information, such as the predicted route polygon and destination grid cell for each trajectory.
The new features are obtained by converting transition probabilities into weighted similarity scores and those into cell-grid centroids, which are then added to the AIS messages within the vessel track.
For example, a point $x_{t}$ at time $t$ from a track $x \in \tau$ is now known to be of shape $x_{t} = \left<\mathcal{L}_{on}, \mathcal{L}_{at}, v, \Delta_{v}, \theta, \Delta_{\theta}, \mathcal{L}_{on}^{R}, \mathcal{L}_{at}^{R}, \mathcal{L}_{on}^{N}, \mathcal{L}_{at}^{N}, \mathcal{L}_{on}^{D}, \mathcal{L}_{at}^{D} \right>$.
Here, $\mathcal{L}^{R}$ indicates the centroid coordinates of the route polygon if the track is predicted to intersect with at least one of the known routes; $\mathcal{L}^{N}$ indicates the centroid coordinates of the grid cell where the vessel currently lies; and, $\mathcal{L}^{D}$ indicates the coordinates of the destination grid cell of the track; where, $C$ is all grid cells, $R \subseteq C$, $N \subseteq C$ and $D \subseteq C$.

Defining the final set of features relies on conditions that depend on the number of routes identified in the dataset and the intersection of the trajectories of interest with the route polygons.
In this sense, defining $\mathcal{L}^{D}$, which is the same for an entire track, depends on a similarity-based score calculation:
\begin{equation}
    \hfill
    \mathbb{S}(D_{j} | C_{i}, S_{\tau}) = 
    \begin{cases} 
      \xi \times \mathbb{P}(D_{j} | C_{i}, R_{k}, S_{\tau}) & \text{if} \,\, \mathbf{max}_{k}[\xi \times \mathbb{P}(R_{k} | C_{i}, S_{\tau})  ] > 0,\\
      \xi \times \mathbb{P}(D_{j} | C_{i}, S_{\tau}) & \text{otherwise}.
    \end{cases}
    \hfill
\end{equation}

\noindent
Here, $\xi = \left(1 - E\left(S_{\tau_n}, S_{\tau_{i}}\right)\right)$ is the similarity value calculated based on the Euclidean distance.
The destination of a track is conditional to the existence of a route polygon that interacts with the trajectory or is independent of it otherwise; knowing that $C_i$ is where the vessel lies, $D_j$ is the possible destination within any of the 310 cells, and $R_k$ is one of the 4 route polygons previously identified in the gulf.
Differently, defining $\mathcal{L}^{R}$, which is also shared with the entire track, depends on the next equation:
\begin{equation}
    \hfill
    \mathbb{S}(R | C_{i}, S_{\tau}) = 
    \begin{cases} 
       \mathbf{max}_{k}[ \xi \times \mathbb{P}(R_{k} | C_{i}, S_{\tau})] & \text{if} \,\, \mathbf{max}_{k}[\xi \times \mathbb{P}(R_{k} | C_{i}, S_{\tau})] > 0,\\
       \mathbf{max}_{j}[\xi \times \mathbb{P}(D_{j} | C_{i}, S_{\tau})] & \text{otherwise}.
    \end{cases}
    \hfill
\end{equation}

\noindent
such a case takes into consideration that not all tracks interact with a known route polygon, and whenever this happens, the centroid of the route polygon $\mathcal{L}^{R}$ is replaced by the centroid of the most probable destination $\mathcal{L}^{D}$ among the grid cell in the gulf to increase the robustness of the probabilistic model.

\subsection*{Trigonometrical Transformations}

Predictive modeling for trajectory data presents challenges in representing geospatial data when measured in degrees or gradians, such as for longitude, latitude, and bearing.
Traditional Recurrent Neural Networks, including Long Short-Term Memory (LSTM) networks, may not be able to properly interpret these variables' cyclical and cardinal nature when directly used in learning, even if $[0, 1]$ normalized.
This can result in artificially induced discontinuities and misinterpreted spatial relationships within the dataset.
That is because the degree space is not contiguous with the physical space, as is the case of the Cartesian plane.

To better handle degree-based data, it is helpful to use transformations that maintain their inherent cyclical relationships.
When dealing with longitude and latitude values, we can project them onto a unit sphere ({\it i.e.}, radius one).
This mapping method transforms the data onto a three-dimensional surface while encoding its geodesic similarities into a nearly Cartesian space using the following sequence of equations:
\begin{equation}
    \hfill
    \begin{aligned}
        \alpha &= \cos\Big(\mathcal{L}_{on}\Big) \times \cos\Big(\mathcal{L}_{at}\Big) \\
        \beta &= \sin\Big(\mathcal{L}_{on}\Big) \times \cos\Big(\mathcal{L}_{at}\Big) \\
        \gamma &= \sin\Big(\mathcal{L}_{at}\Big)
    \end{aligned}
    \label{eq:coordinates}
    \hfill
\end{equation}

\noindent
where $\mathcal{L}_{on}$ is the longitude and $\mathcal{L}_{at}$ is the latitude, and $\alpha, \beta, \gamma \in [0, 1]$.
This type of transformation is designed to maintain the spatial relationships between points on the Earth's surface, ensuring that nearby points remain adjacent in the transformed representation.
However, it becomes inaccurate when dealing with data that are located too close to the Earth's poles, such as in the Arctic Ocean.
Unlike coordinates, bearing values cannot be projected directly onto a Cartesian space.
However, they can be transformed using cosine and sine functions, which enable a sinusoidal representation of the data.

\begin{figure}[!b]
    \centering
    \includegraphics[width=\textwidth]{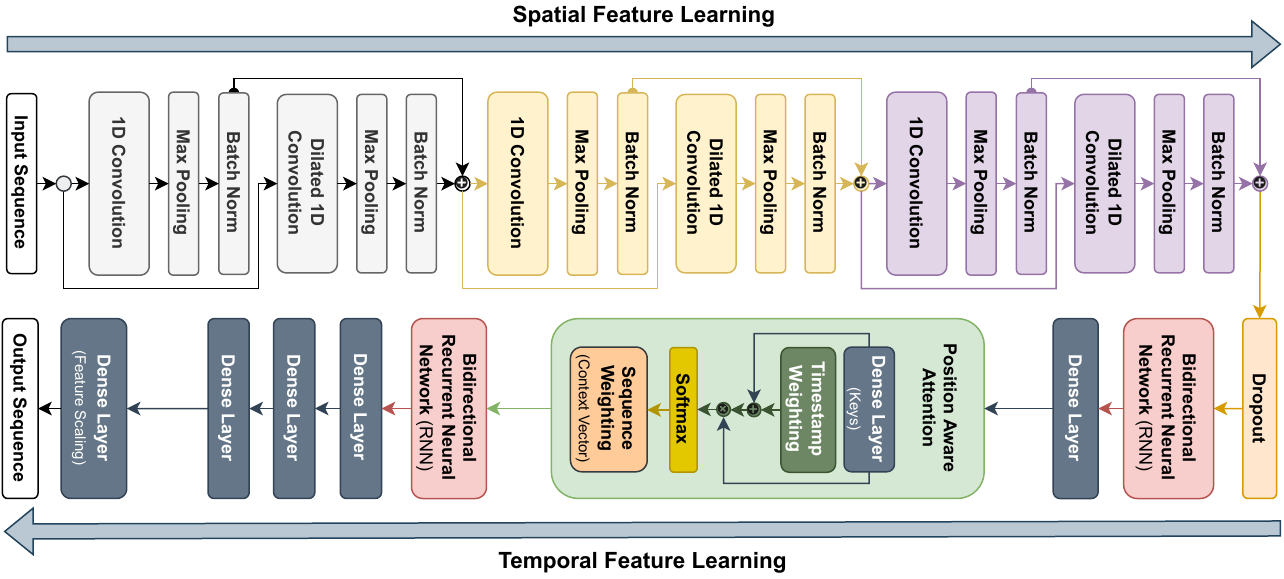}
    \caption{The proposed deep-learning model uses parallel convolutions and a sequential stack of convolution blocks to capture short and long-term spatial features. The extracted features are passed through a dropout layer and a Bi-LSTM with a fully connected layer. The Position-Aware Attention mechanism enhances the importance of later timestamps. The final encoded latent space undergoes a Bi-LSTM decoding phase and a multi-layer fully connected network to produce normalized coordinates.}
    \label{fig:dnn-model}
\end{figure}

\begin{table}[!b]
    \centering
    \caption{Symbols and notations relevant for the deep-learning model formulation understanding.}
        \begin{tabularx}{\textwidth}{c X}
            \toprule
            Notation & Definition \\
            \midrule
            $b$ & Batch size of the input tensor\\
            $w$ & Window size of the input, which represents the number of timestamps \\
            $v$ & Number of features in each input timestamp \\
            $f$ & Size of convolution filters \\
            $s$ & Stride of convolution layers \\
            $m$ & Number of filters in convolution layers \\
            $\mathbf{x}$ & Input tensor to the model, of shape $\mathbb{R}^{b \times w \times v}$ \\
            $\mathbf{W}$ & Representation of the weights of the model layers \\
            $\textbf{b}$ & Representation of the bias parameters of the model layers \\
            $\star$ & Cross-correlation operator used by single-dimensional convolutions \\
            $\widetilde{w}$ & Reduced size of window after convolution \\
            $\epsilon$ & Small constant to avoid division by zero \\
            $\mu_b^d$, $(\sigma_b^d)^2$ & Mean and variance for each filter $d$ within batch $b$ in batch normalization \\
            $\gamma^{~d}$, $\beta^d$ & Scaling and shifting parameters for each filter $d$ in batch normalization \\
            $h$ & Size of max pooling filter \\
            $d$ & Feature dimensions -- changes with each iteration through layers \\
            $\vartheta$ & Dilation factor for kernel expansion in dilated convolutions \\
            $\mathbf{C}$, $\widetilde{\mathbf{C}}$ & Cell state and candidate cell state of the Bi-LSTM \\
            $\mathbf{i}$, $\mathbf{f}$ & Input gate and forget gate of the Bi-LSTM \\
            $\mathbf{o}$, $\mathbf{h}$ & Output gate and hidden state of the Bi-LSTM \\
            $\sigma(\cdot)$ & Sigmoid activation function of the Bi-LSTM\\
            $\phi(\cdot)$ & Activation function used in the model ({\it i.e.}, ReLU) \\
            $(\cdot)\%(\cdot)$ & Integer part of the rest of the division between two numerical values \\
            $\mathbf{A}$ & Attention weights for Position-Aware Attention mechanism \\
            $\mathbf{E}$ & Position-Aware context vector of shape $\mathbb{R}^{b \times z \times d}$\\
            $\otimes$ & Tensor repeating and concatenation operator \\
            $\omega$ & Time factor for establishing a focal point in attention mechanism \\
            $z$ & Window size of the output, which represents the number of output timestamps \\
            $\Xi(\cdot)$ & Re-scaling function for coordinate values generation \\
            $\mathbf{y}$ & Final output tensor representing latitude and longitude, of shape $\mathbb{R}^{b \times z \times 2}$ \\
            \bottomrule
        \end{tabularx}%
    \label{tab:notation-2}%
\end{table}

For speed, applying the logarithmic transformation can be effective, especially when dealing with values spanning across multiple orders of magnitude.
This can ensure that larger-scale patterns do not overly influence the model, allowing local trends to play a part in the decision-making process.
This can be achieved with the equation $v' = \log(1 + |v|)$, where $v'$ is the transformed and $v$ is the original speed.
Through these sets of transformations, the earlier point $x$ at time $t$ from a trajectory $\tau$ become $x_{t} = \Big< \mathcal{L}_{on}, \mathcal{L}_{at}, \alpha, \beta, \gamma, v, \Delta_{v}, \theta, \Delta_{\theta}, v', \Delta_{v'},\\ \cos_\theta, \sin_\theta, \cos_{\Delta_{\theta}}, \sin_{\Delta_{\theta}}, \mathcal{L}_{on}^{R}, \mathcal{L}_{at}^{R},  \alpha^{R}, \beta^{R}, \gamma^{R},  \mathcal{L}_{on}^{N}, \mathcal{L}_{at}^{N}, \alpha^{N}, \beta^{N}, \gamma^{N},$  $ \mathcal{L}_{on}^{D}, \mathcal{L}_{at}^{D}, \alpha^{D}, \beta^{D},  \gamma^{D} \Big>$.
This final representation includes \textit{Standard}, \textit{Probabilistic}, and \textit{Trigonometrical Features}.
It provides an enriched input dataset for the deep-learning model tailored for trajectory forecasting, facilitating its understanding of complex spatial and temporal patterns of the trajectories.
With all feature sets properly defined, we move forward with the formalization of the deep-learning model (see Figure~\ref{fig:dnn-model}), which is based on Table~\ref{tab:notation-2} for describing the symbols and notations used from hereinafter.

\section{Deep Learning Model}
\label{sec:deep-model}

The deep learning model comprises two main parts; spatial and temporal feature learning, as depicted in Figure \ref{fig:dnn-model}.
Our customized architecture uses parallel CNN with recurrent neural networks and focal attention for position-aware temporal learning.
During the model formulation, we use lower-case bold letters to represent vectors, such as $\mathbf{x}$, upper-case bold letters to represent matrices, such as $\mathbf{W}$, and non-bold lower-case values, such as $b$, to indicate a scalar value or an integer index of matrices and vectors.
The equations are defined with indices to help understand how the data changes as it flows through the architecture and to indicate the meaning of each process accomplished by the network module.
We use $\mathbf{x}$ to describe the input tensor, which is propagated and updated at each process step.
The weights and biases, $\mathbf{W}$ and $\mathbf{b}$, are independent of each other but are optimized together during the network training.

Giving an input tensor $\mathbf{x} \in \mathbb{R}^{b \times w \times v}$, where $b$ denotes the batch size, $w$ denotes the window size of the input vectors, and $v$ denotes the number of features.
Our model begins by executing single-dimensional parallel convolutions.
Essentially, we slide a filter (also known as a kernel) of size $f$ across each window element and calculate the dot product across all features with a stride of $s$ for each filter $m$.
This way, we process the input tensor to extract spatial features that will be used in subsequent model layers.
\begin{equation}
    \hfill
    \mathbf{x}_{b,i,m} = \sum_{j=1}^{f}\sum_{n=1}^{v}\mathbf{x}_{b,(i+(s \times (j-1))),n} \cdot \mathbf{W}_{m,j,n} \equiv \mathbf{x}_{b,i} \star \mathbf{W}_{m}
    \hfill
    \label{eq:convolution-1}
\end{equation}

Based on the filter size $f$ and stride $s$, we follow this operation for each window element to yield an output tensor $\mathbf{x} \in \mathbb{R}^{b \times \widetilde{w} \times m}$, where $\widetilde{w}$ represents the new window size and $m$ denotes the kernel size.
The cross-correlation $\star$ operation represents an equivalent of this convolution.
The convolutional layers play a crucial role in extracting spatial features of the input trajectories, especially for long-term trajectory forecasting.
These layers are particularly helpful when the input data is limited to short-term data.
They can learn spatial patterns like curves and slight course changes based on historical data, but they tend to overfit the input data and can only learn patterns that fit their kernel size (short or long-term dependencies).

Subsequently, $\mathbf{x} \in \mathbb{R}^{b \times \widetilde{w} \times m}$ goes through batch normalization to fit the values of the resulting tensor into a standard scale.
This is followed by a max-pooling operation, and the two processes are defined as:
\begin{equation}
    \hfill
    \mathbf{x}_{b,i,m} = \mathbf{max}\left(\frac{\mathbf{x}_{b,p,m} - \mu_b^d}{\sqrt{\big(\sigma_b^d\big)^2 + \epsilon}} \cdot \gamma^{~d} + \beta^d\right) \quad \forall p \in [i, i + h - 1]
    \hfill
    \label{eq:bn-maxpool}
\end{equation}

For each filter $m$ and batch $b$, a window of size $h$ is defined on the feature map starting at position $i$, with the condition that $i + h - 1 \leq \widetilde w$.
The filter-wise mean $\mu_b^d$ and variance $\big(\sigma_b^d\big)^2$ are calculated for batch $b$, along with specific scaling $\gamma^{~d}$ and shifting $\beta^d$ parameters.
Next, the max pooling operation is performed over the window, extracting the maximum value of $p$.
The resulting value is assigned to $\mathbf{x}_{b,i,d}$, which becomes the feature map for the given window of filter index $d$.
Due to working with parallel convolutions that involve different kernel sizes, it is necessary to use a combination of techniques to ensure consistent tensor dimensions.
We use batch normalization and max-pooling to scale down the tensor dimensions into shared tensor dimensions with scaled values, which can be merged later.
This approach guarantees that the branches can be combined seamlessly without stability issues during the training.

As mentioned, single-dimensional CNNs can only learn spatial features within their kernel size.
We employ parallel convolutions in residual blocks to simultaneously capture short-term and long-term spatial features.
This is achieved through dilated convolutional layers, which expand the kernel size by a factor of $\vartheta$ and enable it to capture extended spatial features from the input data.
Both convolutional branches in a block are trained together, but they work independently on copies of the input data so that they are not affected by each other.
To perform dilated convolution, we modify Equation~\ref{eq:convolution-1} as follows:
\begin{equation}
    \hfill
    \mathbf{x}_{b,i,m} = \sum_{j=1}^{f}\sum_{n=1}^{v}\mathbf{x}_{b,(i+(s \times \vartheta \times (j-1))),n} \cdot \mathbf{W}_{m,j,n} \equiv \mathbf{x}_{b,(i + \vartheta)} \star \mathbf{W}_{m}
    \hfill
    \label{eq:convolution-2}
\end{equation}

\noindent
where $j$ is the filter index, and $\vartheta \times (j-1)$ is the expansion of the kernel due to the dilation rate $\vartheta$.
The output from Equation~\ref{eq:bn-maxpool} is then combined into a single tensor as such:
\begin{equation}
    \hfill
    \mathbf{x}_{b,i,m} = \mathbf{x}_{b,i,m}^{T} + \mathbf{x}_{b,i,m}^{D}
    \hfill
    \label{eq:block-joining}
\end{equation}

\noindent
where $T$ and $D$ denote the results obtained from the traditional and dilated convolution operations.
Notice that $\widetilde{w}$ and $m$ are updated each time they go through the CNN blocks.

The series of mathematical operations that end in Equation~\ref{eq:block-joining}, as shown in Figure~\ref{fig:dnn-model}, is repeated three times, each time with different parameters.
At this point, $\widetilde{w}$ and $m$ have dimensions according to the hyperparameters of the convolution block (see Annex).
This stack allows us to extract precise long and short-term spatial features from the input data at each process step.
The parallel CNNs and the sequential blocks work simultaneously to produce a refined output of spatial features that is then sent to the next block for further processing, minimizing further the overall error for the forecasting task.

The subsequent part is where the network learns the temporal features from the input data.
It starts with a dropout layer to prevent overfitting from the spatial feature data.
The dropout layer comprises a binary mask of shape $\mathbb{R}^{b \times \widetilde{w} \times m}$, where each element has a value of 1 with probability $p$ and a value of 0 with probability $1 - p$, which is element-wise multiplied with the input data to generate the output.
Such output is forwarded to a Bidirectional Long-Short Term Memory (Bi-LSTM) network, which takes $\mathbf{x} \in \mathbb{R}^{b \times \widetilde{w} \times m}$ as input and produces the hidden state $\mathbf{h}_{b,i,r} \in \mathbb{R}^{b \times \widetilde{w} \times r}$ for each time step $i \leq \widetilde{w}$, where $r$ is the dimension of the LSTM inner weights.
The Bi-LSTM equations can be formulated as:
\begin{equation}
    \hfill
    \begin{aligned}
        \mathbf{i}_{b,i,r} &= \sigma\Big(\mathbf{W}_{i} \cdot \Big[\mathbf{h}_{b,(i-1),r}~,~ \mathbf{x}_{b,i,m}\Big] + \mathbf{b}_{i}\Big) \\
        \mathbf{f}_{b,i,r} &= \sigma\Big(\mathbf{W}_{f} \cdot \Big[\mathbf{h}_{b,(i-1),r}~,~ \mathbf{x}_{b,i,m}\Big] + \mathbf{b}_{f}\Big) \\
        \widetilde{\mathbf{C}}_{b,i,r} &= \mathbf{tanh}\Big(\mathbf{W}_{C} \cdot \Big[\mathbf{h}_{b,(i-1),r}~,~ \mathbf{x}_{b,i,m}\Big] + \mathbf{b}_{C}\Big) \\
        \mathbf{o}_{b,i,r} &= \sigma\Big(\mathbf{W}_{o} \cdot \Big[\mathbf{h}_{b,(i-1),r}~,~ \mathbf{x}_{b,i,m}\Big] + \mathbf{b}_{o}\Big) \\
        \mathbf{C}_{b,i,r} &= \mathbf{f}_{b,i,r} \cdot \mathbf{C}_{b,(i-1),r} + \mathbf{i}_{b,i,r} \cdot \widetilde{\mathbf{C}}_{b,i,r} \\
        \mathbf{h}_{b,i,r} &= \mathbf{o}_{b,i,r} \cdot \mathbf{tanh}\Big(\mathbf{C}_{b,i,r}\Big) \\
        \mathbf{x}_{b,i,r} &= \phi\Big(\mathbf{W}_{d} \cdot \Big[\mathbf{h}_{b,i,r}^{\rightarrow} + \mathbf{h}_{b,i,r}^{\leftarrow}\Big] + \mathbf{b}_d\Big)
    \end{aligned}
    \hfill
    \label{eq:bi-lstm}
\end{equation}

\noindent
where $\sigma$ is the sigmoid function, $\mathbf{tanh}$ is the hyperbolic tangent function, $[\mathbf{h}_{b,(i-1),r}~,~ \mathbf{x}_{b, i,m}]$ is the concatenation of the previous hidden state with the $i$-th element of the input.
$\mathbf{i}$, $\mathbf{f}$, $\widetilde{\mathbf{C}}$, $\mathbf{o}$, $\mathbf{C}$, and $\mathbf{h}$ represent the input gate, forget gate, candidate cell state, output gate, cell state, and hidden state, respectively.
$\mathbf{x}_{b,i,r}$ is the output of the process, where $\mathbf{W}_d$ and $\mathbf{b}_d$ is the weight and bias for the dense layer after the Bi-LSTM and $\phi$ is the ReLU activation.
Because Bi-LSTMs compute the hidden states in the forward $\mathbf{h}_{b,i,r}^{\rightarrow}$ and backward $\mathbf{h}_{b,i,r}^{\leftarrow}$ direction, the double output is merged before going through the fully-connected layer.

The next step of this process is to have the output of the fully connected layer undergo the focal version of the attention mechanism.
As previously explained, due to working with sequences in which the order is essential, paying greater attention to later timestamps is more critical than early timestamps in the sequence.
As such, we use a Position-Aware Attention mechanism that we propose for this use case, a further contribution of this paper.
Given the input tensor $\mathbf{x}_{b,i,r}$, where $\mathbf{x} \in \mathbb{R}^{b \times \widetilde{w} \times r}$, our model follows as:
\begin{equation}
    \hfill
    \begin{aligned}
        \mathbf{K}_{b,i,r} &= \mathbf{x}_{b,i,r} \cdot \mathbf{W}_{b,i,r} + \mathbf{b}_{b,i,r} \\
        \mathbf{S}_{b,i,r} &= \mathbf{K}_{b,i,r} \cdot \Big(\mathbf{K}_{b,i,r} + \left(\omega \cdot (i~\%~w)\right)\Big) \\
        \mathbf{A}_{b,i,r} &= softmax\Big(\mathbf{S}_{b,i,r}\Big) \\
        \mathbf{E}_{b,z,r} &= \left(\sum_{i=1}^{w}\mathbf{A}_{b,i,r} \cdot \mathbf{x}_{b,i,r}\right) \otimes 1_z
    \end{aligned}
    \hfill
    \label{eq:pa-attention}
\end{equation}

In Equation~\ref{eq:pa-attention}, the attention keys are denoted by $\mathbf{K}_{b,i,m} \in \mathbb{R}^{b \times w \times d}$; and, to establish the focal point of attention, we compute a monotonically increasing score in respect to the index of the input tensor $i$ and the time factor $\omega$.
We apply the dot product operation on the temporal-scaled samples to get the intermediate scores $\mathbf{S}_{b,i,r}$, followed by a \textit{softmax} function to get the attention weights $\mathbf{A}_{b,i,r}$.
Finally, we calculate the context vector $\mathbf{E}_{b,r}$ by performing a weighted summation of input vectors with the attention weights along with the temporal axis $i$.
The context vector is then repeated $z$ times to produce the encoded latent space representation $\mathbf{E}_{b,z,r}$ that will be decoded in the following by the subsequent layers of the model.

After the attention mechanism output ({\it i.e.}, context vector) has been created, it must be decoded to produce meaningful positional results.
That is because, at this point, all the timestamps have the same value.
The output of the attention mechanism is fed into a new Bi-LSTM with a similar formulation as in Equation~\ref{eq:bi-lstm}.
The difference in the decoding part is that the hidden states $\widetilde{\textbf{h}}_{b,z,r} \in \mathbb{R}^{b \times z \times \tilde{r}}$ of the Bi-LSTM goes through three fully connected layers instead of one, each of which can be represented as:
\begin{equation}
    \hfill
        \textbf{x}_{b,z,r}^k = \phi^k\left(\textbf{W}_d^k \cdot \widetilde{\textbf{h}}_{b,z,r} + \textbf{b}_d^k\right)
    \hfill
\end{equation}

\noindent
where $\phi^k$ is the activation function, $\textbf{W}_d^k$ and $\textbf{b}_d^k$ are the weight matrix and the bias for the $k^{th}$ fully connected layer, respectively.
$\textbf{x}_{b,z,d}^k$ is the output of the $k^{th}$ fully connected layer, where $k = 1, 2, 3$, $z$ is the timestamps in the temporal axis, and $d$ the transformed feature dimensions matching the size of the weights of the last fully connected layer.
To produce the final set of forecasted coordinates, $\textbf{x}_{b,z,r}^3$, needs to undergo a last transformation to reduce the axis $d$ into a fixed size of 2, representing the longitude and latitude values:
\begin{equation}
    \hfill
    \textbf{y}_{b,z,\tilde{d}} = \Xi\left(\textbf{W}_{o} \cdot \textbf{x}_{b,z,d}^3 + \textbf{b}_{o}\right)
    \hfill
\end{equation}

\noindent
where $\textbf{y}_{b,i,\tilde{d}} \in \mathbb{R}^{b \times z \times 2}$ and $\Xi$ is a custom re-scaling out function that can map the raw output values defined in $[0,1]$ into the valid range of latitude $[-90,90]$ and longitude $[-180,180]$ coordinate values.
Due to working with a particular focus on the Gulf of St. Lawrence, our re-scaling function decodes the normalized values in the range $[-68, 45]$ for latitude and $[-58, 50]$ for longitude of the coordinate values.

\section{Results}
\label{sec:results}

The evaluation of results begins with the probabilistic features, which constitute the foundation of our proposal and the main catalyst for improved decision-making using neural networks in the task of multi-path long-term vessel trajectory forecasting.
The probability model is trained on the same dataset that is later used to train the deep-learning model.
The crucial difference is that the deep-learning model's success hinges on the probabilistic model's successful implementation.
The probabilistic model is responsible for adding route and destination information in \textit{standard feature} set, while the deep-learning model focuses on trajectory forecasting based on the vessel position and its relevant probabilistic features.
This means that when the deep-learning model has access to data regarding the vessel's probable route and destination, it can significantly decrease the uncertainty in trajectory forecasting, leading to better performance for the forecasting pipeline.

\begin{table}[h!]
    \centering
    \renewcommand{\arraystretch}{1.1}
    \begin{tabular}{r|lccc}
        \toprule
        \multicolumn{5}{c}{\textbf{Probabilistic Model Results}} \\ \hline
        \textbf{Test Type} & \textbf{Coordinate Type} & \textbf{Precision (\%)} & \textbf{Recall (\%)} & \textbf{F1 Score (\%)} \\ \hline
        \multirow{2}{*}{Cargo Test}  & Route  & 83.13                & 83.16           & 81.19                 \\ 
                                     & Destination      & 75.28                & 79.09           & 75.85                 \\ \hline
        \multirow{2}{*}{Tanker Test} & Route  & 86.89                & 89.92           & 87.84                 \\ 
                                     & Destination      & 74.85                & 77.12           & 74.90                 \\ \hline
        \multirow{2}{*}{Cargo Train} & Route  & 81.69                & 83.25           & 80.43                 \\ 
                                     & Destination      & 89.51                & 89.72           & 89.02                 \\ \hline
        \multirow{2}{*}{Tanker Train}& Route  & 83.43                & 84.29           & 82.31                 \\ 
                                     & Destination      & 88.64                & 87.95           & 87.48                 \\ \bottomrule
    \end{tabular}
    \caption{Performance of the probabilistic model in forecasting a vessel's route and destination.}
    \label{tb:proba-results}
\end{table}

The performance evaluations of the probabilistic model across all data segments, for both training and testing datasets, are presented in Table~\ref{tb:proba-results}.
We present the results over the training data for the probabilistic model as it is rule-based and does not store training data.
The overperformance observed in the training datasets --- especially in forecasting the destinations of the trajectories that scored 89\% and 87\% F1 Score for cargo and tanker vessels, respectively --- are expected outcomes.
Forecasting routes in both cases also yield reasonable results, achieving 80\% and 82\%.
More impressively, when utilizing the test data --- entirely unseen for the model --- the outcome for route forecasting exceeded those observed during the training phase, scoring 81\% and 87\% for cargo and tanker vessels, respectively.
However, regarding destination forecasting, we see a consistent decrease in performance, with the scores dropping to 75\% and 74\%, respectively.
The observed performance difference between the training and testing datasets is an expected behavior -- details about the metrics can be found in the Methods section.
Even considering these variations, the results obtained are consistent, advocating for the effectiveness of the probabilistic model.

To evaluate the features' performance in the deep learning model, we conducted experiments that compared all three sets of features by gradually removing parts of the network and observing the results.
This ablation approach allowed us to assess the impact of each feature set on different neural network architectures.
In this context, each model was tested with equal rigor over the {\it Standard}, {\it Probabilistic}, and {\it Trigonometric Feature} sets, denoted as {\bf A}1/{\bf B}1, {\bf A}2/{\bf B}2, and {\bf A}3/{\bf B}3, respectively.
We conducted a series of tests on five different model versions, which we labeled as {\bf C}1 through {\bf C}5.
The details of the tests can be found in Tables~\ref{tab:forecasting-cargos} and~\ref{tab:forecasting-tankers}.
{\bf C}1 represents our proposed model with all its layers included, while {\bf C}2 is the same model but without the parallel convolutions.
On the other hand, {\bf C}3 is the proposed model without the attention mechanism, and {\bf C}4 is the model without both the parallel convolutions and the attention mechanism.
Finally, {\bf C}5 refers to the model that uses simple unidirectional recurrent networks exclusively.

\begin{table}[!t]
    \renewcommand{\arraystretch}{1.1}
    \centering
    \begin{tabular}{c@{\hspace{0.6em}}c@{\hspace{0.55em}}c@{\hspace{0.55em}}c@{\hspace{0.55em}}c@{\hspace{0.55em}}c@{\hspace{0.55em}}c@{\hspace{0.55em}}c@{\hspace{0.55em}}c@{\hspace{0.55em}}}\hline
        \multicolumn{9}{c}{\textit{\textbf{Standard Features}}\hfill||~\textit{\textbf{Cargo Vessels}}} \\ \hline
        & $R^{2}$ Score & MAE & MSE & Mean Err. & 25$^{th}$ Pct. & 50$^{th}$ Pct. & 75$^{th}$ Pct. & Std. Dev. \\ \hline
        \textbf{A}1/\textbf{C}1 & 98.32\% & 0.0794 & 0.0219 & 13.0609 & 4.1290 & 8.1360 & 15.6287 & 16.8534 \\
        \textbf{A}1/\textbf{C}2 & \textbf{98.39\%} & \textbf{0.0738} & \textbf{0.0207} & \textbf{12.0751} & 3.2538 & 7.2295 & \textbf{14.3310} & \textbf{16.7377} \\
        \textbf{A}1/\textbf{C}3 & 97.65\% & 0.0790 & 0.0263 & 12.6353 & 2.6082 & 6.5676 & 15.2610 & 18.7873 \\
        \textbf{A}1/\textbf{C}4 & 97.96\% & 0.0755 & 0.0241 & 12.1253 & \textbf{2.3751} & \textbf{6.1875} & 14.6569 & 18.3074 \\
        \textbf{A}1/\textbf{C}5 & 97.43\% & 0.0844 & 0.0290 & 13.4349 & 2.9712 & 7.2248 & 16.2579 & 19.6830 \\ \hline
        \multicolumn{9}{c}{\textit{\textbf{Probabilistic Features}}\hfill||~\textit{\textbf{Cargo Vessels}}} \\ \hline
        \textbf{A}2/\textbf{C}1 & 98.20\% & 0.0804 & 0.0297 & 13.8892 & 4.0198 & 8.8602 & 16.3392 & 21.7328 \\
        \textbf{A}2/\textbf{C}2 & \textbf{98.21\%} & 0.0816 & \textbf{0.0292} & 13.9605 & 4.2302 & 9.1191 & 16.7682 & \textbf{21.3547} \\
        \textbf{A}2/\textbf{C}3 & 97.89\% & \textbf{0.0763} & 0.0346 & \textbf{12.7730} & \textbf{2.4753} & \textbf{6.2913} & \textbf{14.3849} & 24.7063 \\
        \textbf{A}2/\textbf{C}4 & 97.91\% & 0.0787 & 0.0354 & 13.2785 & 2.6182 & 6.6299 & 15.1287 & 24.9233 \\
        \textbf{A}2/\textbf{C}5 & 97.52\% & 0.0882 & 0.0418 & 14.9181 & 3.1329 & 7.7203 & 17.1389 & 26.7791 \\ \hline
        \multicolumn{9}{c}{\textit{\textbf{Trigonometrical Features}}\hfill||~\textit{\textbf{Cargo Vessels}}} \\ \hline
        \textbf{A}3/\textbf{C}1 & \textbf{98.09\%} & \textbf{0.0737} & \textbf{0.0315} & 12.5173 & 2.9758 & 6.6108 & 13.9366 & \textbf{23.4030} \\
        \textbf{A}3/\textbf{C}2 & 98.08\% & 0.0740 & 0.0323 & \textbf{12.5150} & 2.9509 & 6.6810 & \textbf{13.9237} & 23.8451 \\
        \textbf{A}3/\textbf{C}3 & 97.80\% & 0.0752 & 0.0376 & 12.7279 & \textbf{2.3615} & \textbf{6.0779} & 14.0855 & 26.1927 \\
        \textbf{A}3/\textbf{C}4 & 97.70\% & 0.0801 & 0.0397 & 13.5411 & 2.6477 & 6.6960 & 15.2335 & 26.7318 \\
        \textbf{A}3/\textbf{C}5 & 97.39\% & 0.0874 & 0.0448 & 14.8360 & 3.0500 & 7.5764 & 16.7728 & 28.1661 \\ \hline
    \end{tabular}
    \caption{Comparison of forecasting models over cargo vessels in the Gulf of St. Lawrence. The best performances in each column are highlighted with bold numbers. C1 represents our proposed model with all its layers included, while C2 is the same model but without the parallel convolutions. C3 is the proposed model without the attention mechanism, and C4 is without parallel convolutions and attention mechanisms. Finally, C5 refers to the model that uses simple unidirectional recurrent networks exclusively.}
    \label{tab:forecasting-cargos}
\end{table}

Our first noteworthy observation lies in the slight numerical variation in the $R^2$ Score across different experiments, as shown in Tables~\ref{tab:forecasting-cargos} and~\ref{tab:forecasting-tankers}.
This trend can be attributed to the nature of the data and the geographic traits of the Gulf of St. Lawrence region, which is an area with well-defined shipping lanes where most cargo and tanker vessels exhibit linear travel behavior with minimal or zero variations within the segments used to train our deep-learning model.
Simultaneously, this pattern provides insights into the performance benefits of using probabilistic and trigonometrical features compared to the standard ones.
This is particularly noticeable when evaluating the 25$^{th}$, 50$^{th}$, and 75$^{th}$ percentile values of the distribution of Haversine distances between all observed and all predicted coordinates.
For cargo vessels, as presented in Table~\ref{tab:forecasting-cargos}, we observe variations in the Haversine distance of the predicted to the observed coordinates between 2 to 4 km for the 25$^{th}$, 6 to 8 for the 50$^{th}$, and 14 to 16 km for the 75$^{th}$ percentile across the standard features.
Although error margins slightly increase with probabilistic features, they decrease significantly with trigonometric features integration, indicating better vessel route prediction.

\begin{table}[!t]
    \renewcommand{\arraystretch}{1.1}
    \centering
    \begin{tabular}{c@{\hspace{0.6em}}c@{\hspace{0.55em}}c@{\hspace{0.55em}}c@{\hspace{0.55em}}c@{\hspace{0.55em}}c@{\hspace{0.55em}}c@{\hspace{0.55em}}c@{\hspace{0.55em}}c@{\hspace{0.55em}}}\hline
        \multicolumn{9}{c}{\textit{\textbf{Standard Features}}\hfill||~\textit{\textbf{Tanker Vessels}}} \\ \hline
        & $R^{2}$ Score & MAE & MSE & Mean Err. & 25$^{th}$ Pct. & 50$^{th}$ Pct. & 75$^{th}$ Pct. & Std. Dev. \\ \hline
        \textbf{B}1/\textbf{C}1 & 98.17\% & 0.0751 & 0.0215 & 12.3757 & 3.9366 & 7.7236 & 14.5098 & 17.3276 \\
        \textbf{B}1/\textbf{C}2 & \textbf{98.32\%} & \textbf{0.0691} & \textbf{0.0198} &\textbf{ 11.3193 }& 3.0421 & 6.7370 & \textbf{13.4186} & \textbf{17.0405} \\
        \textbf{B}1/\textbf{C}3 & 97.10\% & 0.0747 & 0.0259 & 11.9052 & 2.4920 & 6.2030 & 14.2152 & 18.7462 \\
        \textbf{B}1/\textbf{C}4 & 97.74\% & 0.0707 & 0.0232 & 11.3954 & \textbf{2.2510} & \textbf{5.8280} & 13.5352 & 18.3798 \\
        \textbf{B}1/\textbf{C}5 & 96.92\% & 0.0799 & 0.0283 & 12.6215 & 2.7878 & 6.7369 & 15.0114 & 19.6633 \\ \hline
        \multicolumn{9}{c}{\textit{\textbf{Probabilistic Features}}\hfill||~\textit{\textbf{Tanker Vessels}}} \\ \hline
        \textbf{B}2/\textbf{C}1 & \textbf{98.42\%} & 0.0748 & \textbf{0.0242} & 12.9008 & 3.8051 & 8.4623 & 15.4267 & 19.5181 \\
        \textbf{B}2/\textbf{C}2 & 98.36\% & 0.0765 & \textbf{0.0242} & 13.0027 & 4.0693 & 8.6144 & 15.6594 & \textbf{19.3429} \\
        \textbf{B}2/\textbf{C}3 & 98.06\% & \textbf{0.0705} & 0.0293 & \textbf{11.7623} & \textbf{2.3384} & \textbf{5.9108} & \textbf{13.3525} & 22.8520 \\
        \textbf{B}2/\textbf{C}4 & 98.03\% & 0.0740 & 0.0307 & 12.4650 & 2.5137 & 6.3644 & 14.3824 & 23.2749 \\
        \textbf{B}2/\textbf{C}5 & 97.72\% & 0.0833 & 0.0348 & 14.0422 & 3.0119 & 7.4313 & 16.4624 & 24.2642 \\ \hline
        \multicolumn{9}{c}{\textit{\textbf{Trigonometrical Features}}\hfill||~\textit{\textbf{Tanker Vessels}}} \\ \hline
        \textbf{B}3/\textbf{C}1 & \textbf{98.29\%} & \textbf{0.0691} & \textbf{0.0260} & \textbf{11.6770} & 2.8218 & 6.2823 & \textbf{13.1752} & \textbf{21.2020} \\
        \textbf{B}3/\textbf{C}2 & 98.21\% & 0.0696 & 0.0277 & 11.7550 & 2.8866 & 6.4927 & 13.2350 & 22.1301 \\
        \textbf{B}3/\textbf{C}3 & 97.97\% & 0.0698 & 0.0321 & 11.7930 & \textbf{2.2478} & \textbf{5.7668} & 13.1974 & 24.3013 \\
        \textbf{B}3/\textbf{C}4 & 97.92\% & 0.0749 & 0.0332 & 12.6534 & 2.5354 & 6.3981 & 14.4932 & 24.4186 \\
        \textbf{B}3/\textbf{C}5 & 97.55\% & 0.0827 & 0.0384 & 13.9672 & 2.9248 & 7.2341 & 16.0181 & 25.9926 \\ \hline
    \end{tabular}
    \caption{Comparison of forecasting models over tanker vessels in the Gulf of St. Lawrence. The best performances in each column are highlighted with bold numbers. C1 represents our proposed model with all its layers included, while C2 is the same model but without the parallel convolutions. C3 is the proposed model without the attention mechanism, and C4 is without parallel convolutions and attention mechanisms. Finally, C5 refers to the model that uses simple unidirectional recurrent networks exclusively.}
    \label{tab:forecasting-tankers}
\end{table}

Turning to tanker vessels, as shown in Table~\ref{tab:forecasting-tankers}, we find that these models outperform cargo vessels, reflected by lower mean and standard deviation errors.
However, percentile errors remain similar, generally varying by half a kilometer to one kilometer.
These results affirm that our model benefits from incorporating probabilistic features, particularly trigonometrical ones, leading to more accurate route approximation than the standard features.
The best models using only standard features exhibit similar scores compared to the models using the trigonometrical features (such as shown in Table \ref{tab:forecasting-cargos}), indicating that the deep learning model is able to replicate the intricate relationships that we estimate with the probabilistic model.
These results, however, are not consistent across several model runs.
Using the trigonometrical features and the probabilistic model, on the other hand, results in consistent results across several model runs.

More specifically, in Table~\ref{tab:forecasting-cargos}, the {\bf A}1/{\bf C}2 model showed the best performance under the standard features for cargo vessels, with the lowest Mean Absolute Error (MAE) of 0.0738, Mean Squared Error (MSE) of 0.0207, Mean Error of 12.0751 km, and Standard Deviation of 16.7377 km.
For the probabilistic features, model {\bf A}2/{\bf C}3 marginally outperformed others, achieving the lowest Mean Error of 12.7730 km and 25$^{th}$ percentile error of 2.4753 km, while model {\bf A}2/{\bf C}2 recorded the lowest MSE of 0.0292 and Standard Deviation of 21.3547 km.
Among the trigonometric features, model {\bf A}3/{\bf C}1 showed the best results with the lowest MAE of 0.0737 and MSE of 0.0315, while model {\bf A}3/{\bf C}2 showed the lowest Mean Error of 12.5150 km.
In Table~\ref{tab:forecasting-tankers}, the {\bf B}1/{\bf C}2 model under standard features exhibited the best performance for tankers, while under probabilistic features, the {\bf B}2/{\bf C}1 and {\bf B}2/{\bf C}2 models outperformed others in most metrics, and model {\bf B}2/{\bf C}3 depicted the lowest MAE of 0.0705 and Mean Error of 11.7623 km.
Under the trigonometric features, model {\bf B}3/{\bf C}1 showed the lowest values across most evaluation metrics, including MAE of 0.0691, MSE of 0.0260, Mean Error of 11.6770 km, and 75$^{th}$ percentile error of 13.1752 km.

\begin{figure}[!t]
    \vspace{-1.25cm}\hspace{-.65cm}
    \begin{tabular}{@{}c@{\hspace{0.5cm}}c@{\hspace{0.15cm}}c@{\hspace{0.15cm}}c@{}}
    & \subcaptionbox*{\textit{(a) Standard Features}}[0.3\linewidth][c]{} 
    & \subcaptionbox*{\textit{(b) Probabilistic Features}}[0.3\linewidth][c]{}
    & \subcaptionbox*{\textit{(c) Trigonometrical Features}}[0.3\linewidth][c]{} \\
        \rotatebox[origin=c]{90}{\hspace{2.75cm}\textit{Line \#1}}\hspace{-.4cm}
        & \includegraphics[width=0.33\linewidth,trim={0 .5cm 0 1.35cm},clip]{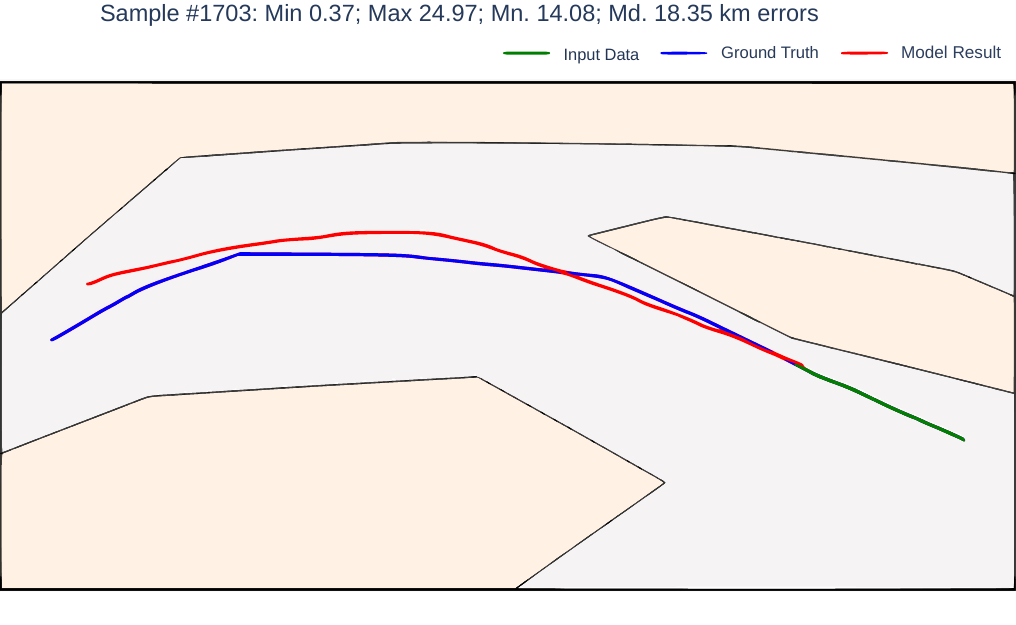}
        & \includegraphics[width=0.33\linewidth,trim={0 .5cm 0 1.35cm},clip]{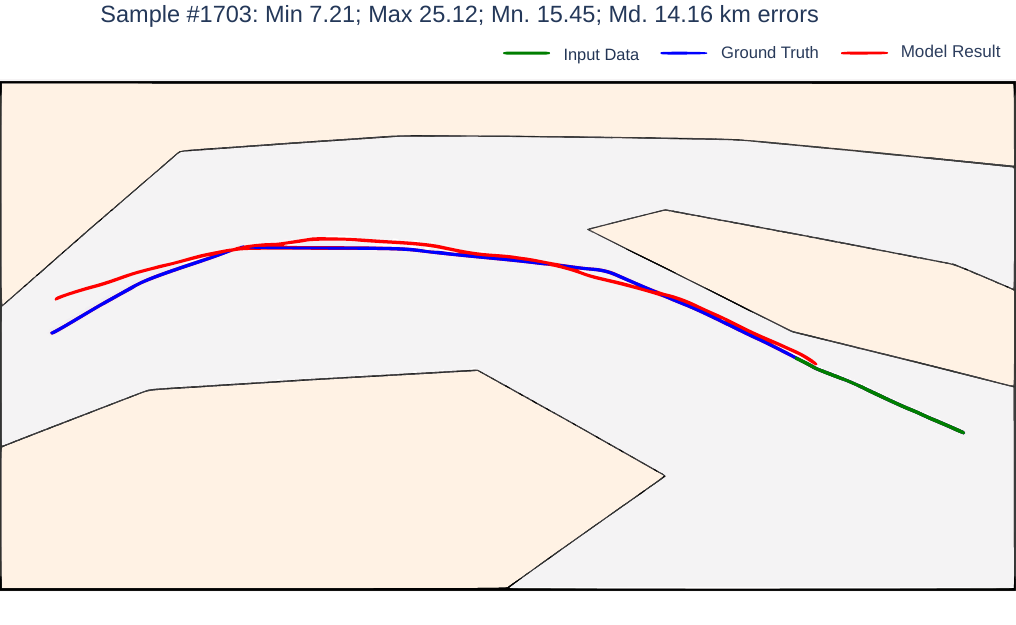}
        & \includegraphics[width=0.33\linewidth,trim={0 .5cm 0 1.35cm},clip]{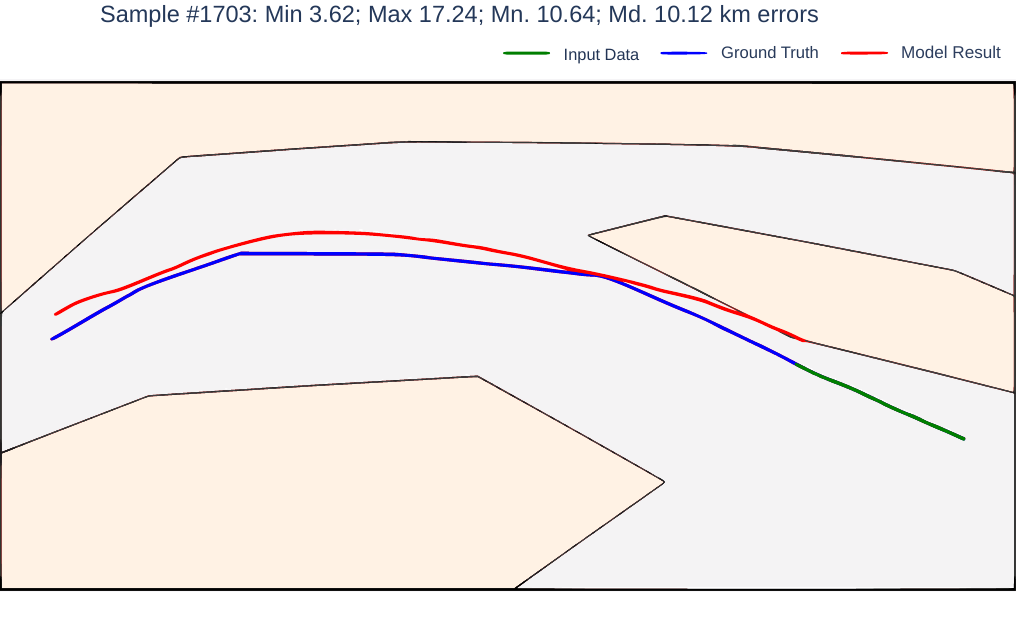}
        \vspace{-1.75cm} \\
        \rotatebox[origin=c]{90}{\hspace{2.75cm}\textit{Line \#2}}\hspace{-.4cm}
        & \includegraphics[width=0.334\linewidth,trim={0 .5cm 0 1.35cm},clip]{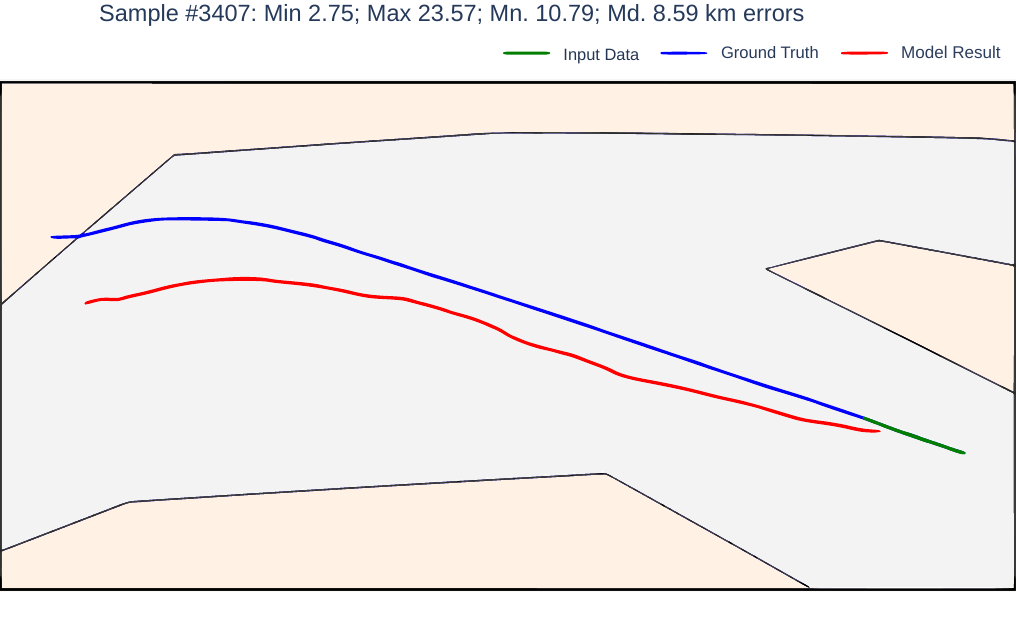}
        & \includegraphics[width=0.33\linewidth,trim={0 .5cm 0 1.35cm},clip]{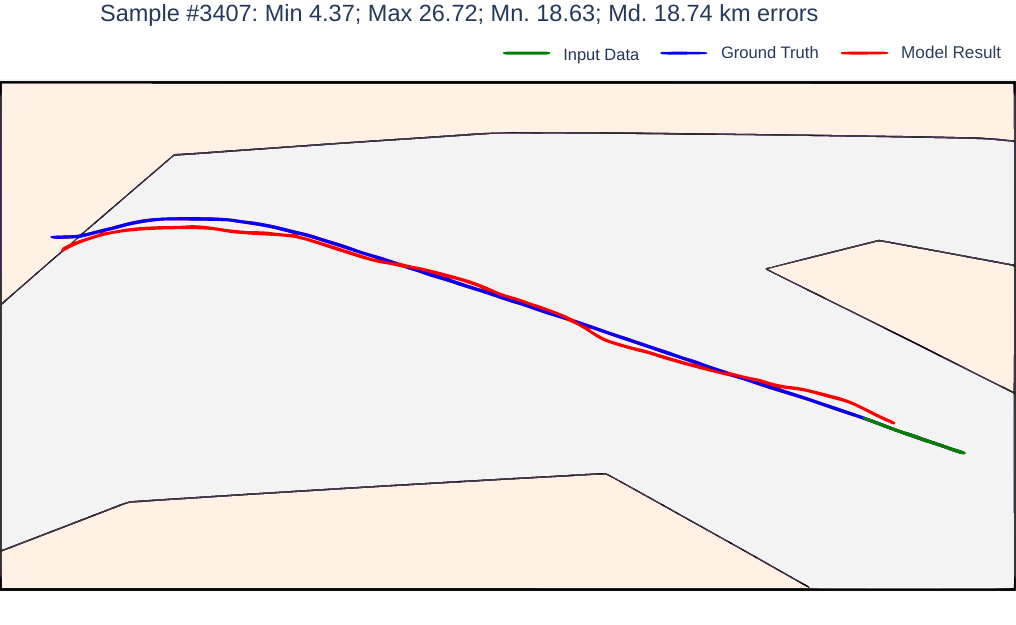}
        & \includegraphics[width=0.33\linewidth,trim={0 .5cm 0 1.35cm},clip]{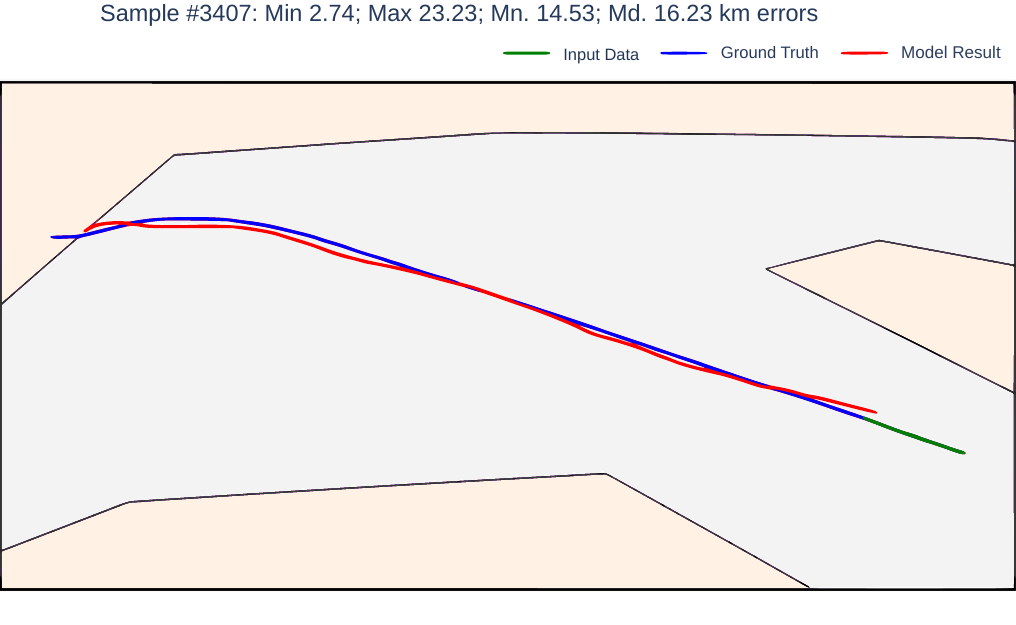}
        \vspace{-1.75cm} \\
        \rotatebox[origin=c]{90}{\hspace{2.75cm}\textit{Line \#3}}\hspace{-.4cm}
        & \includegraphics[width=0.33\linewidth,trim={0 .5cm 0 1.35cm},clip]{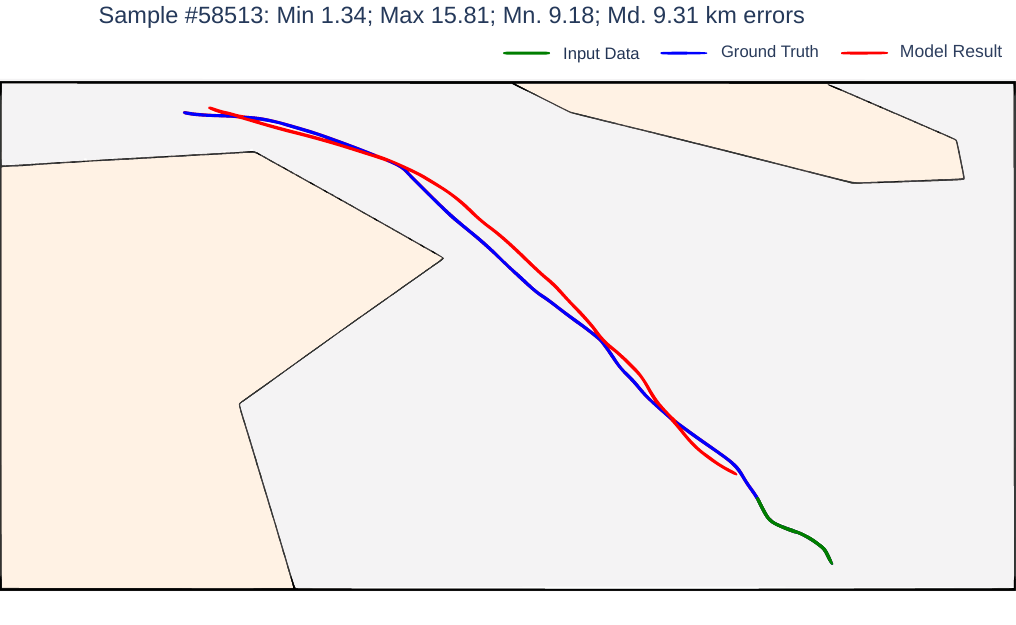}
        & \includegraphics[width=0.33\linewidth,trim={0 .5cm 0 1.35cm},clip]{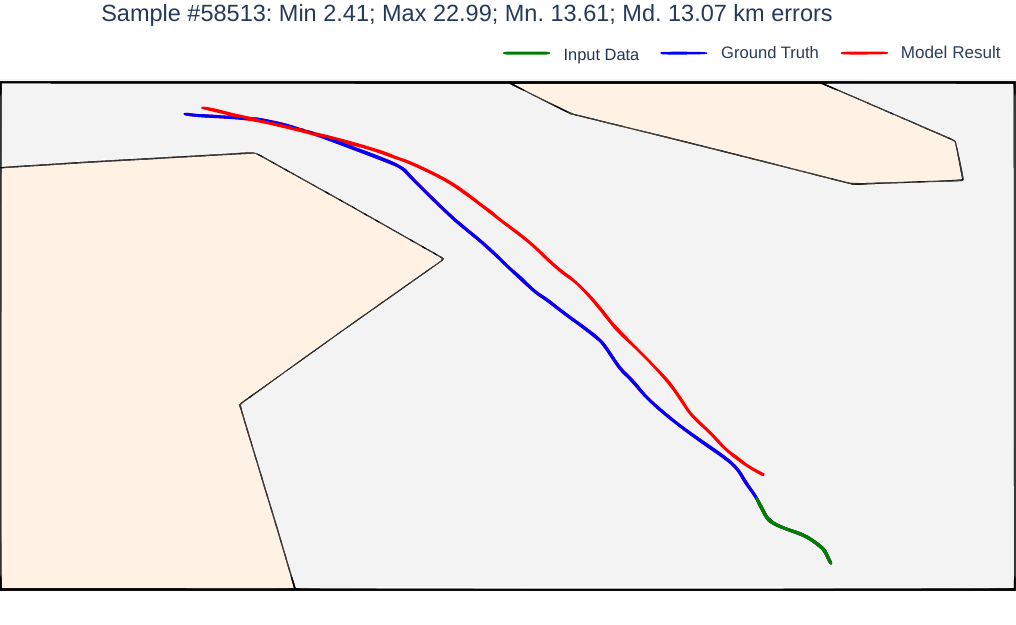}
        & \includegraphics[width=0.33\linewidth,trim={0 .5cm 0 1.35cm},clip]{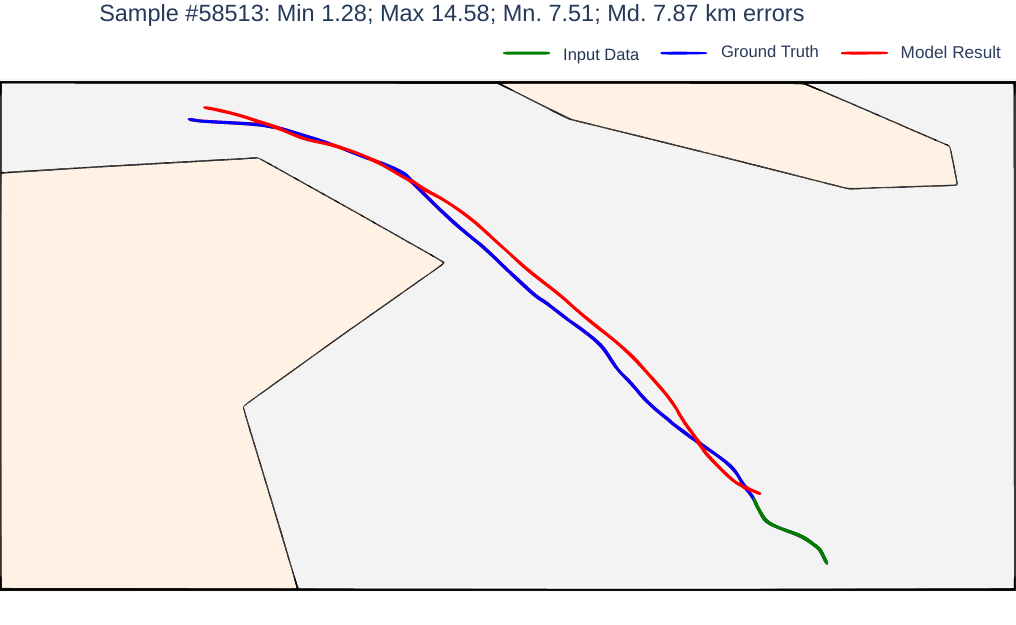}
        \vspace{-1.75cm} \\
        \rotatebox[origin=c]{90}{\hspace{2.75cm}\textit{Line \#4}}\hspace{-.4cm}
        & \includegraphics[width=0.33\linewidth,trim={0 .5cm 0 1.35cm},clip]{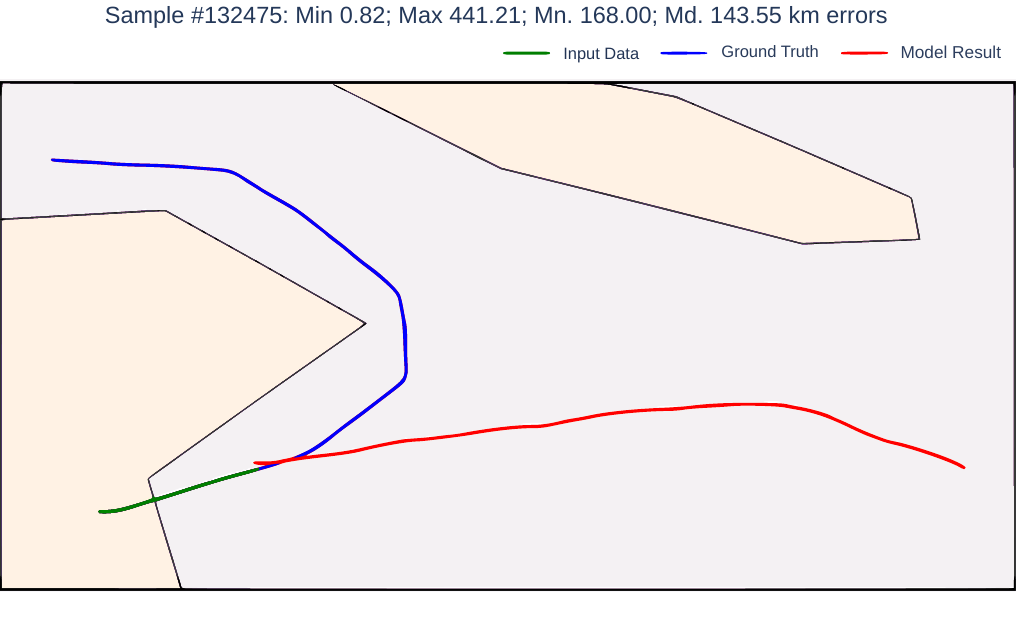}
        & \includegraphics[width=0.33\linewidth,trim={0 .5cm 0 1.35cm},clip]{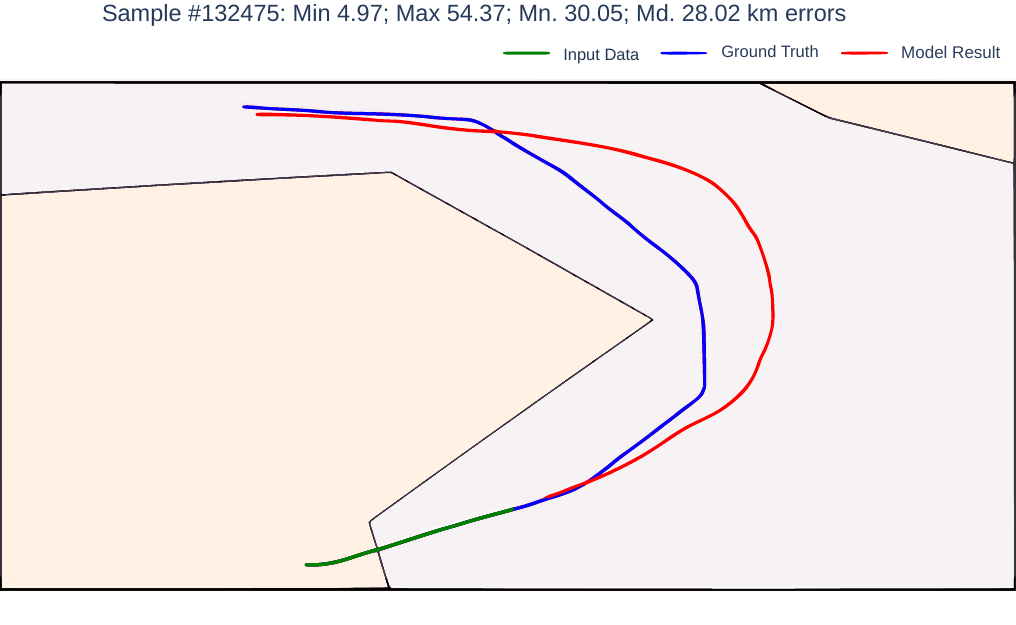}
        & \includegraphics[width=0.33\linewidth,trim={0 .5cm 0 1.35cm},clip]{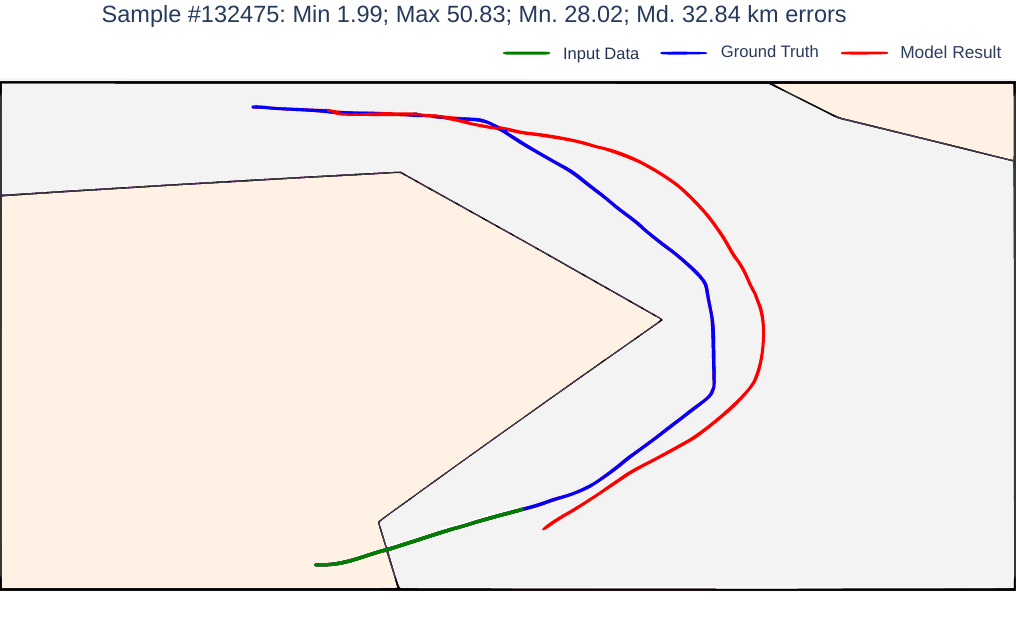}
        \vspace{-1.75cm} \\
        \rotatebox[origin=c]{90}{\hspace{2.75cm}\textit{Line \#5}}\hspace{-.4cm}
        & \includegraphics[width=0.33\linewidth,trim={0 .5cm 0 1.35cm},clip]{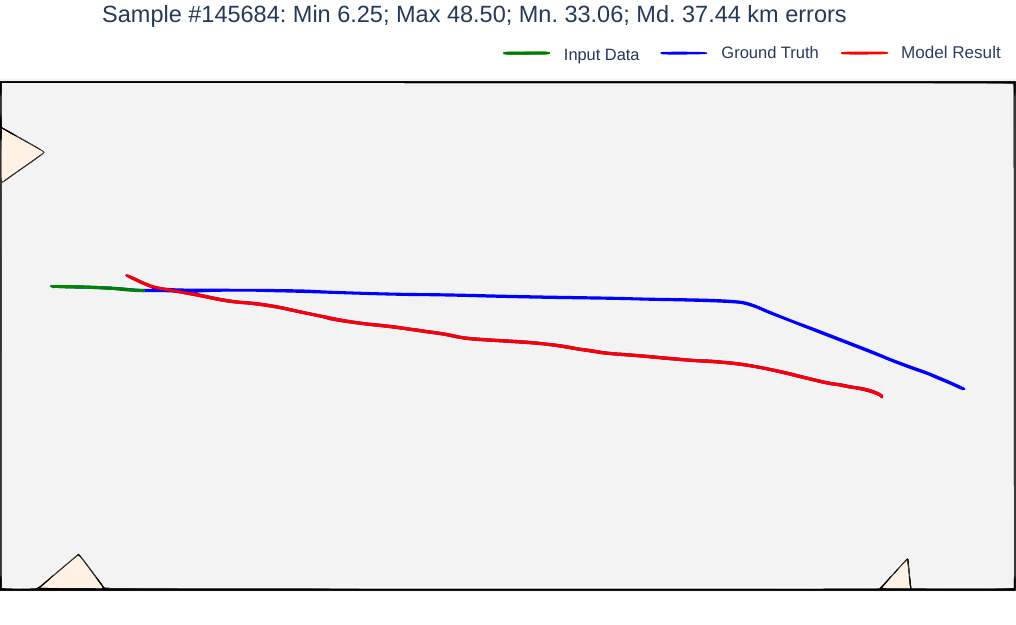}
        & \includegraphics[width=0.33\linewidth,trim={0 .5cm 0 1.35cm},clip]{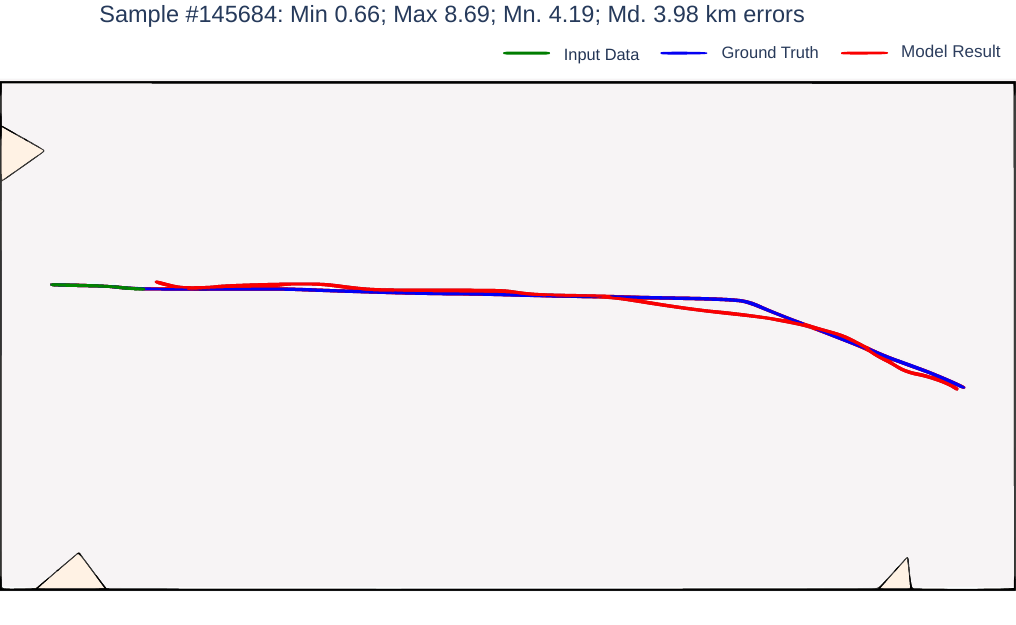}
        & \includegraphics[width=0.33\linewidth,trim={0 .5cm 0 1.35cm},clip]{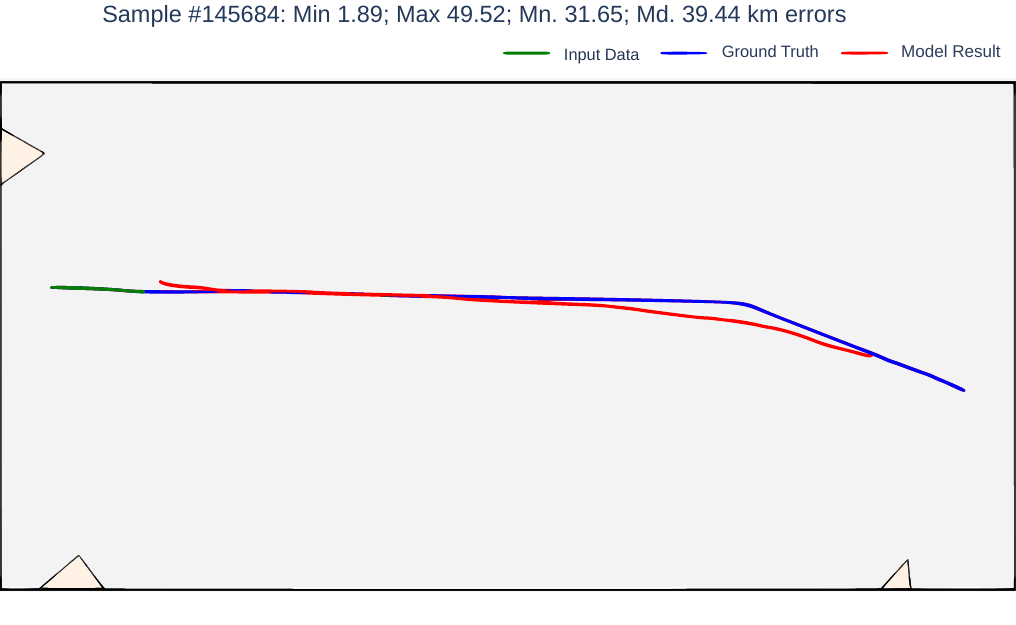}
        \vspace{-1.75cm} \\
    \end{tabular}
    \caption{Proposed Model \textbf{with} \textit{Convolutional Layers} \textbf{and} \textit{Positional-Aware Attention} ({\bf C}1).\\\textbf{Legend:} \textcolor{green!40!black}{\textbf{Green}} lines represent the input data, \textcolor{blue!60!black}{\textbf{Blue}} the ground-truth, and \textcolor{red!80!black}{\textbf{Red}} the model forecasting.}
    \label{fig:dnn-complete}
\end{figure}

Upon conducting a thorough examination of these various models, we can conclude that the complexity of the models varies based on the feature sets, vessel types, and individual model attributes.
Nonetheless, despite these differences, the models consistently exhibited high-performance levels, with dissimilarities primarily arising from the preference for specific feature sets.
Furthermore, our analysis unveiled a deeper facet of the models' capabilities beyond the raw performance statistics.
The models engage in intricate decision-making processes when working with diverse feature sets.
For example, the increased standard deviation observed for trigonometric and probabilistic features does not necessarily indicate poor performance.
Rather, it represents a unique form of dynamic decision-making that results in complex, erratic, yet precise trajectory shapes ({\it i.e.}, routes) that may not be reflected in numerical consistency.

\begin{figure}[!t]
    \vspace{-1.25cm}\hspace{-.65cm}
    \begin{tabular}{@{}c@{\hspace{0.5cm}}c@{\hspace{0.15cm}}c@{\hspace{0.15cm}}c@{}}
    & \subcaptionbox*{\textit{(a) Standard Features}}[0.3\linewidth][c]{} 
    & \subcaptionbox*{\textit{(b) Probabilistic Features}}[0.3\linewidth][c]{}
    & \subcaptionbox*{\textit{(c) Trigonometrical Features}}[0.3\linewidth][c]{} \\
        \rotatebox[origin=c]{90}{\hspace{2.75cm}\textit{Line \#1}}\hspace{-.4cm}
        & \includegraphics[width=0.33\linewidth,trim={0 .5cm 0 1.35cm},clip]{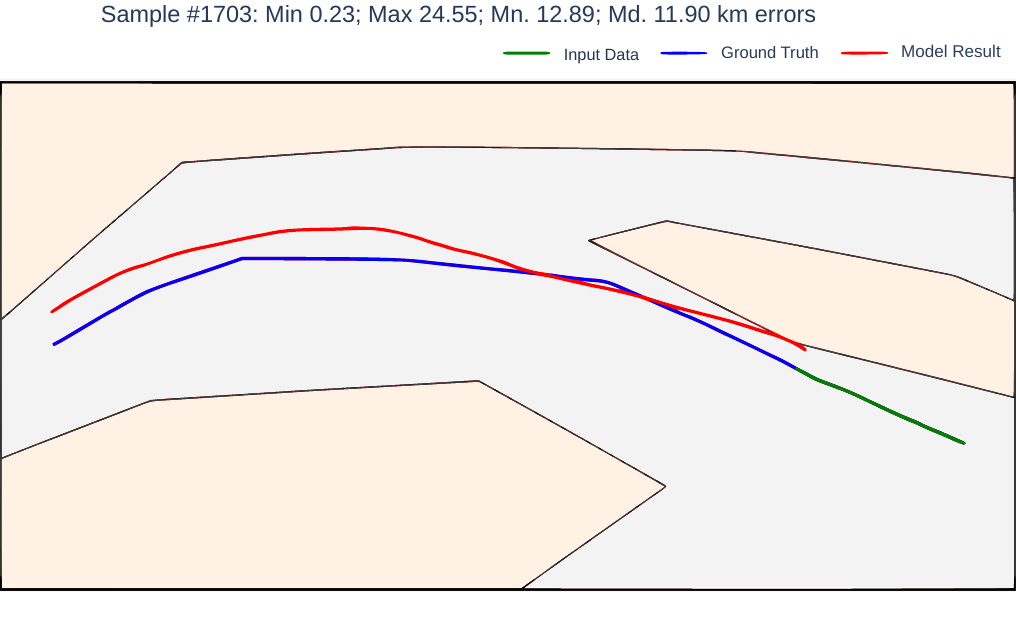}
        & \includegraphics[width=0.33\linewidth,trim={0 .5cm 0 1.35cm},clip]{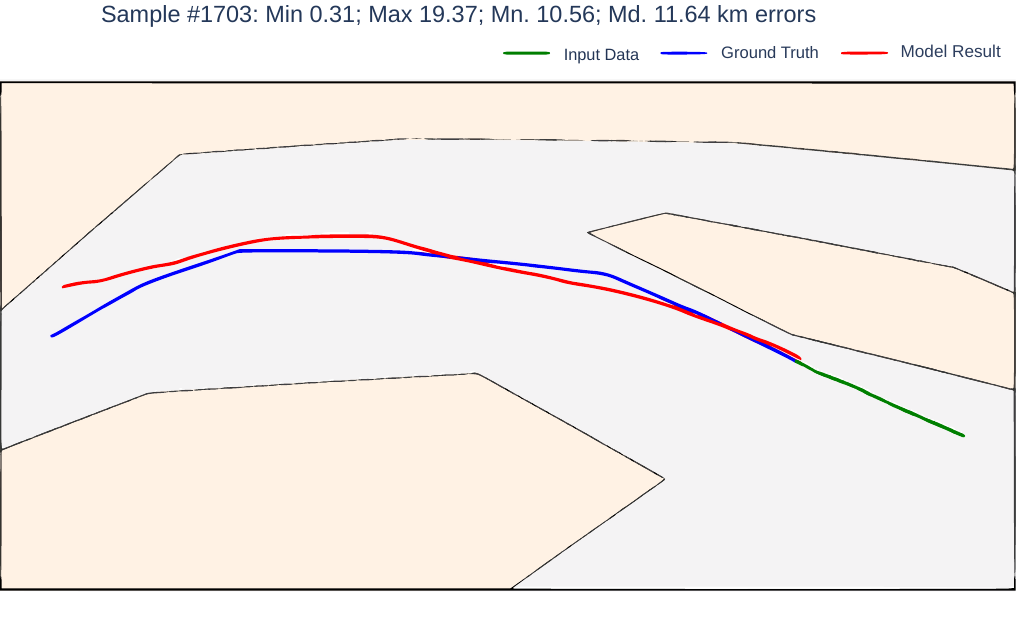}
        & \includegraphics[width=0.33\linewidth,trim={0 .5cm 0 1.35cm},clip]{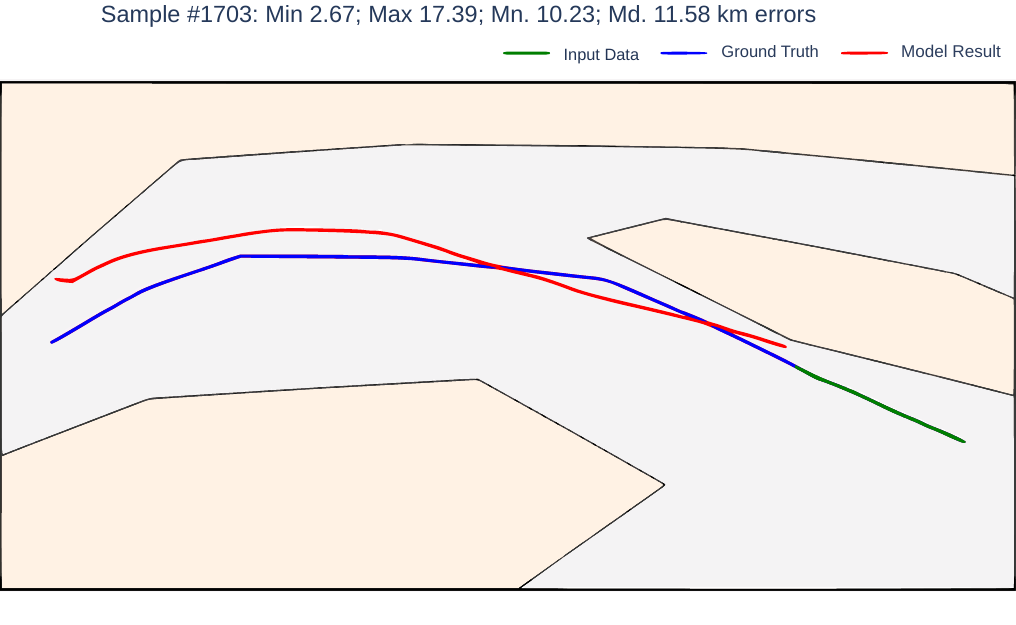}
        \vspace{-1.75cm} \\
        \rotatebox[origin=c]{90}{\hspace{2.75cm}\textit{Line \#2}}\hspace{-.4cm}
        & \includegraphics[width=0.33\linewidth,trim={0 .5cm 0 1.35cm},clip]{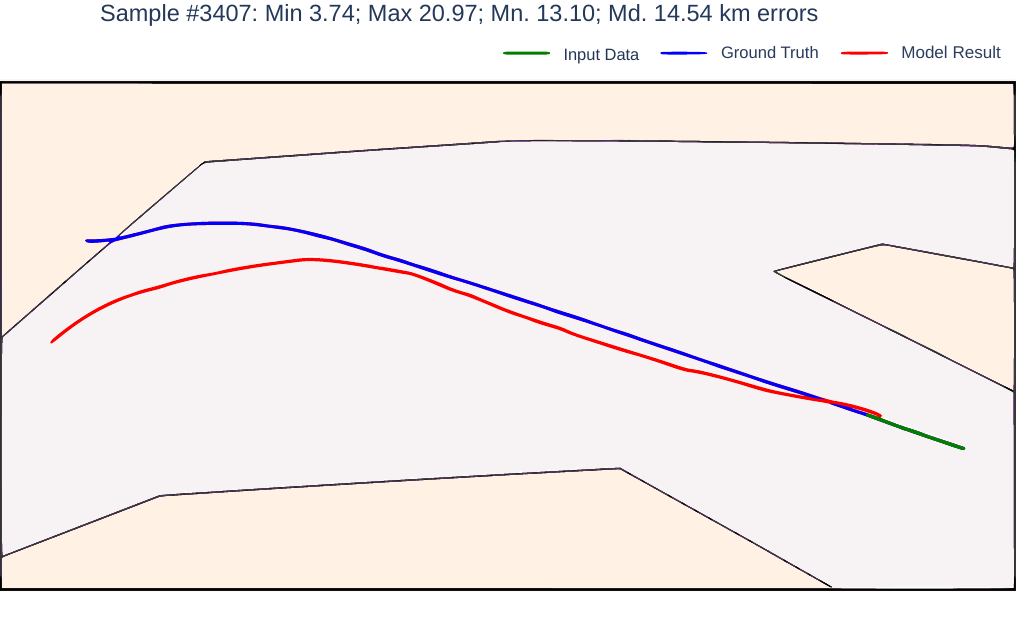}
        & \includegraphics[width=0.33\linewidth,trim={0 .5cm 0 1.35cm},clip]{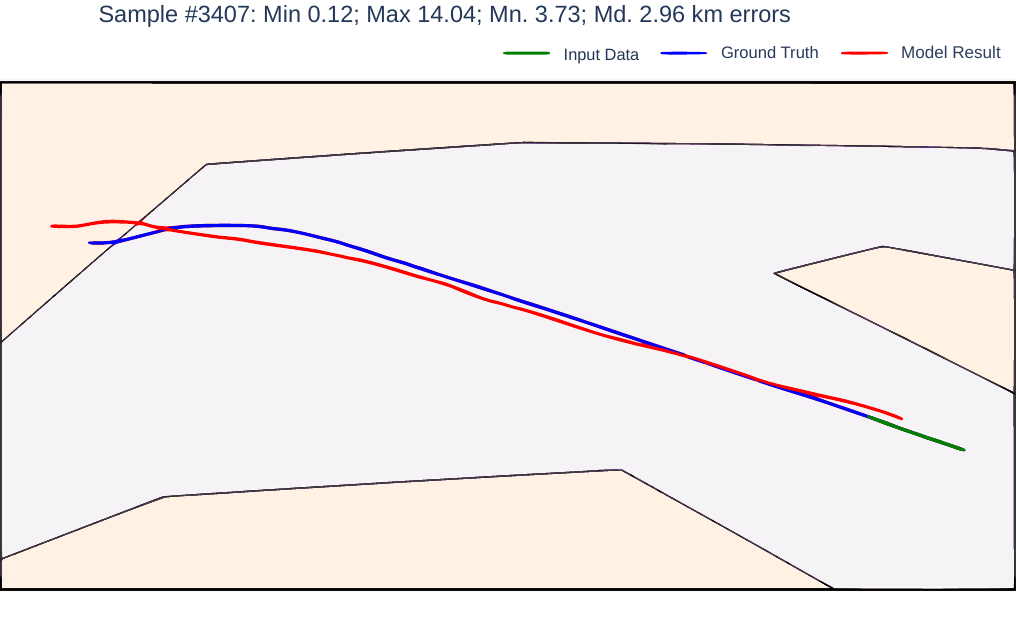}
        & \includegraphics[width=0.33\linewidth,trim={0 .5cm 0 1.35cm},clip]{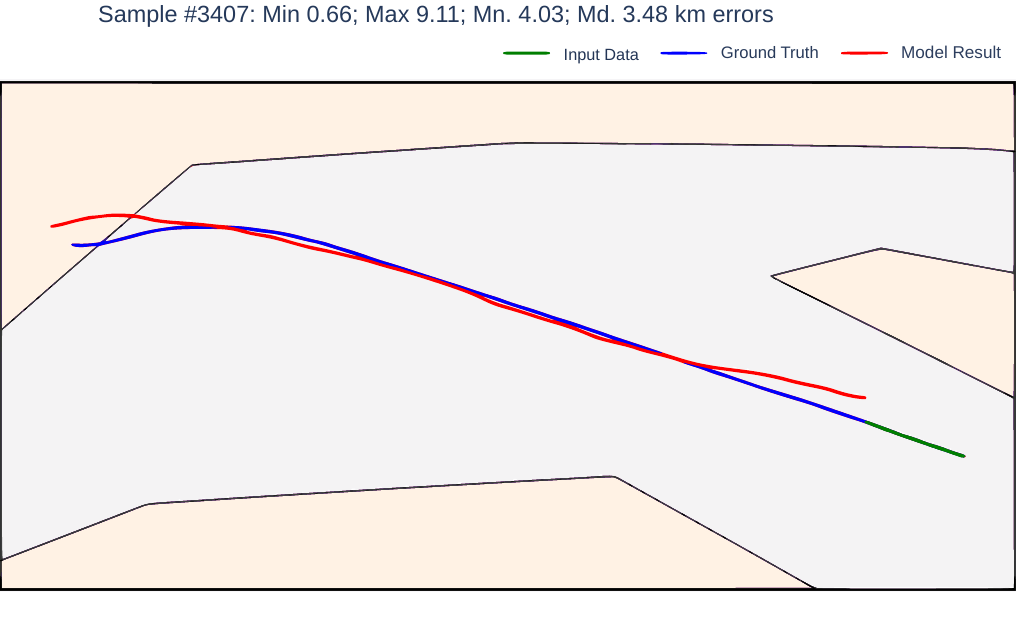}
        \vspace{-1.75cm} \\
        \rotatebox[origin=c]{90}{\hspace{2.75cm}\textit{Line \#3}}\hspace{-.4cm}
        & \includegraphics[width=0.33\linewidth,trim={0 .5cm 0 1.35cm},clip]{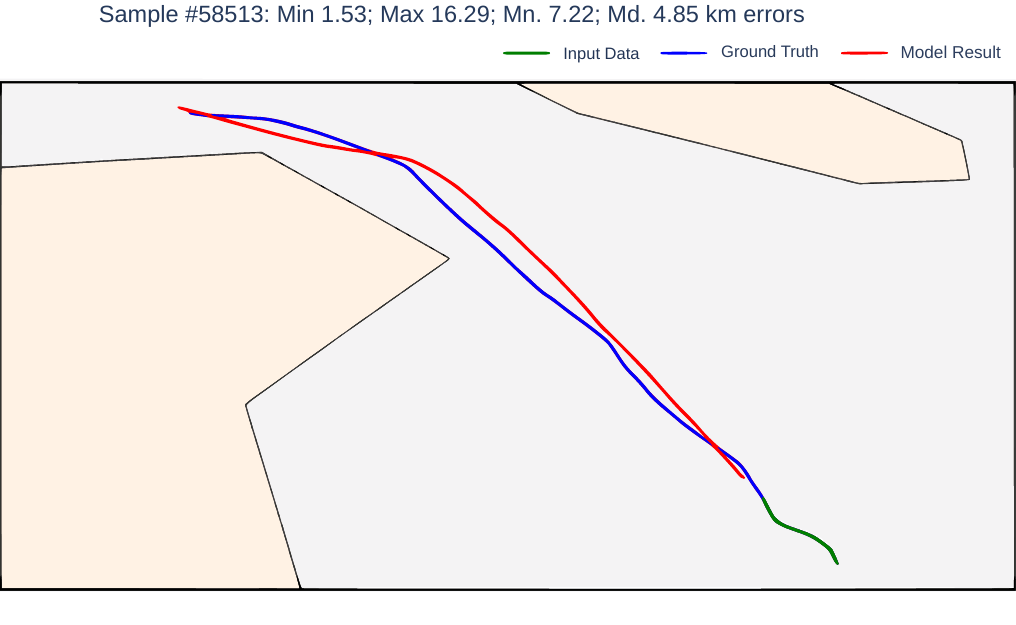}
        & \includegraphics[width=0.33\linewidth,trim={0 .5cm 0 1.35cm},clip]{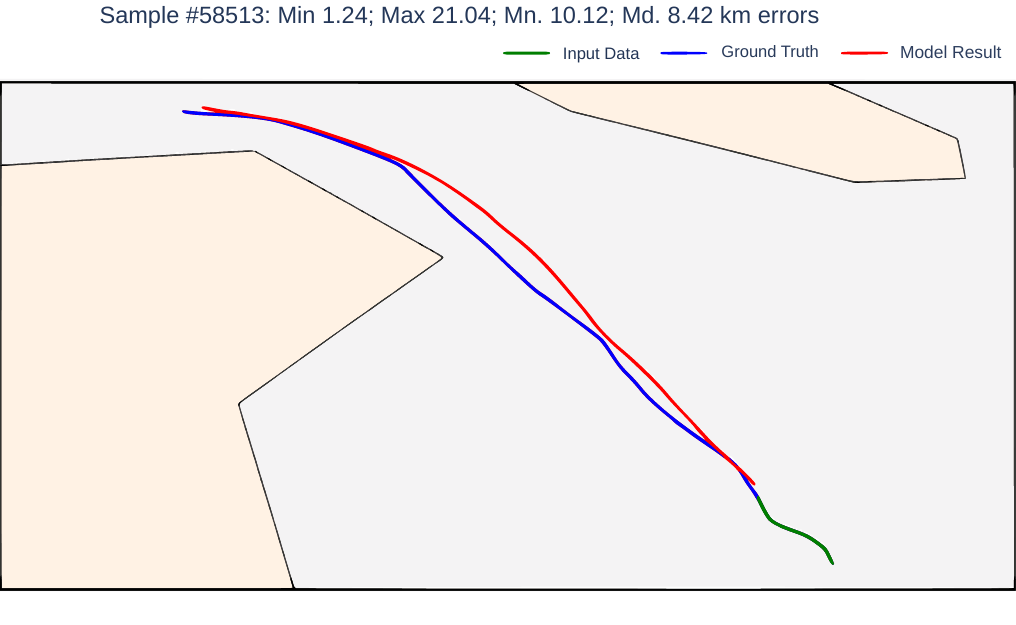}
        & \includegraphics[width=0.33\linewidth,trim={0 .5cm 0 1.35cm},clip]{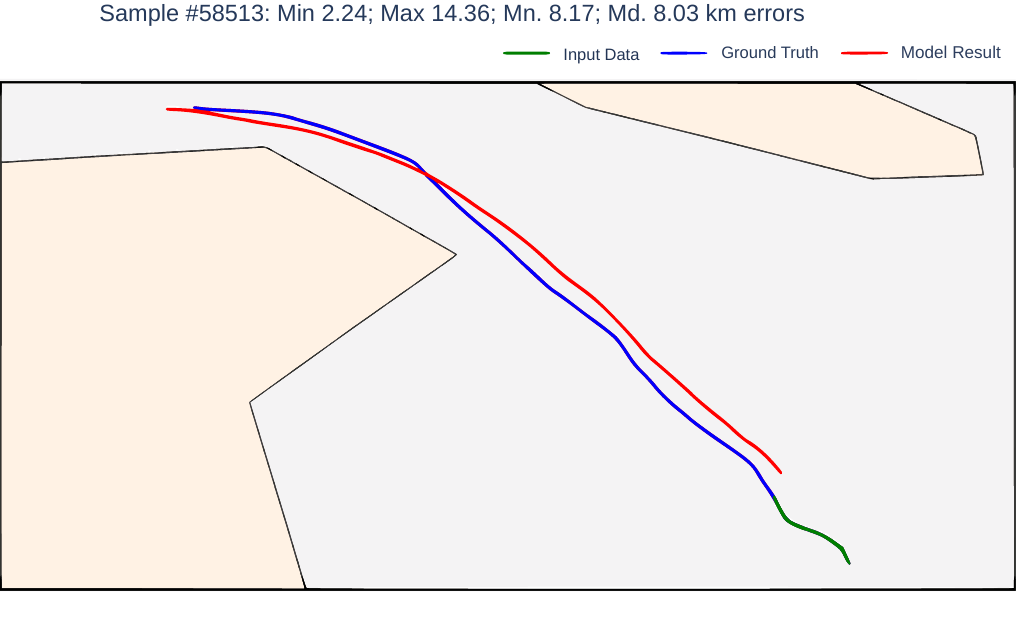}
        \vspace{-1.75cm} \\
        \rotatebox[origin=c]{90}{\hspace{2.75cm}\textit{Line \#4}}\hspace{-.4cm}
        & \includegraphics[width=0.33\linewidth,trim={0 .5cm 0 1.35cm},clip]{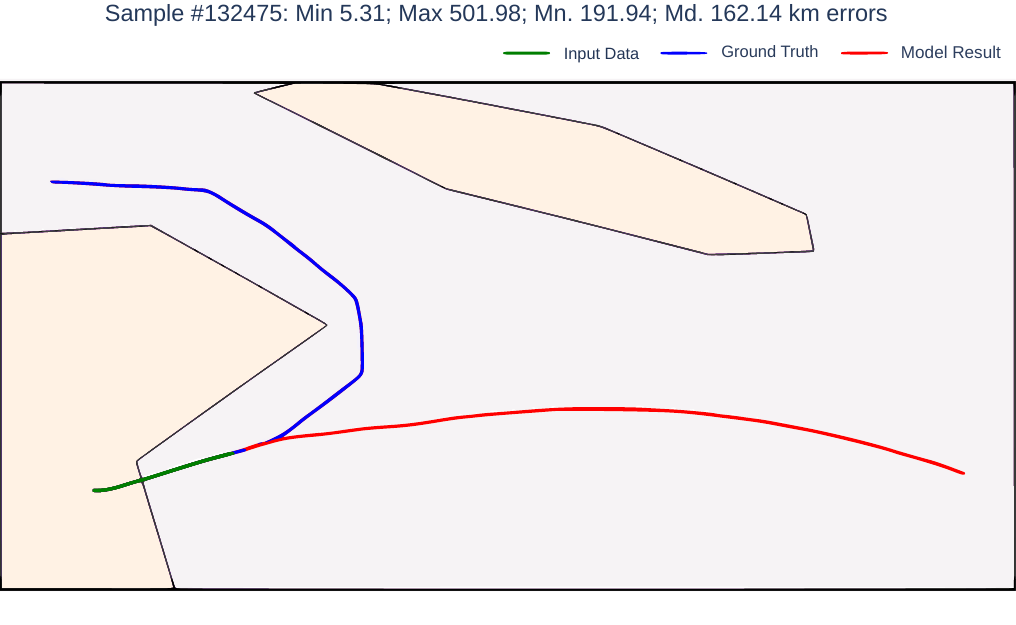}
        & \includegraphics[width=0.33\linewidth,trim={0 .5cm 0 1.35cm},clip]{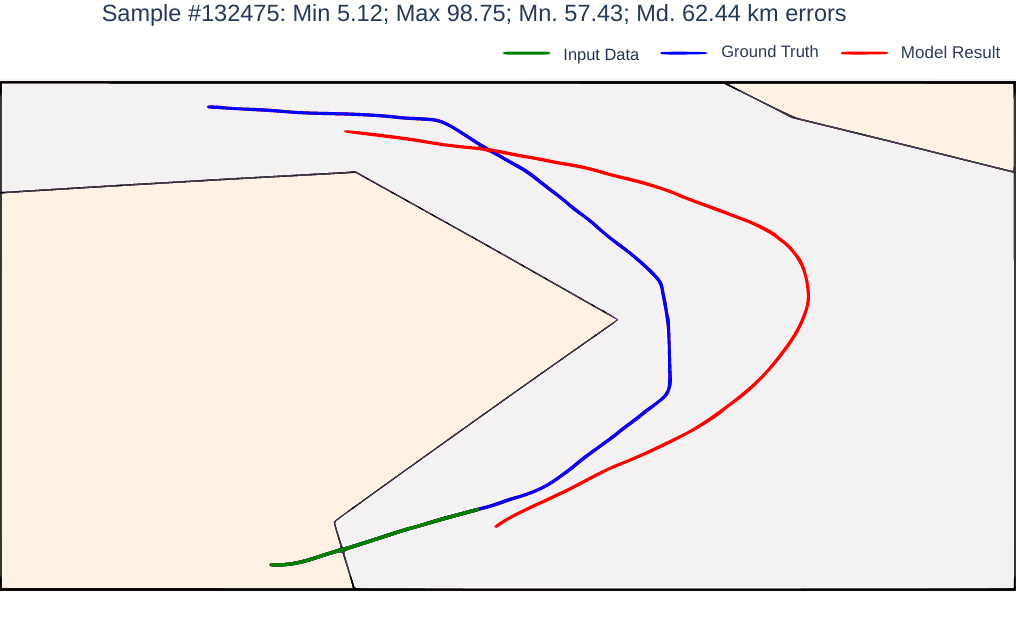}
        & \includegraphics[width=0.33\linewidth,trim={0 .5cm 0 1.35cm},clip]{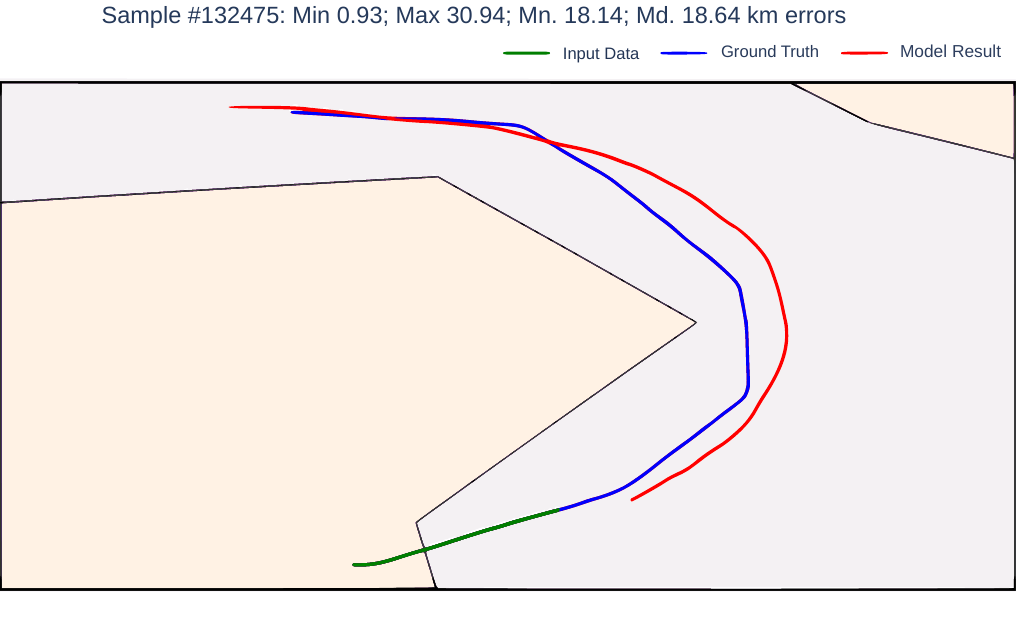}
        \vspace{-1.75cm} \\
        \rotatebox[origin=c]{90}{\hspace{2.75cm}\textit{Line \#5}}\hspace{-.4cm}
        & \includegraphics[width=0.33\linewidth,trim={0 .5cm 0 1.35cm},clip]{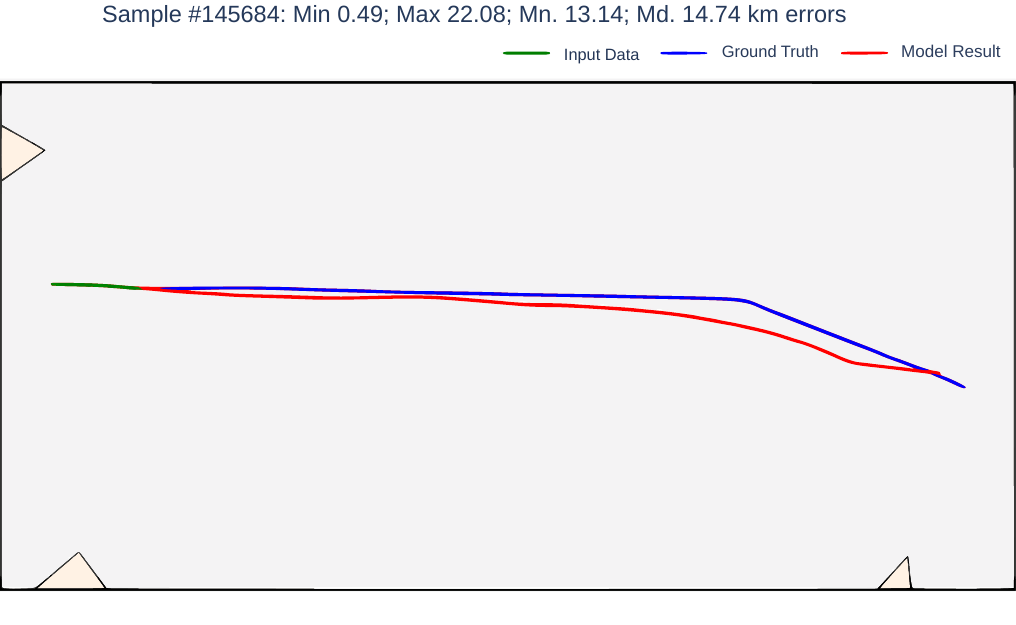}
        & \includegraphics[width=0.33\linewidth,trim={0 .5cm 0 1.35cm},clip]{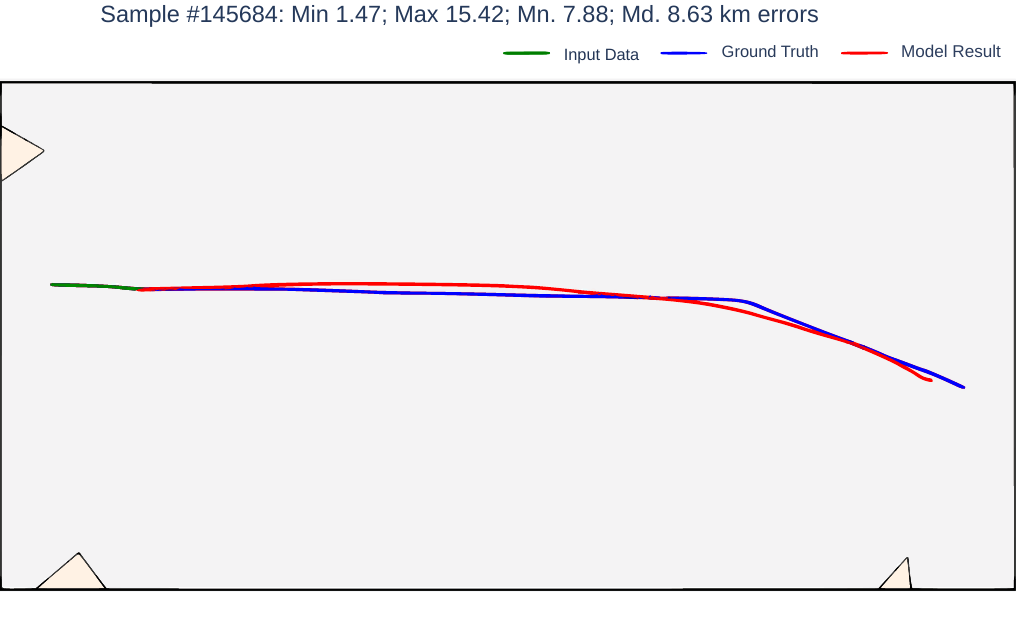}
        & \includegraphics[width=0.33\linewidth,trim={0 .5cm 0 1.35cm},clip]{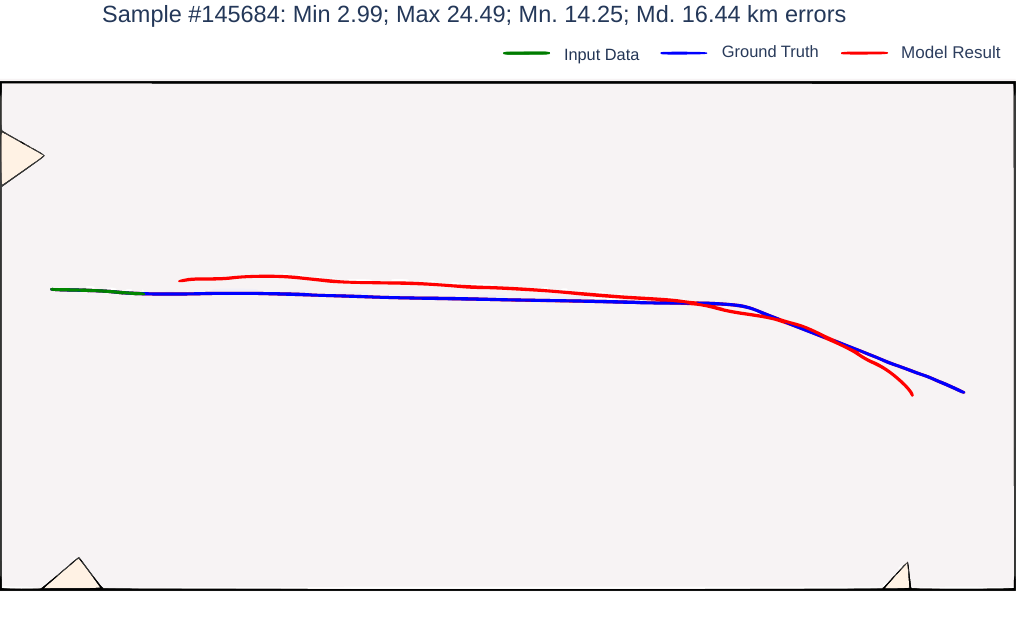}
        \vspace{-1.75cm} \\
    \end{tabular}
    \caption{Proposed Model \textbf{with} \textit{Convolutional Layers} \textbf{and without} \textit{Positional-Aware Attention} ({\bf C}2).\\\textbf{Legend:} \textcolor{green!40!black}{\textbf{Green}} lines represent the input data, \textcolor{blue!60!black}{\textbf{Blue}} the ground-truth, and \textcolor{red!80!black}{\textbf{Red}} the model forecasting.}
    \label{fig:dnn-no-cnn}
\end{figure}

\begin{figure}[!t]
    \vspace{-1.25cm}\hspace{-.65cm}
    \begin{tabular}{@{}c@{\hspace{0.5cm}}c@{\hspace{0.15cm}}c@{\hspace{0.15cm}}c@{}}
    & \subcaptionbox*{\textit{(a) Standard Features}}[0.3\linewidth][c]{} 
    & \subcaptionbox*{\textit{(b) Probabilistic Features}}[0.3\linewidth][c]{}
    & \subcaptionbox*{\textit{(c) Trigonometrical Features}}[0.3\linewidth][c]{} \\
        \rotatebox[origin=c]{90}{\hspace{2.75cm}\textit{Line \#1}}\hspace{-.4cm}
        & \includegraphics[width=0.33\linewidth,trim={0 .5cm 0 1.35cm},clip]{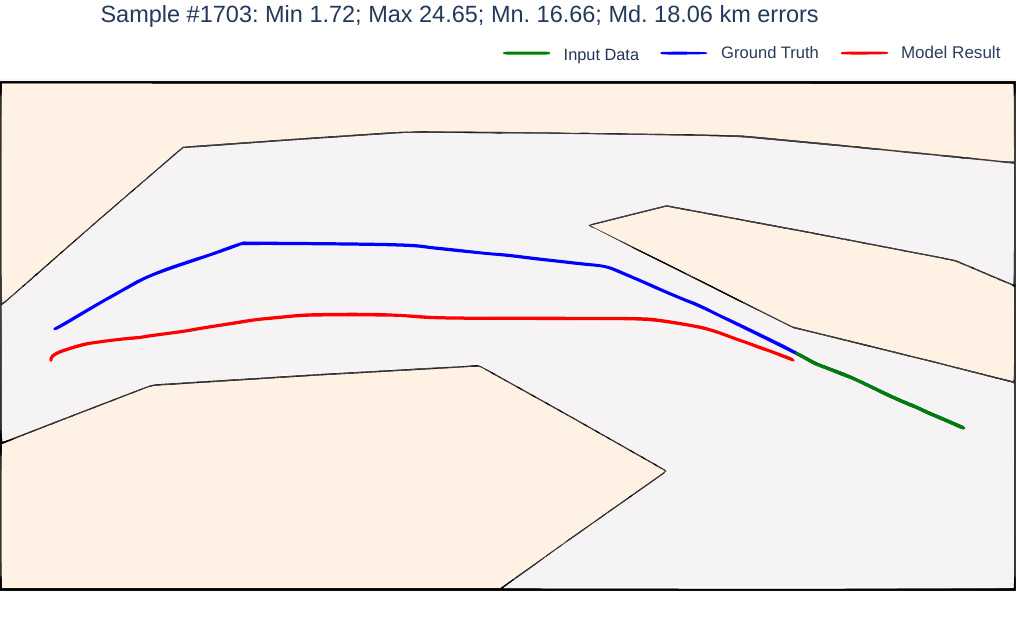}
        & \includegraphics[width=0.33\linewidth,trim={0 .5cm 0 1.35cm},clip]{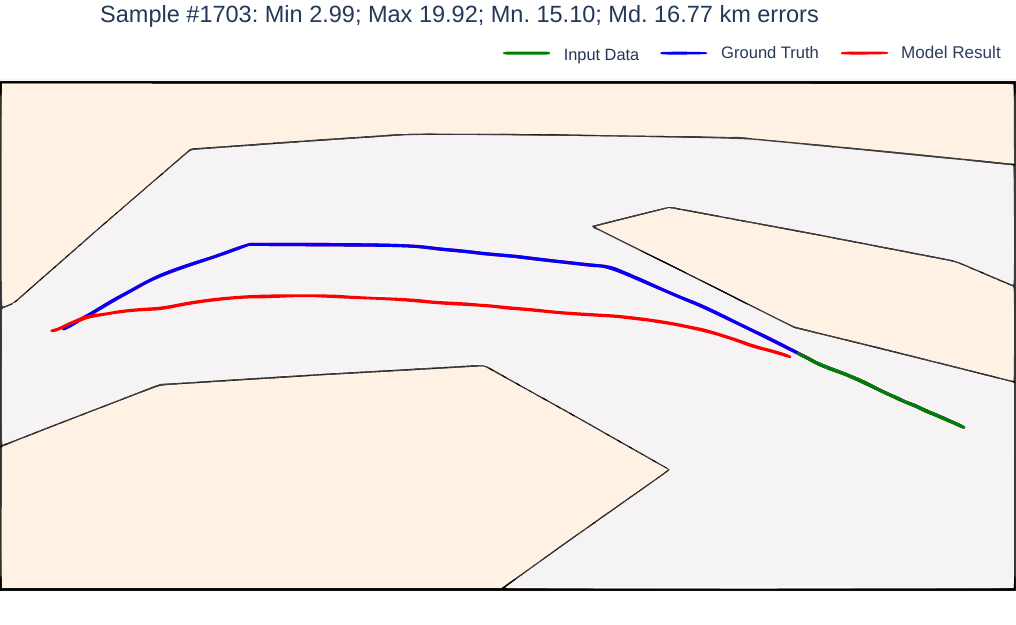}
        & \includegraphics[width=0.33\linewidth,trim={0 .5cm 0 1.35cm},clip]{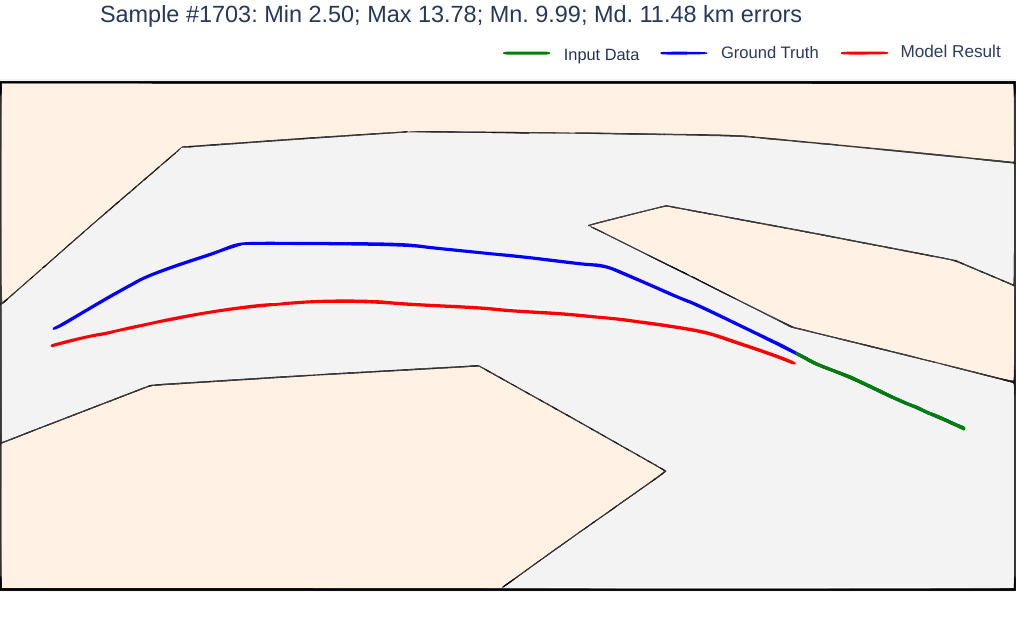}
        \vspace{-1.75cm} \\
        \rotatebox[origin=c]{90}{\hspace{2.75cm}\textit{Line \#2}}\hspace{-.4cm}
        & \includegraphics[width=0.33\linewidth,trim={0 .5cm 0 1.35cm},clip]{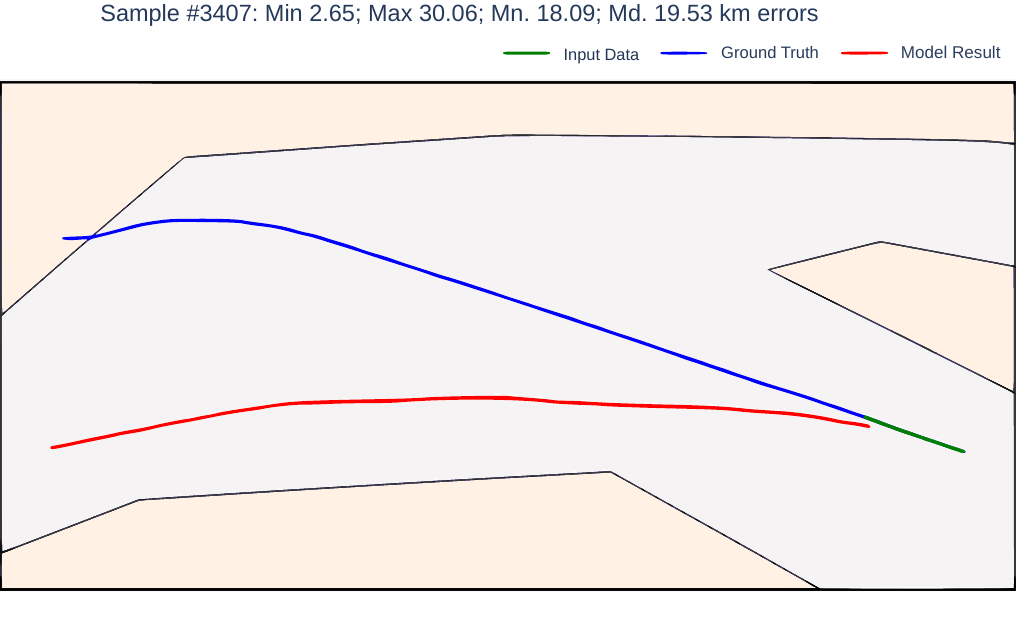}
        & \includegraphics[width=0.33\linewidth,trim={0 .5cm 0 1.35cm},clip]{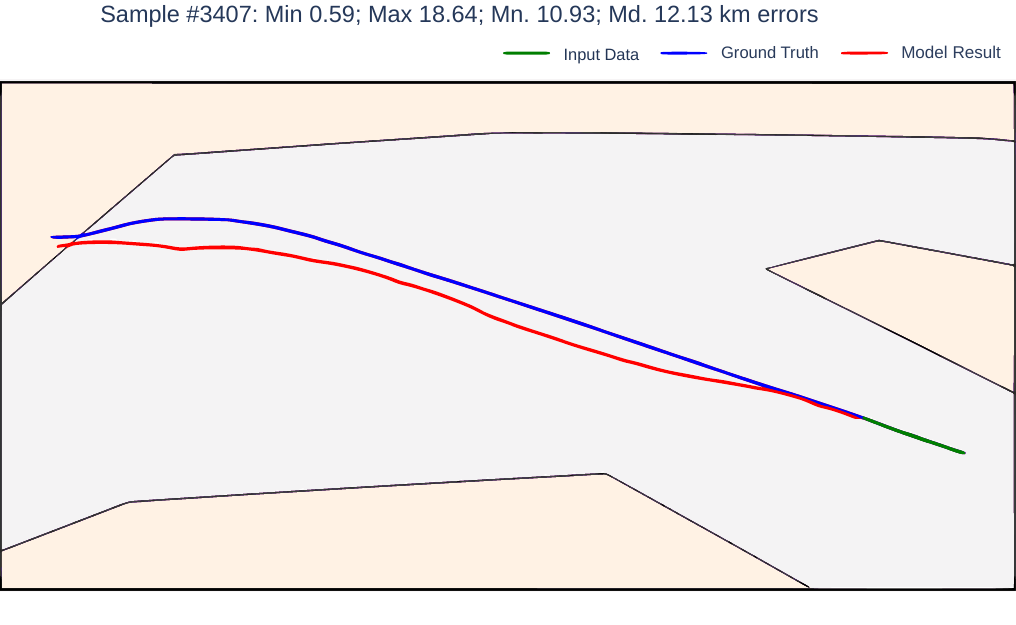}
        & \includegraphics[width=0.33\linewidth,trim={0 .5cm 0 1.35cm},clip]{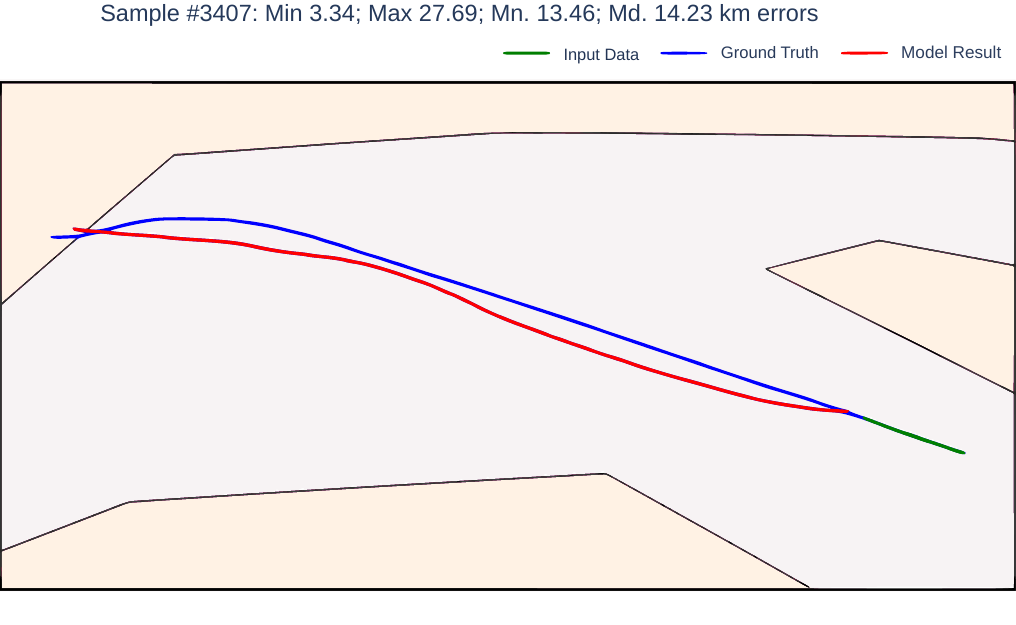}
        \vspace{-1.75cm} \\
        \rotatebox[origin=c]{90}{\hspace{2.75cm}\textit{Line \#3}}\hspace{-.4cm}
        & \includegraphics[width=0.33\linewidth,trim={0 .5cm 0 1.35cm},clip]{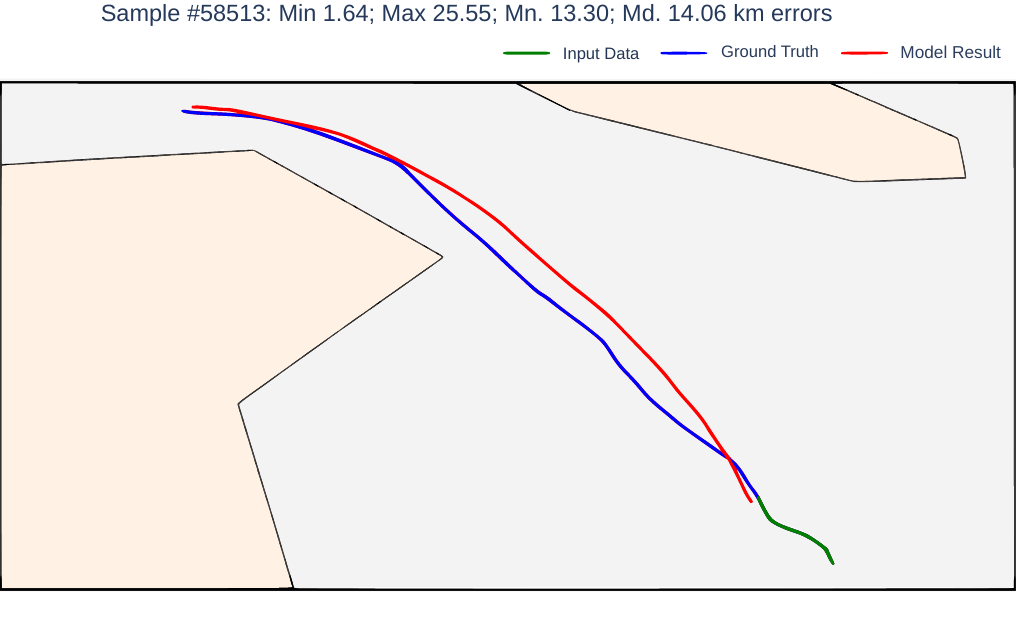}
        & \includegraphics[width=0.33\linewidth,trim={0 .5cm 0 1.35cm},clip]{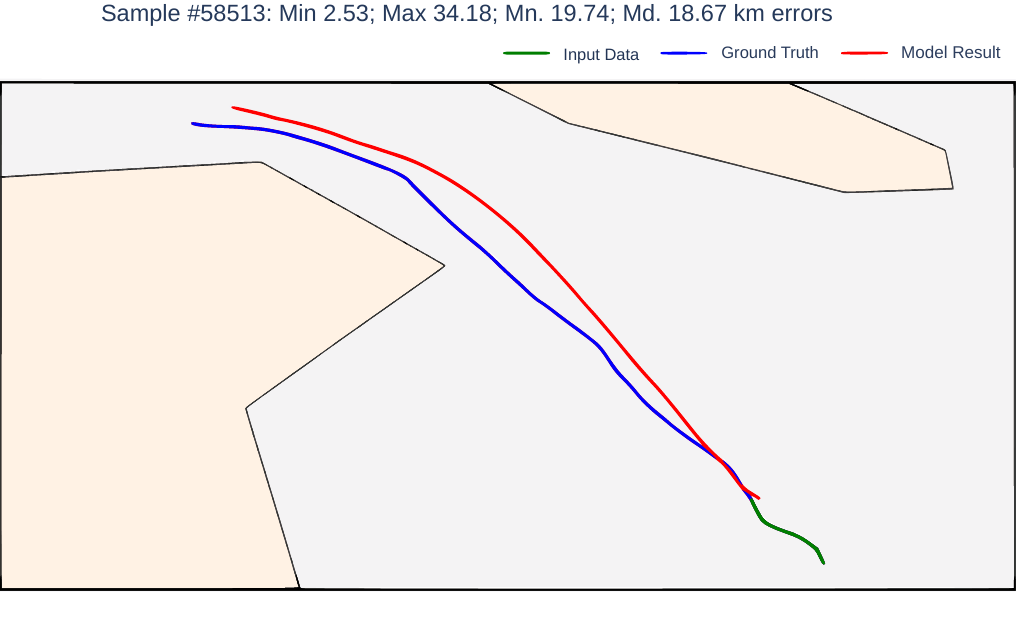}
        & \includegraphics[width=0.33\linewidth,trim={0 .5cm 0 1.35cm},clip]{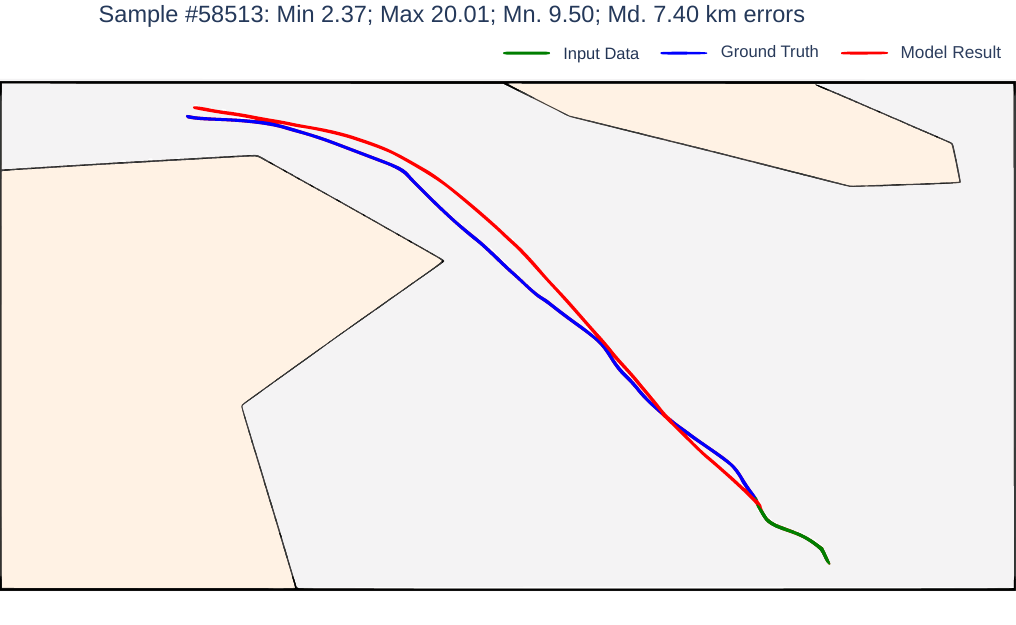}
        \vspace{-1.75cm} \\
        \rotatebox[origin=c]{90}{\hspace{2.75cm}\textit{Line \#4}}\hspace{-.4cm}
        & \includegraphics[width=0.33\linewidth,trim={0 .5cm 0 1.35cm},clip]{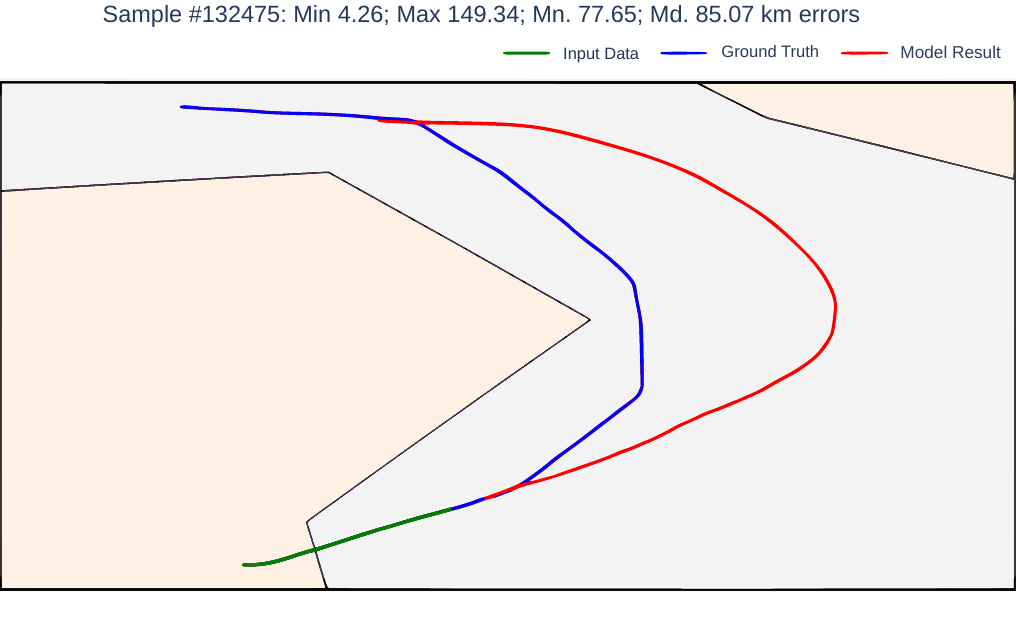}
        & \includegraphics[width=0.33\linewidth,trim={0 .5cm 0 1.35cm},clip]{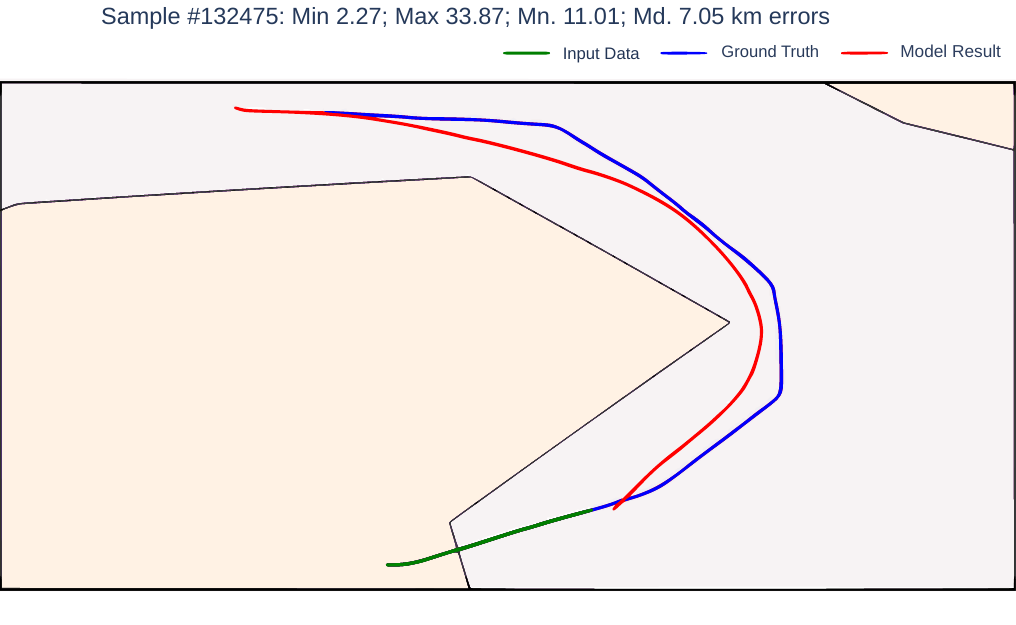}
        & \includegraphics[width=0.33\linewidth,trim={0 .5cm 0 1.35cm},clip]{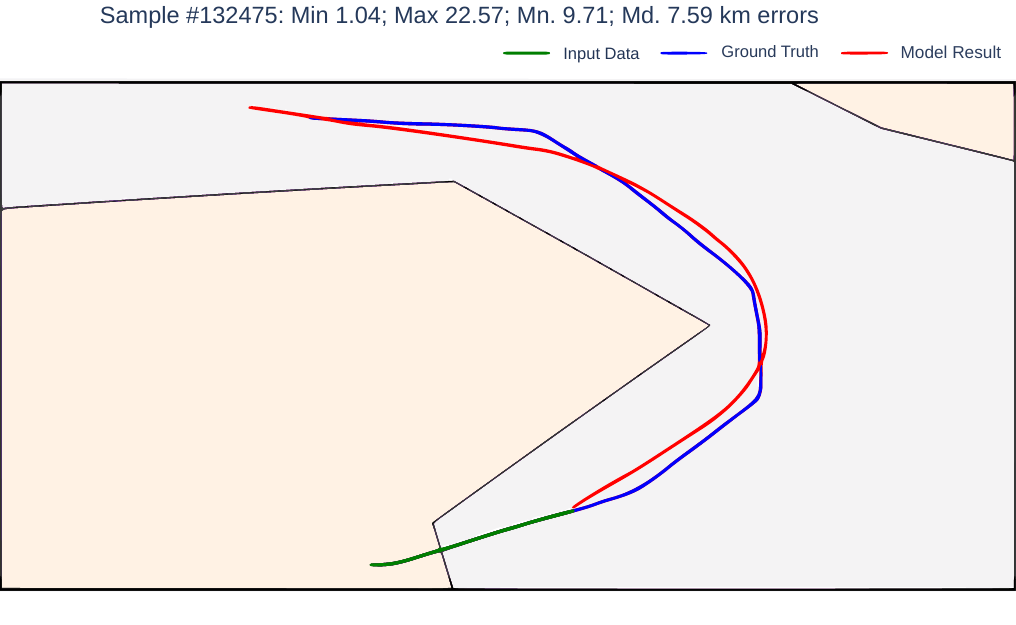}
        \vspace{-1.75cm} \\
        \rotatebox[origin=c]{90}{\hspace{2.75cm}\textit{Line \#5}}\hspace{-.4cm}
        & \includegraphics[width=0.33\linewidth,trim={0 .5cm 0 1.35cm},clip]{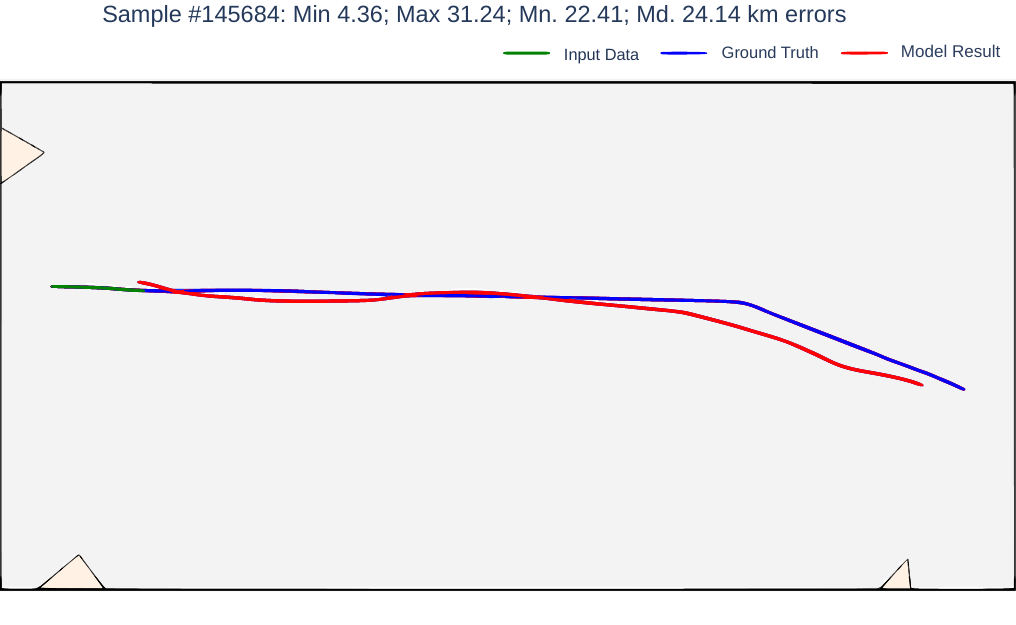}
        & \includegraphics[width=0.33\linewidth,trim={0 .5cm 0 1.35cm},clip]{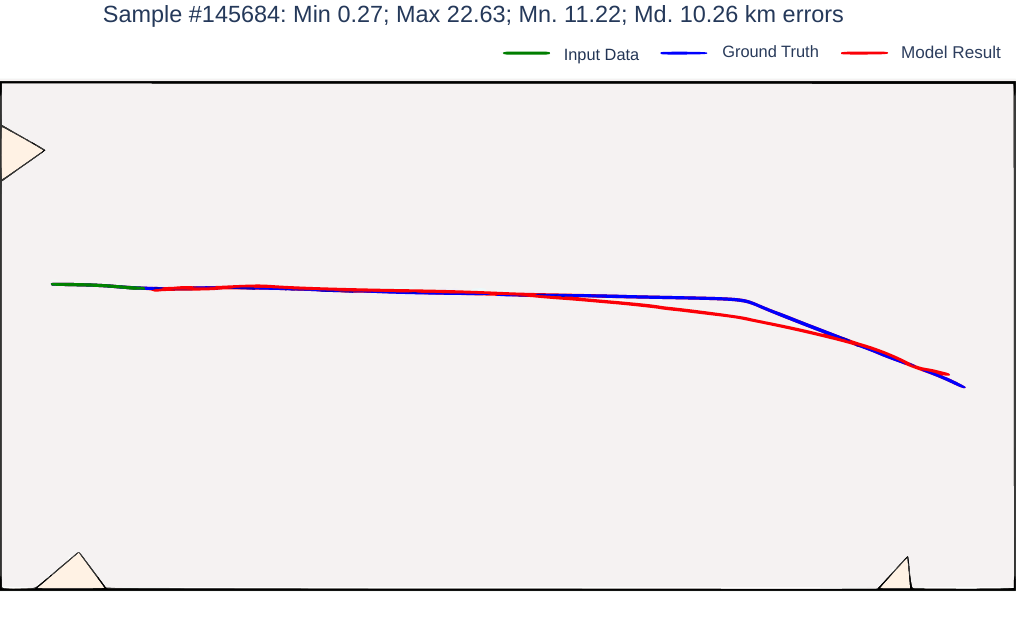}
        & \includegraphics[width=0.33\linewidth,trim={0 .5cm 0 1.35cm},clip]{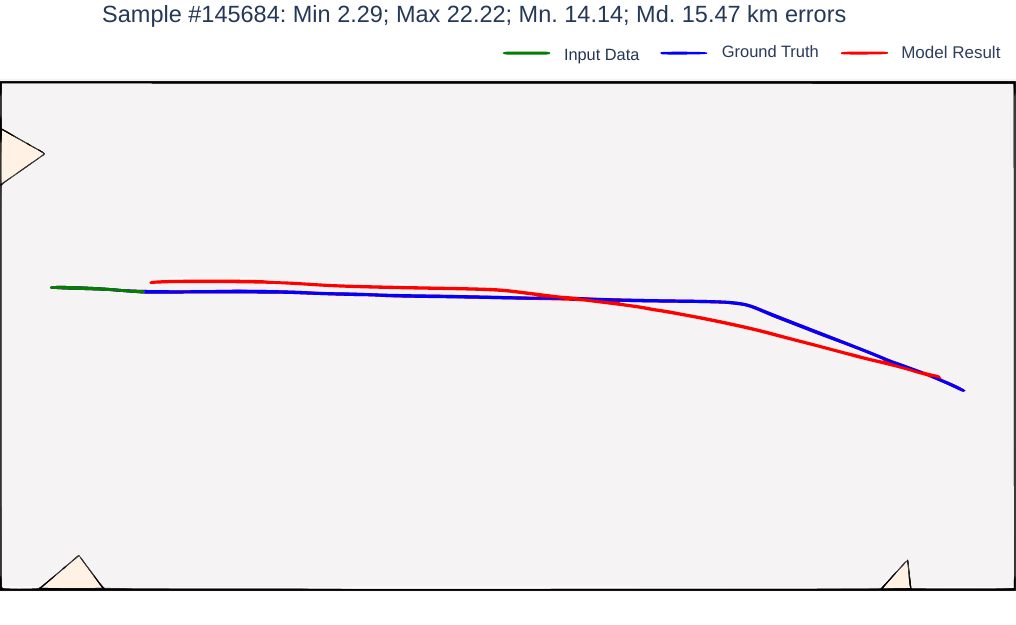}
        \vspace{-1.75cm} \\
    \end{tabular}
    \caption{Proposed Model \textbf{without} \textit{Convolutional Layers} \textbf{and with} \textit{Positional-Aware Attention} ({\bf C}3).\\\textbf{Legend:} \textcolor{green!40!black}{\textbf{Green}} lines represent the input data, \textcolor{blue!60!black}{\textbf{Blue}} the ground-truth, and \textcolor{red!80!black}{\textbf{Red}} the model forecasting.}
    \label{fig:dnn-no-attention}
\end{figure}

\begin{figure}[!t]
    \vspace{-1.25cm}\hspace{-.65cm}
    \begin{tabular}{@{}c@{\hspace{0.5cm}}c@{\hspace{0.15cm}}c@{\hspace{0.15cm}}c@{}}
    & \subcaptionbox*{\textit{(a) Standard Features}}[0.3\linewidth][c]{} 
    & \subcaptionbox*{\textit{(b) Probabilistic Features}}[0.3\linewidth][c]{}
    & \subcaptionbox*{\textit{(c) Trigonometrical Features}}[0.3\linewidth][c]{} \\
        \rotatebox[origin=c]{90}{\hspace{2.75cm}\textit{Line \#1}}\hspace{-.4cm}
        & \includegraphics[width=0.33\linewidth,trim={0 .5cm 0 1.35cm},clip]{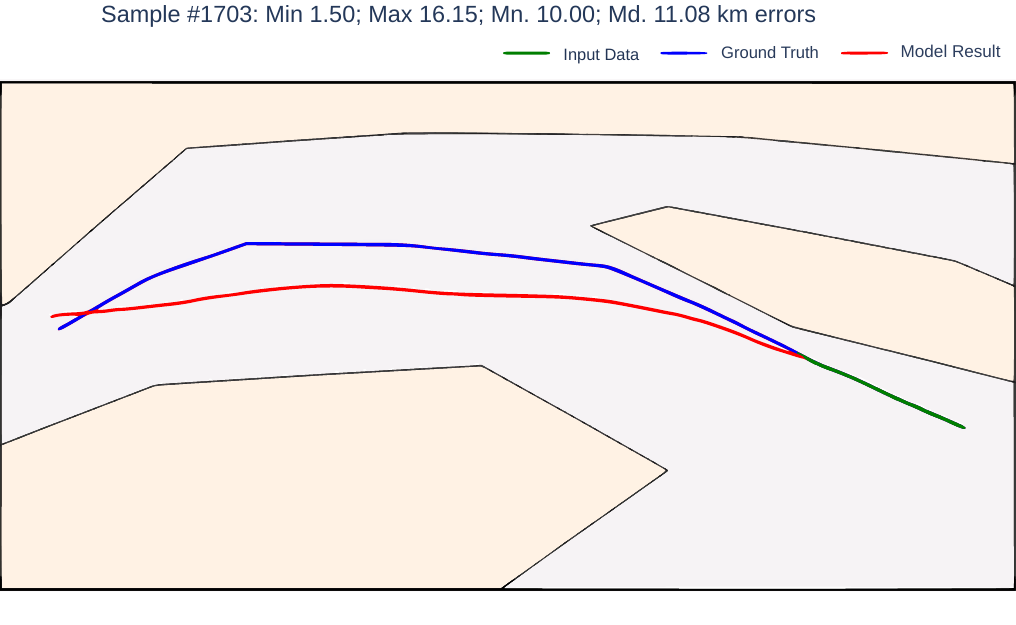}
        & \includegraphics[width=0.33\linewidth,trim={0 .5cm 0 1.35cm},clip]{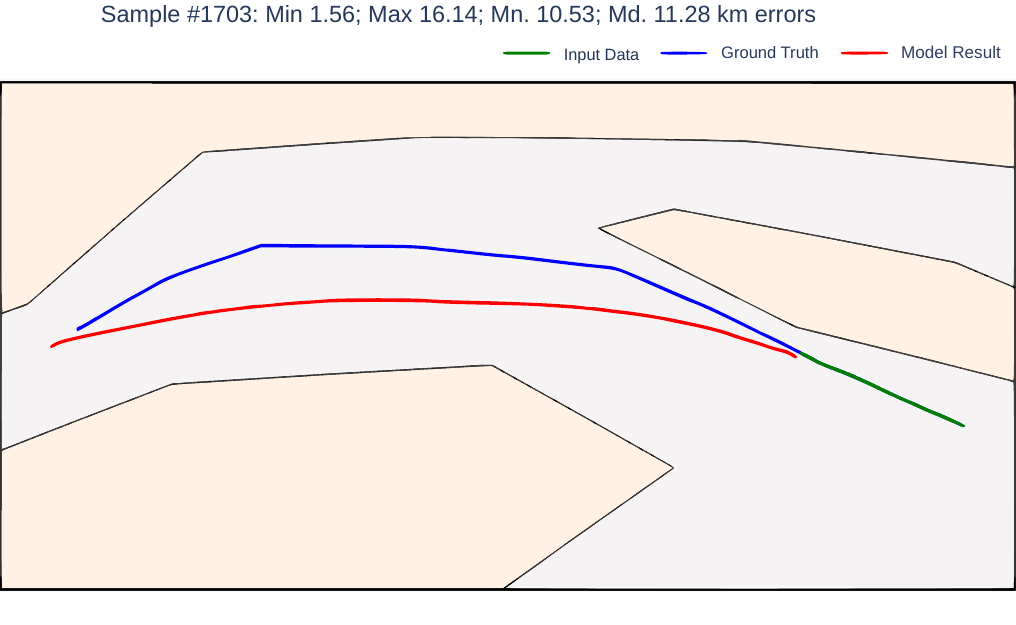}
        & \includegraphics[width=0.33\linewidth,trim={0 .5cm 0 1.35cm},clip]{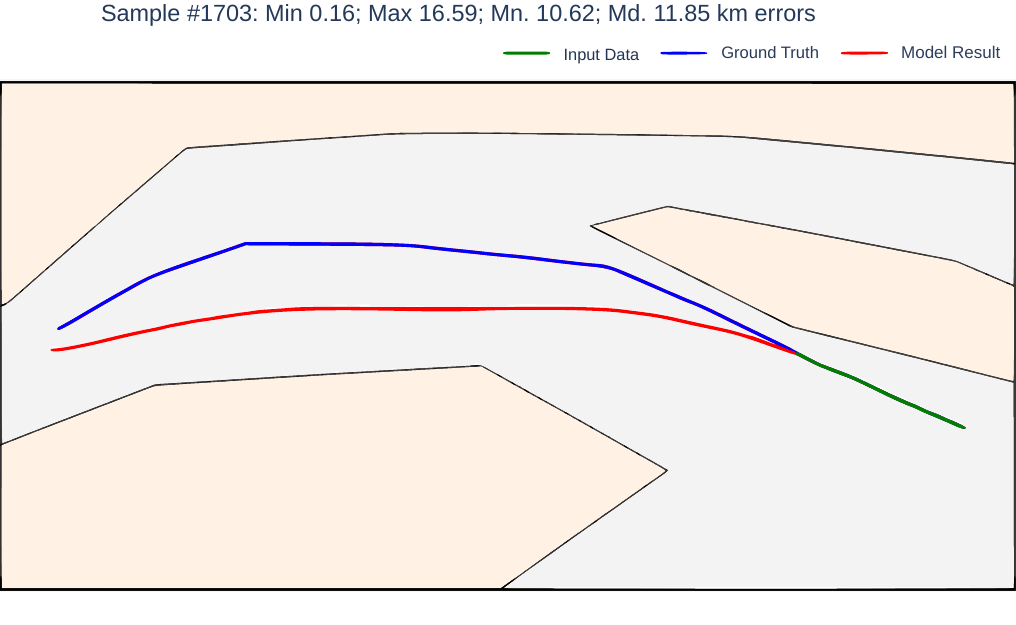}
        \vspace{-1.75cm} \\
        \rotatebox[origin=c]{90}{\hspace{2.75cm}\textit{Line \#2}}\hspace{-.4cm}
        & \includegraphics[width=0.33\linewidth,trim={0 .5cm 0 1.35cm},clip]{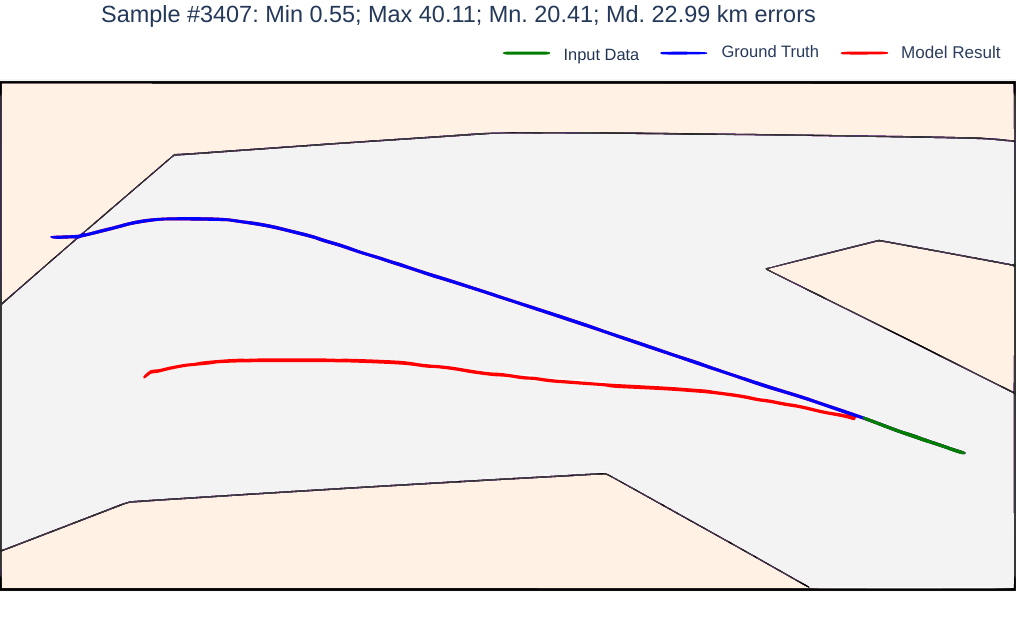}
        & \includegraphics[width=0.33\linewidth,trim={0 .5cm 0 1.35cm},clip]{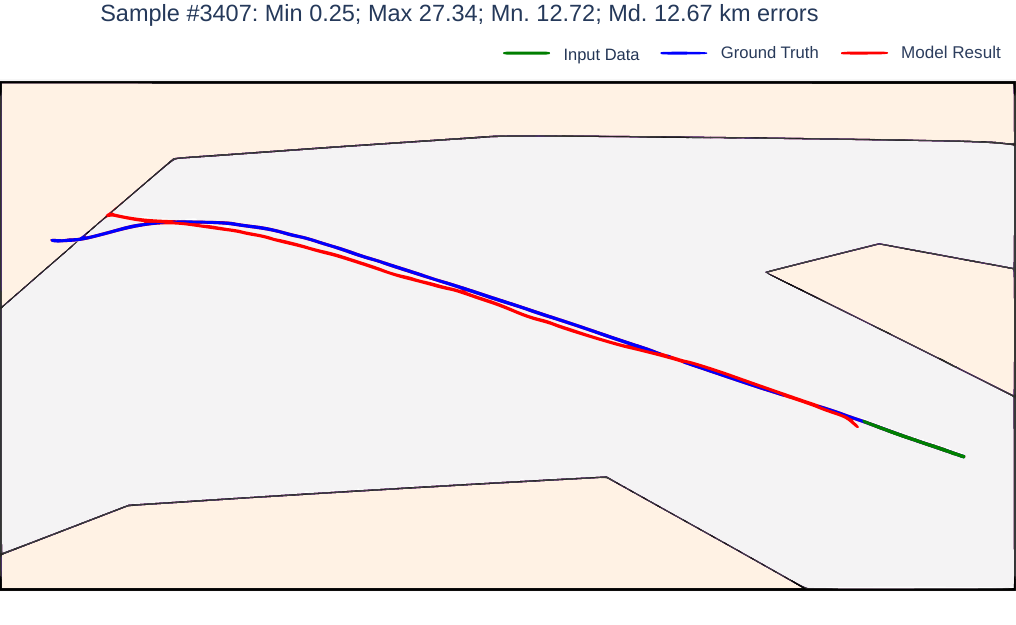}
        & \includegraphics[width=0.33\linewidth,trim={0 .5cm 0 1.35cm},clip]{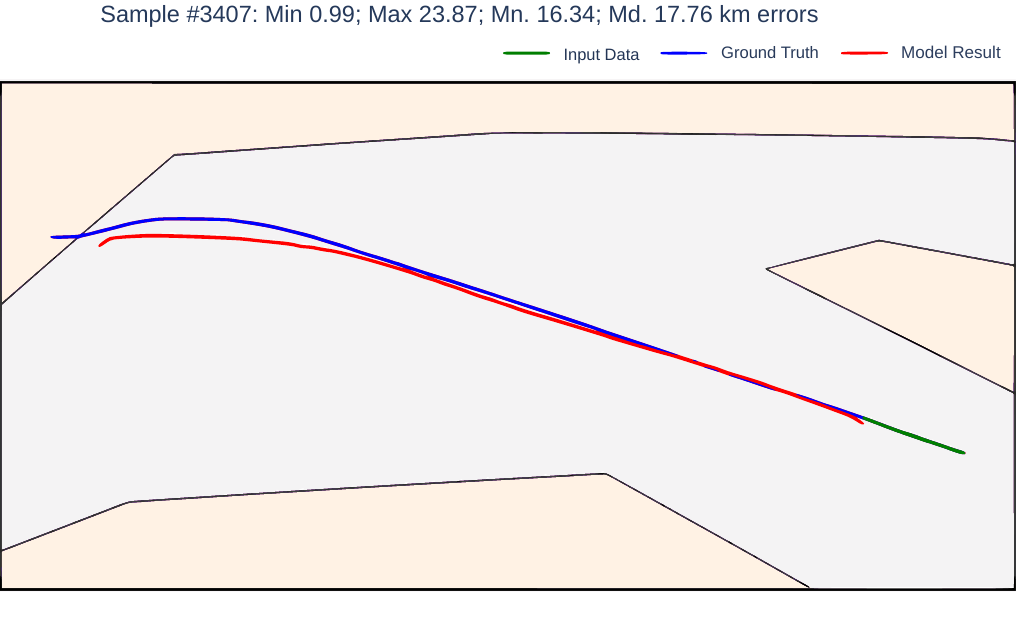}
        \vspace{-1.75cm} \\
        \rotatebox[origin=c]{90}{\hspace{2.75cm}\textit{Line \#3}}\hspace{-.4cm}
        & \includegraphics[width=0.33\linewidth,trim={0 .5cm 0 1.35cm},clip]{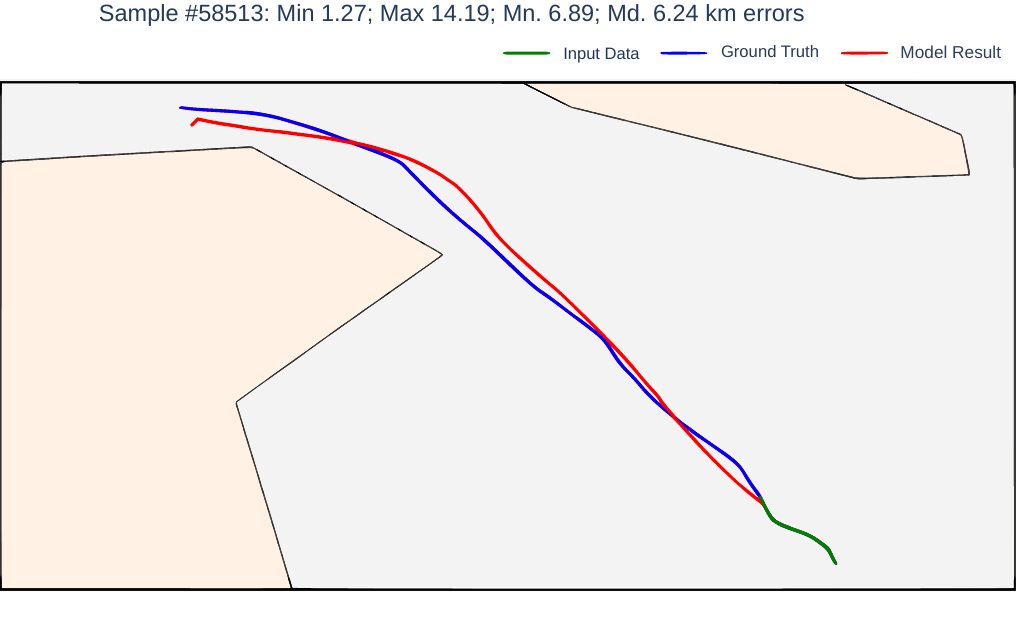}
        & \includegraphics[width=0.33\linewidth,trim={0 .5cm 0 1.35cm},clip]{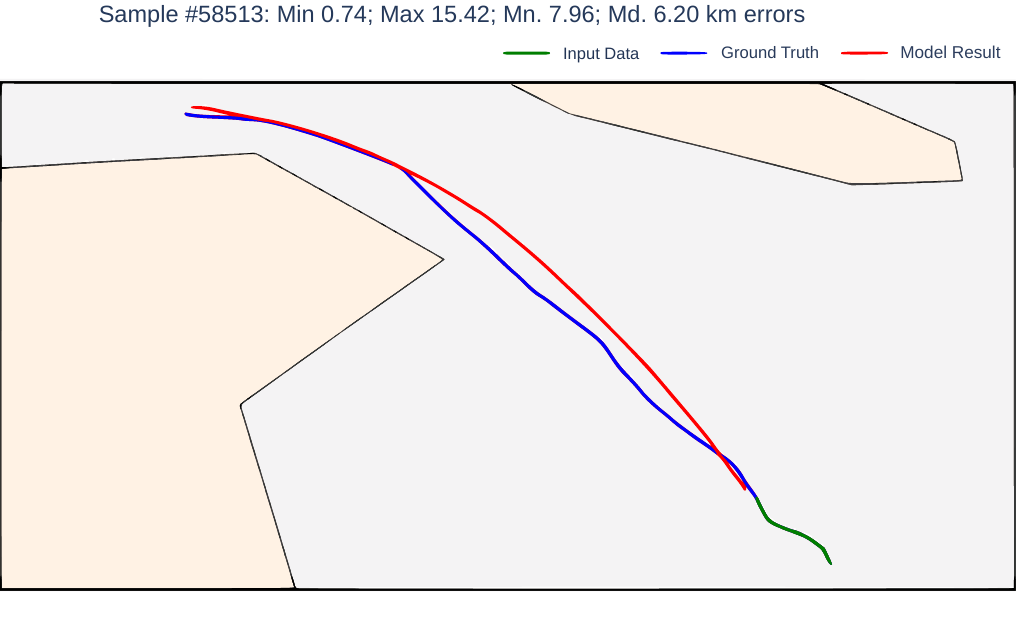}
        & \includegraphics[width=0.33\linewidth,trim={0 .5cm 0 1.35cm},clip]{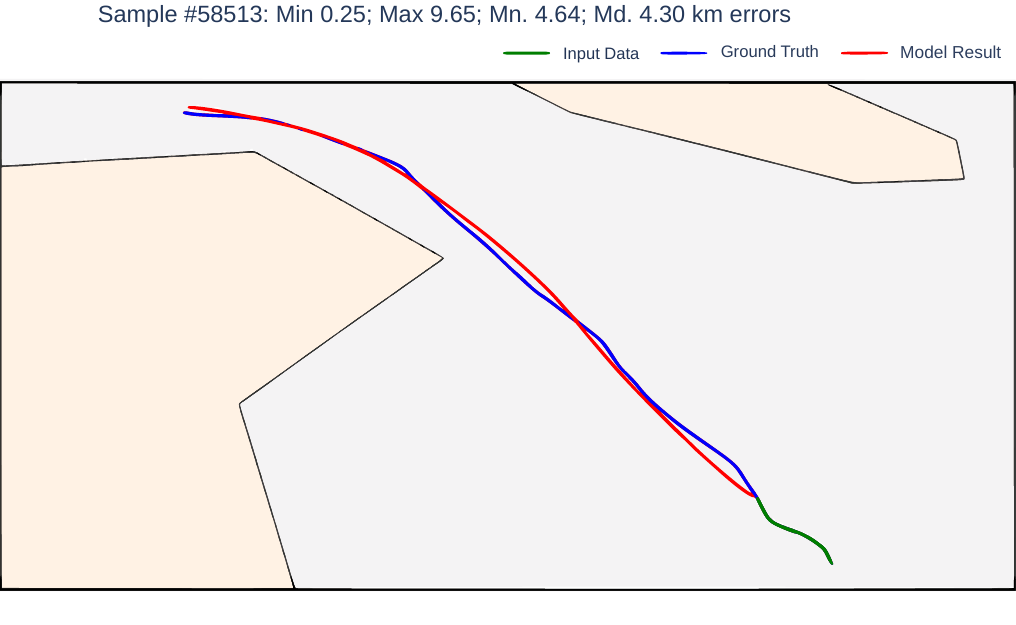}
        \vspace{-1.75cm} \\
        \rotatebox[origin=c]{90}{\hspace{2.75cm}\textit{Line \#4}}\hspace{-.4cm}
        & \includegraphics[width=0.33\linewidth,trim={0 .5cm 0 1.35cm},clip]{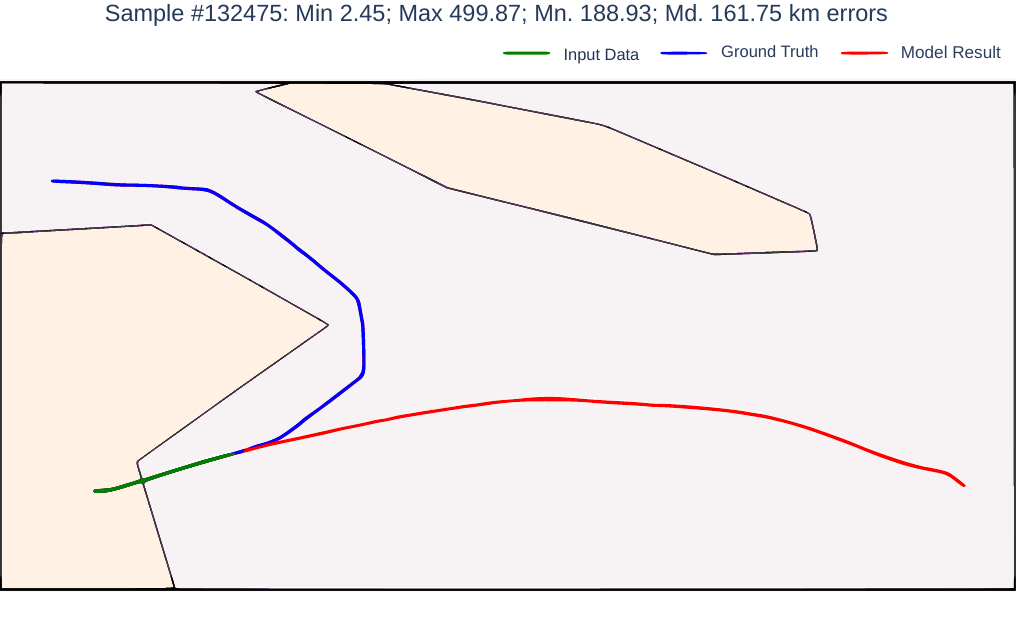}
        & \includegraphics[width=0.33\linewidth,trim={0 .5cm 0 1.35cm},clip]{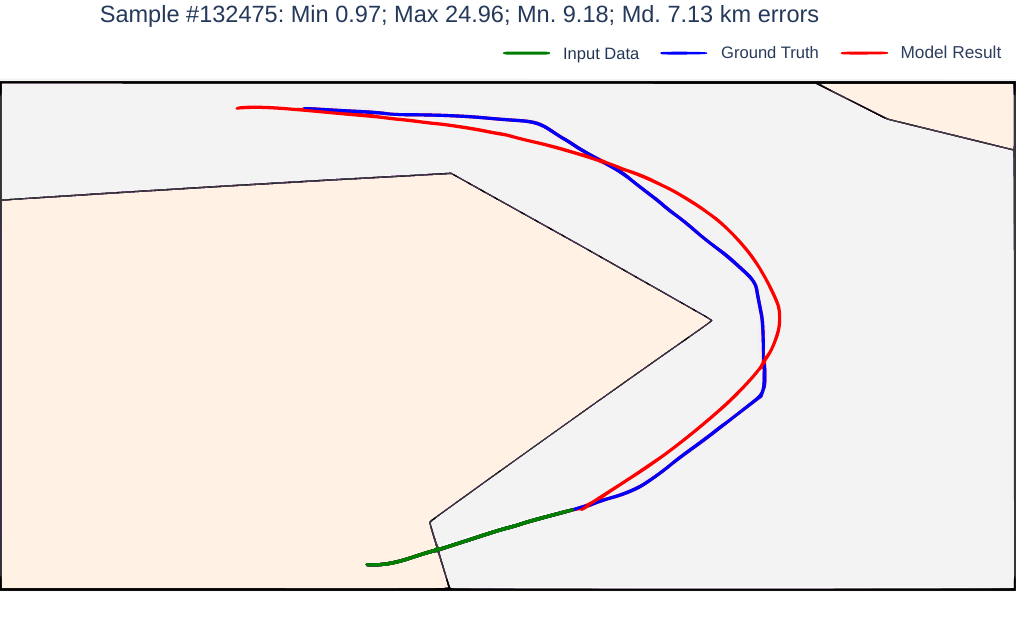}
        & \includegraphics[width=0.33\linewidth,trim={0 .5cm 0 1.35cm},clip]{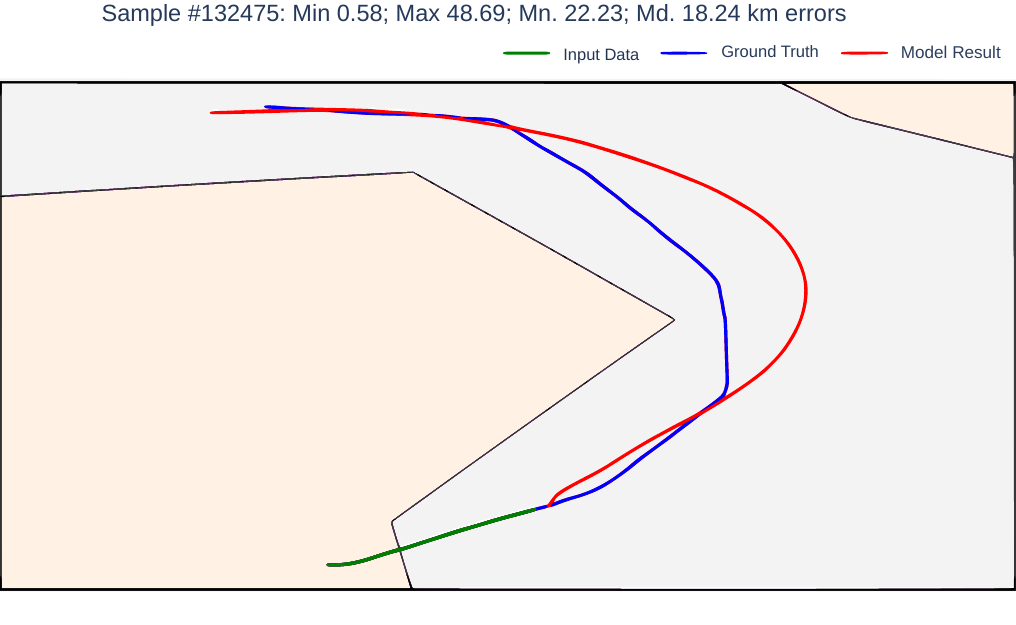}
        \vspace{-1.75cm} \\
        \rotatebox[origin=c]{90}{\hspace{2.75cm}\textit{Line \#5}}\hspace{-.4cm}
        & \includegraphics[width=0.33\linewidth,trim={0 .5cm 0 1.35cm},clip]{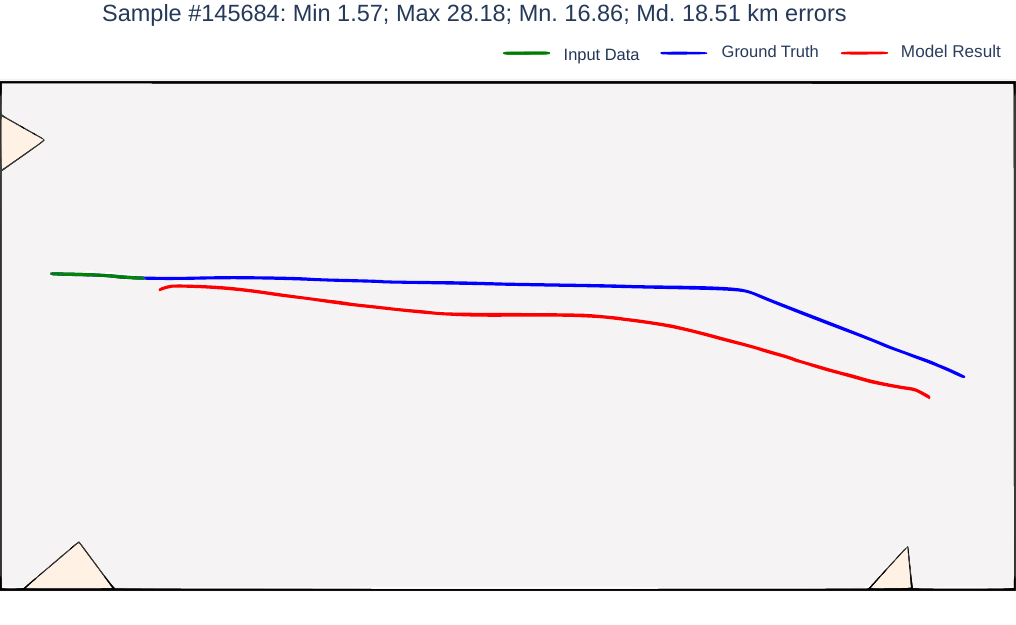}
        & \includegraphics[width=0.33\linewidth,trim={0 .5cm 0 1.35cm},clip]{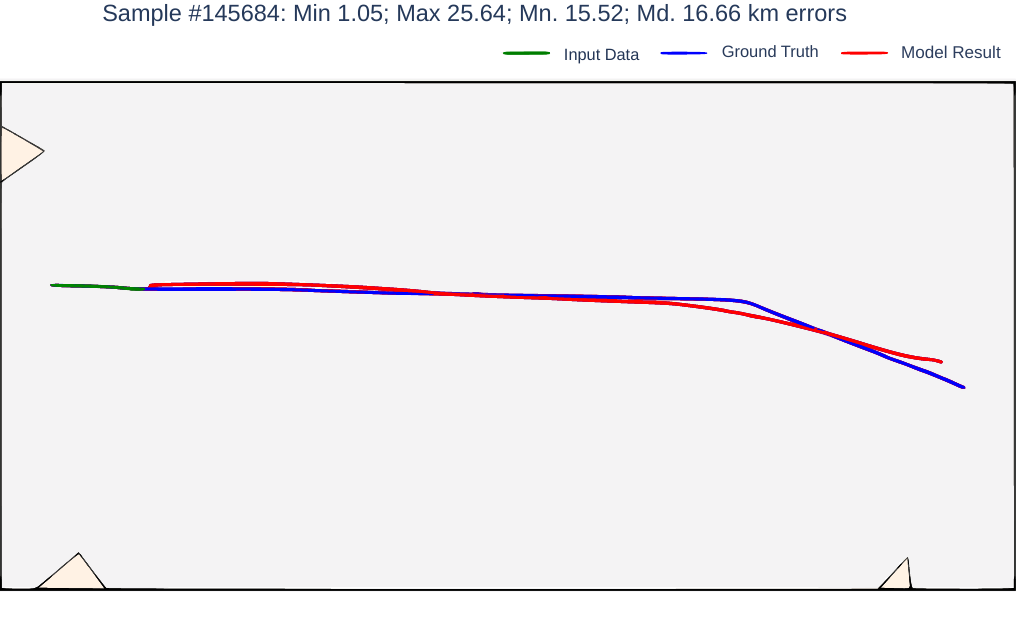}
        & \includegraphics[width=0.33\linewidth,trim={0 .5cm 0 1.35cm},clip]{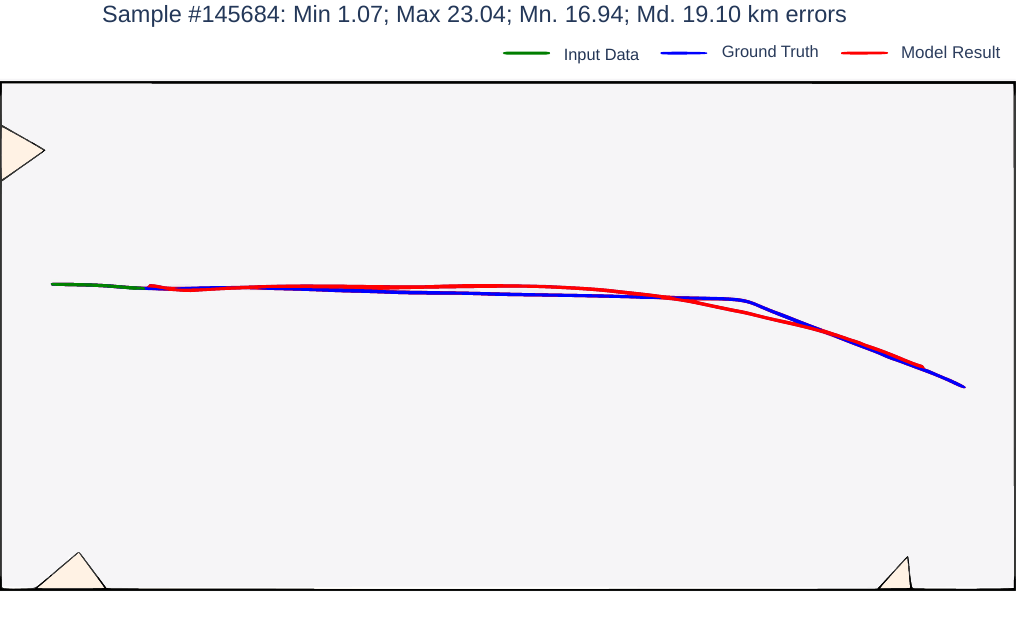}
        \vspace{-1.75cm} \\
    \end{tabular}
    \caption{Proposed Model \textbf{without} \textit{Convolutional Layers} \textbf{and} \textit{Positional-Aware Attention} ({\bf C}4).\\\textbf{Legend:} \textcolor{green!40!black}{\textbf{Green}} lines represent the input data, \textcolor{blue!60!black}{\textbf{Blue}} the ground-truth, and \textcolor{red!80!black}{\textbf{Red}} the model forecasting.}
    \label{fig:dnn-nothing}
\end{figure}

As further experiments, we designed and tested five different scenarios, which are illustrated in Figures~\ref{fig:dnn-complete}, \ref{fig:dnn-no-cnn}, \ref{fig:dnn-no-attention}, and \ref{fig:dnn-nothing}.
These scenarios aim to compare and validate the decision-making capability of different models and highlight that the evaluation metrics used for numerical analysis can be misleading in certain cases.
Figure~\ref{fig:dnn-complete} shows the results for model {\bf C}1,
Figure~\ref{fig:dnn-no-cnn} corresponds to {\bf C}2,
Figure~\ref{fig:dnn-no-attention} represents {\bf C}3,
while Figure~\ref{fig:dnn-nothing} displays the model {\bf C}4.
As for model {\bf C}5, we did not provide a visual representation due to its limited performance.
However, the detailed results for this model are available in Tables~\ref{tab:forecasting-cargos} and~\ref{tab:forecasting-tankers}.

In Figure~\ref{fig:dnn-complete}, we can observe the performance of the proposed model ({\bf C}1) on three sets of distinct features.
The image highlights two incorrect path decisions made when using the Standard Features, which are images \textbf{\textit{2(a)}} and \textbf{\textit{4(a)}}.
Additionally, it shows poor forecasting performance in image \textbf{\textit{5(a)}}.
However, the other two feature sets exhibit perfect path decisions, with clear variations in the approximation of the forecasted path with the observed ones.
In this case, the Trigonometrical Features are better suited for approximating the track in image \textbf{\textit{4(c)}} than the Probabilistic Features as seen in image \textbf{\textit{4(b)}}.
In contrast, the Probabilistic Features are better for image \textbf{\textit{4(b)}}, albeit with a broader curve shape that increases the error.

In Figure~\ref{fig:dnn-no-cnn}, the proposed model is shown without the CNN layers ({\bf C}2).
It is evident that the lack of these layers for spatial feature extraction produces a more noisy approximation of several paths.
Specifically, the \textbf{\textit{1(b)}}/\textbf{\textit{1(c)}} and \textbf{\textit{2(b)}}/\textbf{\textit{2(c)}} paths from the Probabilistic and Trigonometrical Features are poorly fitted.
The previously well-fitted, curvy-shaped forecast for the Probabilistic Features is now loosely related to the input trajectory and the expected forecasting, as seen in image \textbf{\textit{4(b)}}.
On the other hand, the Trigonometrical Features show a better curve fitting in image \textbf{\textit{5(c)}}, yet with highly imperfect forecastings due to the loss of spatial features sourced from the stacked convolutional layers.
Moreover, the standard feature set still struggles to forecast the correct paths.
Although there were minor improvements in images \textbf{\textit{1(a)}} and \textbf{\textit{5(a)}}, the curvy-shaped \textbf{\textit{4(a)}} and the port-entering one \textbf{\textit{2(a)}} remain incorrect.

The image in Figure~\ref{fig:dnn-no-attention} shows a slight change in the setup, where there is no attention mechanism, but instead, the stack of convolution layers ({\bf C}3) is used.
The image also reveals that all three sets of features fail equally to fit images \textbf{\textit{1(a)}}, \textbf{\textit{1(b)}}, and \textbf{\textit{1(c)}}.
This suggests that the model might repeat a dominant pattern in the dataset, failing to account for changes in the features required to predict the correct trajectory.
On the other hand, images \textbf{\textit{2(b)}} and \textbf{\textit{2(c)}}, which depict a port entry, are correctly predicted for both Probabilistic and Trigonometrical features.
However, for the standard feature set in image \textbf{\textit{2(a)}}, it shows the worst result seen so far.
Interestingly, the Trigonometrical Features are extremely well-fitted to the curvy-shape pattern, reasonably fitted by the Probabilistic Features, and somewhat fitted by the Standard Features, tracks shown in images \textbf{\textit{4(c)}}, \textbf{\textit{4(b)}}, and \textbf{\textit{4(a)}}, respectively.
This indicates that spatial patterns play a larger role in this set of experiments.
Still, it can lead to an overfitting scenario as CNNs tend to repeat dominant patterns from the training dataset.
In the complete model with all layers intact ({\bf C}1), the attention mechanism functions as a filter for the spatial features, helping to control overfitting in the forecasting.

In Figure~\ref{fig:dnn-nothing}, the probabilistic features, as seen in image \textbf{\textit{4(b)}}, perform better in curvy-shaped forecasting when compared to the trigonometrical features, as seen in image \textbf{\textit{4(c)}}.
However, all three feature sets fail to provide accurate information for the forecasting of images \textbf{\textit{1(a)}}, \textbf{\textit{1(b)}}, and \textbf{\textit{1(c)}}.
Similarly, the standard features fail to forecast images \textbf{\textit{2(a)}} and \textbf{\textit{4(a)}}.
Although image \textbf{\textit{5(a)}} moves in the correct direction, it fails to fit accurately as the other two sets of features managed to achieve in images \textbf{\textit{5(b)}} and \textbf{\textit{5(c)}}.

Our set of experiments shows that the numerical variations in Tables~\ref{tab:forecasting-cargos} and \ref{tab:forecasting-tankers}  reflect the partial performance of the models in terms of long-term multi-path trajectory forecasting.
Such discrepancy arises as the assessment metrics often fail to capture the inherent decision-making complexity of the deep-learning models, especially in scenarios involving predominantly linear trajectories.
Even when the dataset is balanced and stratified, these metrics require additional inspection to fully appreciate the models' ability to handle complex tasks.
Moreover, the plethora of experiments has permitted us to identify the superior model for this study -- model {\bf C}1, trained on trigonometrical features.
This model is proficient in accurately forecasting and visually representing all possible trajectories.
In addition, the model performs better, yielding lower errors when trained on trigonometric features instead of probabilistic features.
Therefore, the {\bf C}1 model demonstrates superior path-decision capabilities, supported by the minimal approximation errors observed across the complete dataset of reserved vessels used for validation.

\begin{figure}[htbp]
    {\small{\textbf{Legend:} \textcolor{green!40!black}{\textbf{Green}} lines is the input data, \textcolor{blue!60!black}{\textbf{Blue}} the ground-truth, and \textcolor{red!80!black}{\textbf{Red}} the model forecasting.}}
    \centering
    \begin{subfigure}[b]{0.32\textwidth}
        \centering
        \includegraphics[width=\linewidth]{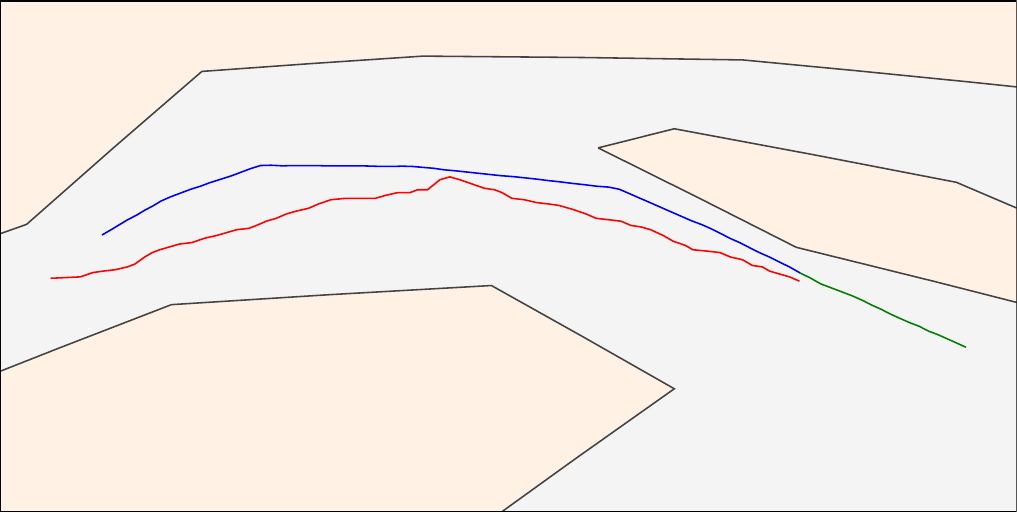}
        \caption{Test case \#1}
    \end{subfigure}
    \hfill
    \begin{subfigure}[b]{0.32\textwidth}
        \centering
        \includegraphics[width=\linewidth]{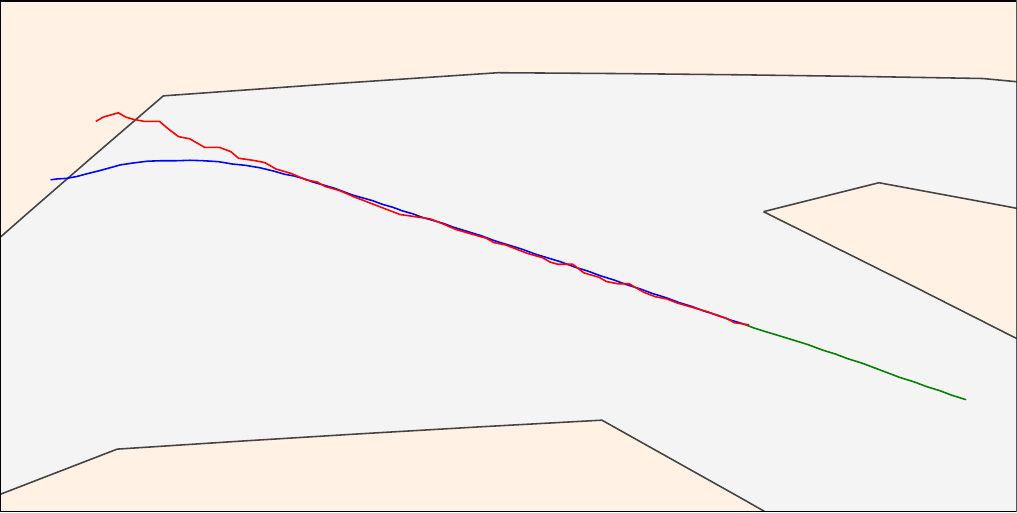}
        \caption{Test case \#2}
    \end{subfigure}
    \hfill
    \begin{subfigure}[b]{0.32\textwidth}
        \centering
        \includegraphics[width=\linewidth]{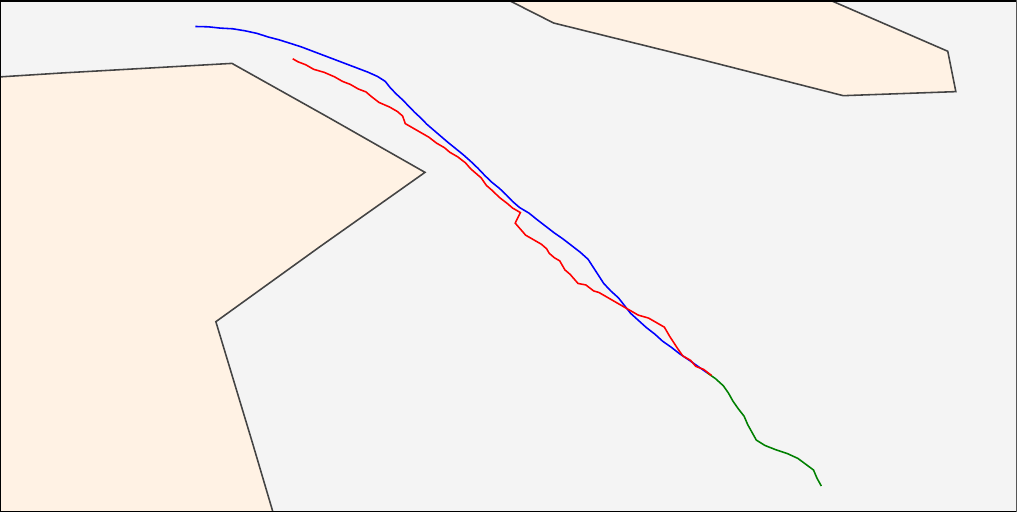}
        \caption{Test case \#3}
    \end{subfigure}
    {\centering
        \begin{subfigure}[b]{0.32\textwidth}
            \centering
            \includegraphics[width=\linewidth]{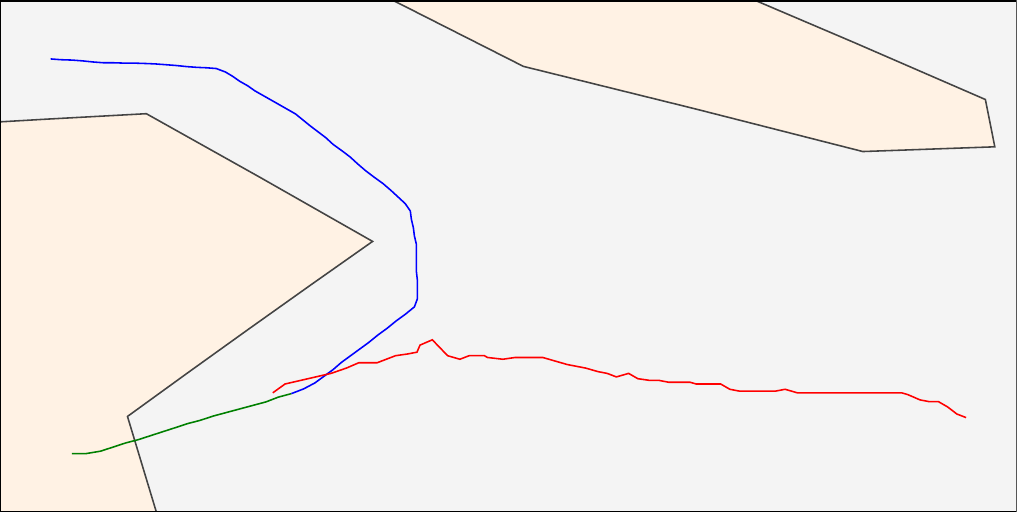}
            \caption{Test case \#4}
        \end{subfigure}
        \hspace{.15cm}
        \begin{subfigure}[b]{0.32\textwidth}
            \centering
            \includegraphics[width=\linewidth]{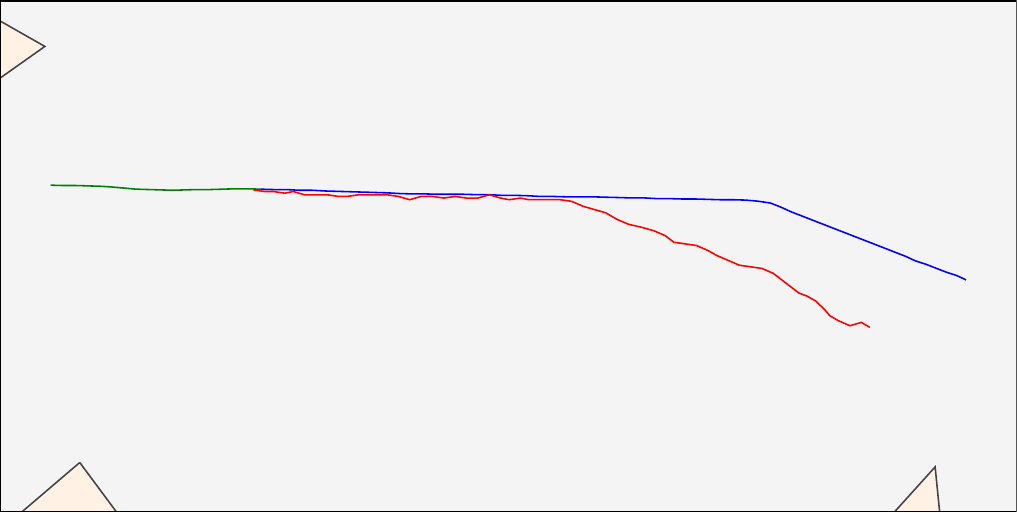}
            \caption{Test case \#5}
        \end{subfigure}
    }
    \caption{TrAISformer~\cite{nguyen2024} model over the five test cases using the {\it (a) Default Features} for comparison.}
    \label{fig:traisformer}
\end{figure}

For a finer and more complete evaluation, in Figure~\ref{fig:traisformer}, we show the results we achieved over the same test cases using the {\it (a) Default Features} with the TrAISformer model, the previous state-of-the-art on the task of long-term trajectory forecasting.
Several adaptations were required to make the model suitable for the task we approach in this paper.
Details about our adaptations to their model are available in the Methods.
As can be observed in the images, the TrAISformer shows a similar performance to a simple sequence-to-sequence auto-encoder model when not trained with the probabilistic and trigonometric features.
This means that their model, as is, is not capturing additional information that can help in the multi-path long-term trajectory forecasting problem.
Besides this, TrAISformer has $7.6\times10^7$ parameters, taking 1 hour and 30 minutes per training epoch and requiring 30 hours of training over the complete dataset.
In contrast, the models tested and presented in the current paper are limited to $1.5\times10^7$ parameters, \textit{five times fewer trainable parameters}, in most cases requiring half a million trainable parameters, taking not more than 15 minutes per epoch and 3 hours to achieve a fully trained neural network.
Details on the computational environment used for this experiment and computation can be found within the Methods.

Table~\ref{tab:forecasting-TrAISformer} presents the results obtained from the complete test dataset of cargo and tankers.
Unlike the previous models, TrAISformer shows an $R^{2}$ result lower than the previously observed 98\% for our proposed model.
This is because the model was trained using teacher-forcing, which forecasts navigational metrics instead of approximating them using the trajectory coordinates.
Therefore, the trajectories are geometrically noisy, even though some display correct direction and forecasted long-term paths.
Based on the outcomes shown in Table~\ref{tab:forecasting-cargos} and~\ref{tab:forecasting-tankers}, TrAISformer has achieved results similar to \textbf{A}1/\textbf{C}5 and \textbf{B}1/\textbf{C}5; they refer to simple unidirectional recurrent networks on an auto-encoder structure and do not rely on convolutional networks and the attention mechanism.
Based on these results, we can conclude that utilizing probabilistic augmentation for trajectory data, such as AIS data, and employing different spatial projections for feature learning in auto-encoder sequence-to-sequence models achieves state-of-the-art results in long-term multi-path-trajectory forecasting.
This approach proves to be the most effective strategy for handling uncertainty in open waters where there are multiple paths between origin and destination.

\begin{table}[!ht]
    \renewcommand{\arraystretch}{1.1}
    \centering
    \begin{tabular}{c@{\hspace{0.6em}}c@{\hspace{0.55em}}c@{\hspace{0.55em}}c@{\hspace{0.55em}}c@{\hspace{0.55em}}c@{\hspace{0.55em}}c@{\hspace{0.55em}}c@{\hspace{0.55em}}c@{\hspace{0.55em}}}\hline
        & $R^{2}$ Score & MAE & MSE & Mean Err. & 25$^{th}$ Pct. & 50$^{th}$ Pct. & 75$^{th}$ Pct. & Std. Dev. \\ \hline
        Tanker & 96.67\% & 0.0966 & 0.0399 & 16.1471 & 3.6852 & 9.0399 & 19.9771 & 24.2283\\
        Cargo & 97.23\% & 0.0945 & 0.0342 & 15.6350 & 3.5165 & 8.7594 & 19.6720 & 21.4226 \\ \hline
    \end{tabular}
    \caption{Results from TrAISformer~\cite{nguyen2024} over the entire dataset of Tanker and Cargo vessels.}
    \label{tab:forecasting-TrAISformer}
\end{table}

\section{Conclusion and Remarks}
\label{sec:conclusions}
Our study highlights the significance of incorporating a probabilistic model within the trajectory forecasting process to enhance multi-path long-term vessel trajectory forecasting.
During training, ground-truth data is used to construct transition probability matrices, providing the deep learning model with a nuanced perspective on potential vessel movements without the risk of overfitting.
As the model is probabilistically driven, it stores motion statistics without relying heavily on the exact details of training data.
This enables the model to distill a set of generalized patterns and rules from the data, which, while retaining a degree of uncertainty, does not compromise the model's prediction capabilities.
This controlled uncertainty enhances its ability to forecast under various conditions and adapt to the complexity of maritime navigation.

Our deep-learning model offers a unique advantage by combining exact coordinates from the Standard Features with conditional probabilities from the output of probabilistic model (route and endpoint grid cell), allowing for a blend of precision and flexibility in predictive capabilities.
This approach enables the model to identify intricate route patterns and make informed decisions that simpler baseline models often overlook.
Through the Probabilistic Features, the model learns to navigate a domain where certainty is balanced with possibilities, reflecting real-world conditions where numerous factors can alter a vessel's trajectory.
This dynamic interplay between certainty and uncertainty makes the deep learning model more robust and less likely to fail.

The dual-model architecture with shared predictive responsibilities strengthens the overall system's resilience, making it well-suited for real-time applications that process streaming AIS data.
In this sense, our research contributes to trajectory forecasting by introducing a spatio-temporal model that integrates probabilistic feature engineering with deep learning techniques, significantly improving routing decisions on long-term vessel trajectory forecasting.
We tested the model in various conditions with multiple feature sets and two vessel types, demonstrating its robustness and generalization in different contexts.

It is important to note that relying solely on numerical indicators can be misleading when deciding over complex scenarios.
As we showed in the results, our model efficiently makes correct long-term path decisions, though it may only sometimes generate as straight paths as baseline models due to its noisiness.
The ability of our model comes not only from the features but also the positional-aware attention mechanism in our deep-learning model, which significantly improves its ability to capture temporal dependencies, enabling it to learn nuanced temporal patterns that improve trajectory-fitting capabilities.

Furthermore, our study has practical implications concerning maritime safety and preserving biodiversity.
Improved trajectory forecasting can significantly enhance navigational safety and prevent unintentional harm to marine life, particularly endangered species inhabiting busy sea routes such as the NARW.
Our work inspires further research to improve the model's performance, such as adding more route polygons across the Gulf of St.
Lawrence and exploring techniques like Dynamic Time Warping as a loss function to reduce noise in generated trajectories.
Furthermore, our model offers a foundation to harness the potential of sophisticated forecasting models, advancing maritime safety and proactively addressing ecological preservation concerns.
In conclusion, our approach sets a positive precedent for fostering reproducible research in vessel trajectory forecasting for the broader scientific community while harnessing the potential of deep learning and advancing maritime safety.

\section*{Computing Environment}
We conducted experiments on probabilistic and deep-learning models for trajectory forecasting in a high-performance computing environment. The environment was equipped with an NVIDIA A100 GPU GPU with 40GB of memory and 42 CPUs with 128GB of RAM running under Ubuntu 20.04 LTS.

\section*{Data Licensing and Training Data Disclosure}
The \textit{smartWhales} initiative and its many research fronts have different license agreements with varying companies and governmental agencies. As a matter of clarification, the research we describe in this paper has used data acquired from Spire under MERIDIAN's fair-use and non-disclose data license, which is research-intended high-resolution data that is hosted at the Institute for Big Data Analytics (IBDA) from Dalhousie University. Due to licensing agreements, we are not allowed to share the raw data, but we can provide a trained model and means to retrain the models on open-source data from akin maritime regions.

\section*{Acknowledgments}
{
The nucleus of this paper was the Canadian Space Agency-funded \textit{smartWhales} – Stream 2: Prediction and Modeling -- project, led by WSP and DHI, who provided partial funding and expert supervisory (i.e., representatives from DHI and Dr. Ronald Pelot) support for this work. This funding/support was arranged through a Dalhousie-WSP MITACS Accelerate grant from the Natural Sciences and Engineering Research Council of Canada (NSERC), grant number RGPIN-2022-03909. This project received further funding from the Institute for Big Data Analytics (IBDA), and the Ocean Frontier Institute (OFI) at Dalhousie University in Halifax, NS, Canada. Furthermore, this research was supported by the São Paulo Research Foundation (FAPESP) grant number \#2016/17078-0 and \#2022/12374-1. Additionally, it received further funding from the Canada First Research Excellence Fund (CFREF) and the Canadian Foundation for Innovation's MERIDIAN cyberinfrastructure.
We thank all funding sources and {\it Spire} for providing the vessel track dataset under MERIDIAN's fair-use and non-disclose data license. The authors would like to thank Dr. Romina Gehrmann for reading the manuscript and providing detailed feedback.
}

\section*{Author contributions statement}
Conceptualization, G.S., J.K.;
methodology, G.S., J.K., R.P., and S.M.;
validation, G.S., J.K., A.S., R.F., and D.E.;
formal analysis, G.S.;
data modeling, G.S., J.K., and D.E.;
model optimization, G.S.;
data curation, G.S., J.K., D.E., J.B., and T.F.;
writing --- original draft preparation, G.S. and J.K.;
writing --- review and editing, all authors;
visualization, G.S. and J.K.;
supervision, J.B., T.F., R.F., R.P., S.M.;
funding acquisition, G.S., R.P., S.M.;
All authors have read and agreed to this version of the manuscript.

\section*{Competing interests}
The authors declare no competing interests.

\appendix

\section{Dataset Description}
Our dataset was collected from satellite-based Automatic Identification System (AIS) messages sent by vessels operating in the Gulf of St. Lawrence between 2015 and 2020.
The AIS data was acquired from Spire\footnote{~See https://spire.com.}, and it provides detailed records, including identification number, speed, location, vessel type, and timestamps.
After cleaning the tracks, our final dataset consisted of 953 cargo and 391 tanker vessels for training, 51 cargo and 21 tanker vessels for validation, and 252 cargo and 103 tanker vessels for testing.
The testing subset represents 20\% of the entire dataset, while the validation subset (for deep-learning use) represents 20\% of the non-testing data, and the rest constitutes the training dataset.

\begin{algorithm}[!ht]
\caption{Calculate grid cell Probability} 
 \begin{algorithmic}[1]
    \Require{$C_{i}$ current grid cell of new track $\tau_{n}$; $S^{C_{i}}_{\tau_{n}}$ statistics of a new track $\tau_{n} \in C_{i}$; $k$ number of destinations;}
    \Function{Possible-Destination}{$C_{i},S^{C_{i}}_{\tau_{n}},k$}
    \State $T_{C_{i}} = \left\{ ( S_{\tau} \rightarrow D_{j} ) | \tau_{C_{i}} = C_{i} , \tau \in \widetilde{\mathbf{M}} \right\} $
    \State $C_{dest}= \left\{ D_{j}| D_{j} \in T_{C_{i}} \right\}$
    \State $\mathbf{D} = \phi$
    \For {$ (S_{\tau_{i}} \rightarrow D_{j}) \in  T_{C_{i}} $}
    \State $dist =Euclidean(S_{\tau_{i}}, S^{C_{i}}_{\tau_{n}} )$
    \State $\mathbf{D} = \mathbf{D} \cup (D_{j},S_{\tau_{i}} ) \rightarrow dist $
    \EndFor
    \State $CD = \phi$
    \For {$D_{i} \in C_{dest}$}
    \State $D_{i}^{min} = min\left\{dist| dist \rightarrow \mathbf{D}(D_{j}, S_{\tau}), D_{j} = D_{i} \right\}$
    \State $CD = CD \cup (D_{i} \rightarrow D_{i}^{min})$
    \EndFor
    \State $\delta= \mu( \left\{  D_{i}^{min} | (D_{i} \rightarrow D_{i}^{min}) \in CD \right\})$
    \State $CD_{updated} = \phi$
    \For{$D_{i} \in C_{dest}$}
    \State $D_{i}^{min}=D_{i}\in CD$
    \If{$D_{i}^{min} < \delta$ }
    \State $prob_{cx} = \frac{\left\|  \left\{ \tau | \tau \in T_{C_{i}}, D_{j} \leftarrow \tau, D_{j} = D_{i} \right\} \right\| } {|T_{C_{i}}|}$
    
    \State $ \mathsf{dist}_{i} = D_{i}^{min} \times (1 - prob_{cx}) $
    \State $CD_{updated} = CD_{updated} \cup (D_{i} \rightarrow \mathsf{dist}_{i})$ 
    \EndIf
    \EndFor

    \State return $\left\{ f: D_{i} | (D_{i} \leftarrow \mathsf{dist}_{i}) \in CD_{updated}, \mathsf{dist}_{j} > \mathsf{dist}_{i} , j \neq i , i = 1, ..., k \right\}$ 
    \EndFunction
	\end{algorithmic} 
 \label{algo:pdestination}
\end{algorithm}

\section{Estimation of Route and Destination}
Our probabilistic model's main objective is to predict the vessel's most likely future location based on the current position, direction data, and historical AIS data.
We require the ocean to be segmented into deterministic grids to achieve this.
These grids integrate the movement patterns of various vessels over time based on their position data from historical AIS records.
Each cell in the grid records when and how frequently vessels have visited that particular cell, storing their motion statistics along with it.
The model operates on two levels.
Firstly, it predicts the vessel's route polygon along the trajectory.
Secondly, it also predicts the most likely end-grid cell, which represents the completion of the vessel's journey.
Both are represented by their centroids, calculated separately for each AIS message in the vessel trajectory.
The step-by-step procedure to find possible routes and destinations for a new trajectory ($\tau_{n}$) is provided in Algorithm~\ref{algo:pdestination}, which has inputs: $C_{i}$, $S^{C_{i}}_{\tau_{n}}$, and $k$.

Herein, $C_{i}$ is the grid cell related to the current position of a vessel; $S^{C_{i}}_{\tau_{n}}$ is motion statistics of the new trajectory ($\tau_{n}$) in the grid cell $C_i$; lastly, $k$ represents the number of top probable destinations to consider.
In Line 2,  $T_{C_{i}}$ is a subset from $\widetilde{\mathbf{M}}$ (defined in Table \ref{tab:notation-1}) containing historical vessel motion statistics and destination related to grid cell $C_i$.
We gather all unique possible destinations (end grid cell) $C_{dest}$ passing through $C_i$ in Line 3.
In Line 4, we initialize the set $D$ as an empty set.
Subsequently, we calculated the similarity between each historical track and the new trajectory in the current cell.
The distance for each historical track (moving to a particular destination $D_{j}$) is stored in $D$.
Lines 5--8 depicts the procedure where for each historical track's statistics $S^{C_{i}}_{\tau_{i}}$, we calculate the Euclidean distance, labeled as $\mathsf{e}$, between $S_{\tau_{i}}$ and $S^{C_{i}}_{t_{n}}$.
In the historical tracks, more than one track may pass through $C_{i}$ to a particular destination $D_{j}$.
Thus, for each unique $D_{j} \in C_{dest}$, from all historical tracks that go to $D_{j}$, we consider only one track having the minimum distance $D_{i}^{min}$ (see Lines 10--13).
The mean ($\mu$) of the population's minimum distances in $CD$ is determined in Line 14.
In $CD$, we only consider those destination $D_{j}$ whose distance is less than the mean $\delta$ --- see Equation~\ref{eq:dest-only}.
Line 16-23 depicts the procedure to store the score in $CD_{updated}$.
The score is calculated by multiplying the similarity score and probability tracking passing cell $C_{i}$ to reach a particular $D_{i}$.
After processing, the function returns a set of tuples.
Each tuple is uniquely identified by a pair of indices $(i, j)$, where $i \in [1,k]$ and $j$ denote the tuple's rank.

\section{TrAISformer Adaptations}
As already mentioned, there are only a hand-full of models that are tailored for the long-term trajectory forecasting of AIS data due to the increased uncertainty in working on a large geographical environment over a long temporal range~\cite{DBLP:journals/tits/ZhangFXXQ22}, making this problem very challenging to tackle and susceptible to many variables not intrinsically inherent to the positioning data provided by AIS messages. The prior state-of-the-art model~\cite{nguyen2024}, for instance, was developed to classify instead of regress over data, adopting a space quantization strategy that also helps reduce uncertainty and achieve better results. Such an approach adopts a sparse representation of AIS data and trains the transformer with multi-head attention. Due to problem constraints, we have adapted their loss function as follows:
\begin{equation}
    L_{total} = ( \mathbb{L}_{lon} + \mathbb{L}_{lat} + \mathbb{L}_{v}+ \mathbb{L}_{\theta} + \mathbb{L}_{\Delta_{v}} + \mathbb{L}_{\Delta_{\theta}} )
\end{equation}
The change was necessary because the original implementation of their model failed to converge over our dataset. In such cases, the loss function we adapt measures the variation within the prediction of each feature required by their model. This is a requirement because the authors are using teacher-forcing training, which is also the reason for the increased variability observed in the output results of their model. In our version of the loss function, $\mathbb{L}$ indicates the cross-entropy loss function applied over the quantized model-output results. A last modification for better performance and stability of their model over our dataset refers to masking the embedding using a left-padding strategy instead of a right-padding, matching the input and output of the different models under testing.

\section{Evaluation Metrics}
When evaluating the performance of a probabilistic model, three metrics are used: \textit{Precision}, \textit{Recall}, and \textit{F1 Score}.
\textit{Precision} measures the model's ability to accurately identify only the relevant instances and is calculated by dividing true positive (TP) predictions by the sum of true positives and false positives (FP).
\textit{Recall}, also known as sensitivity, measures the model's ability to identify all relevant instances and is calculated by dividing true positives by the sum of true positives and false negatives (FN).
\textit{F1 Score} is a performance assessment metric that combines both \textit{Precision} and \textit{Recall}, giving a more comprehensive evaluation of the model's performance by taking the harmonic mean between these two values.
To ensure a fair model evaluation, each of the metrics used was evaluated in a weighted manner.

\noindent
\begin{minipage}{0.3\linewidth}
    \begin{equation}
        \hfill
        \mathcal{P} = \frac{TP}{TP+FP}
        \hfill
    \end{equation}
    \end{minipage}
    \begin{minipage}{0.3\linewidth}
    \begin{equation}
        \hfill
        \mathcal{R} = \frac{TP}{TP + FN}
        \hfill
    \end{equation}
    \end{minipage}
    \begin{minipage}{0.4\linewidth}
    \begin{equation}
        \hfill
        F1\ Score = 2 \times \frac{{\mathcal{P} \times \mathcal{R}}}{{\mathcal{P} + \mathcal{R}}}
        \hfill
    \end{equation}
\end{minipage}
\vspace{.1cm}

To evaluate the performance of our deep-learning model, we use: Mean Absolute Error -- MAE, Mean Squared Error -- MSE, and Coefficient of Determination -- $R^2$.
The MAE measures the average size of errors in a set of predictions.
It is calculated as the average absolute differences between the predicted and actual observations.
Each difference carries equal weight.
The MAE is derived from the number of observations (n), the actual observations ($y_i$), and the predicted observations ($\hat{y}_i$).
The MSE measures the average of the squares of the errors.
It is a quality measure of an estimator, representing the average squared difference between the estimated values and the actual value.
The MSE is always non-negative, and lower values are better.
The $R^2$ score is a statistical measure that shows the proportion of the variance in a dependent variable explained by one or more independent variables in a regression model.
It indicates the model's goodness-of-fit and how well-unseen samples are likely to be predicted.
The value of $R^2$ ranges from 0 to 1, where 1 indicates the linear regression fits the data perfectly, and 0 indicates that the average is a better predictor.
In this case, $\bar{y}$ represents the mean of the observed data.

\noindent\hfill
\begin{minipage}{0.4\linewidth}
    \begin{equation}
        \hfill
        MAE = \frac{1}{n} \sum_{i=1}^{n} \left| y_i - \hat{y}_i \right|
        \hfill
    \end{equation}
\end{minipage}
\begin{minipage}{0.4\linewidth}
    \begin{equation} 
        \hfill
        MSE = \frac{1}{n} \sum_{i=1}^{n} \left( y_i - \hat{y}_i \right)^2
        \hfill
    \end{equation}
\end{minipage}
\hspace{1.5cm}

\hspace{-1.5cm}
\begin{minipage}{0.2\linewidth}\end{minipage}\hfill
\begin{minipage}{0.4\linewidth}
    \begin{equation}
        \hfill
        R^2 = 1 - \frac{\sum_{i=1}^{n} \left( y_i - \hat{y}_i \right)^2}{\sum_{i=1}^{n} \left( y_i - \bar{y} \right)^2}
        \hfill
    \end{equation}
\end{minipage}\hfill
\begin{minipage}{0.4\linewidth}\end{minipage}

\section{Speed}
The speed of a vessel is determined by the ratio of the distance between two points with respect to time.
The distance between two AIS messages (points) on the earth's surface can be determined by the distance between the coordinates of $a,b \in \tau_i$ using the great circle distance.
The Equation~\ref{eq:geodesic_distance} uses the latitudes $\mathcal{L}_{at}^{a}$ and $\mathcal{L}_{at}^{b}$, the difference $\bigtriangleup^{\mathcal{L}_{on}}_{ab}$ between the longitudes $\mathcal{L}_{on}^{a}$ and $\mathcal{L}_{on}^{b}$, and the Earth's radius $\mathbf{R}$ (6,371 km), values are given in radians.
\begin{equation}
    \hfill
    \mathbb{D}(a,b) =  \left( sin\left(\mathcal{L}_{at}^{a}\right) \times sin\left(\mathcal{L}_{at}^{b}\right) + 
    cos\left(\mathcal{L}_{at}^{a}\right) \times cos\left(\mathcal{L}_{at}^{b}\right) \times cos\left({\bigtriangleup^{\mathcal{L}_{on}}_{qk}}\right) \times \mathbf{R} \right)
    \label{eq:geodesic_distance}
    \hfill
\end{equation}
Using Equation~\ref{eq:geodesic_distance}, we calculate speed of a vessel to reach point $a$ from $b$:
\begin{equation}
    \hfill
    v(b,a) = \frac{\mathbb{D}(a, b)}{(time_{b} - time_{a})}
    \hfill
\end{equation}
Furthermore, the change in speed between locations is calculated as $\Delta_{v} = v_{2}-v_{1}$.

\section{Bearing}
The {\it bearing} is the direction of a vessel relative to the earth's surface, calculated as:
\begin{equation}
    \hfill
    \begin{aligned}
        \theta(a,b)~=~atan_2\Big(&
            sin\Big(\bigtriangleup_{\mathcal{L}_{on}}\Big) \times cos\Big(\mathcal{L}_{at}^{b}\Big),\\
            &cos\Big(\mathcal{L}_{at}^{a}\Big) \times sin\Big(\mathcal{L}_{at}^{b}\Big) - sin\Big(\mathcal{L}_{at}^{a}\Big) \times cos\Big(\mathcal{L}_{at}^{b}\Big) \times cos\Big(\bigtriangleup_{\mathcal{L}_{on}}\Big)\Big)
    \end{aligned}
    \hfill
\end{equation}

\noindent
where $\bigtriangleup_{\mathcal{L}_{on}}$ is the difference between the longitudes of two points, $\Big(\mathcal{L}_{on}^{b} - \mathcal{L}_{on}^{a}\Big)$.
Furthermore, the rate of bearing $\Delta_{\theta}$ is defined as the difference between two bearing values $\Big(\Delta_{\theta} = \theta_{2} - \theta_{1}\Big)$, adjusted to lie in the range of $[-180, 180]$ by adding or subtracting 360 from $\Delta_{\theta}$ as needed.

\section{Optimization Strategy}
The optimization function we use encapsulates the strategy of independently minimizing the Mean Absolute Error (MAE) for latitude and longitude, such as formalized below:
\begin{equation}
    \hfill
    \min_{\Theta} \left(\frac{1}{N}\sum_{i=1}^{N} \left| 
    \mathcal{L}_{at}^i - \widehat{\mathcal{L}_{at}^i} \right| + \frac{1}{N}\sum_{j=1}^{N} \left| \mathcal{L}_{on}^j - \widehat{\mathcal{L}_{on}^j} \right|\right)
    \hfill
\end{equation}
\noindent
$y=\left<\mathcal{L}_{at}, \mathcal{L}_{on}\right>$ is the original coordinates, and $\widehat y=\left<\widehat{\mathcal{L}_{at}}, \widehat{\mathcal{L}_{on}}\right>$ is the predicted ones, while $\Theta$ stands as the network parameters.
For the optimization process, we utilize the Adam optimizer with the learning rate of $1e^{-3}$ and weight decay of $1e^{-4}$ to prevent overfitting.

\section{Model Hyperparameters}
Consistent hyperparameters were utilized across all experiments to ensure an unbiased evaluation when constructing the deep learning model. 
L2 regularization was enforced with a regularization strength of 0.0005.
Incremental convolutional blocks were employed with various numbers of filter sizes (256, 256, 128) and kernel sizes (7, 5, 5), with each block being subject to the stipulated L2 regularization.
A Dropout Layer was added with a probability rate of 10\%.
The encoder-decoder architecture utilized bidirectional long- and short-term memory layers with 128 cells.
Both the kernel and the recurrent weights in these layers used the ReLU activation function and L2 regularization. 
A Position-Aware Attention Mechanism was also incorporated to enhance the model's attention to the input data's temporal sequence.
A time factor of 25\% was employed for a balance between self-attention and the capturing of longer-term dependencies within the data sequence.
Following the decoder, three dense layers were sequentially stacked.
These layers had a decreasing number of neurons (128, 128, 64), and the ReLU function was activated for each.

\bibliographystyle{elsarticle-num}
\bibliography{reference}
\end{document}